\documentclass[3p]{elsarticle} 
\journal{arXiv}
\usepackage{silence}
\usepackage{amssymb,float,url,ctable,tabularx,multirow,lineno,amsmath,amsthm,array,graphicx} 
\usepackage[utf8]{inputenc}
\usepackage[ruled,titlenumbered,linesnumbered,lined,boxed,vlined]{algorithm2e}
\let\linenumbers\nolinenumbers\nolinenumbers
\usepackage{framed,multirow,latexsym,adjustbox}

\usepackage[caption=false,font=footnotesize]{subfig}

\newcommand{\titulo}{Evaluating k-NN in the Classification of Data Streams with Concept Drift}

\usepackage{etoolbox}
\BeforeBeginEnvironment{tabularx}{\centering\fontsize{8}{5.6}\selectfont}
\BeforeBeginEnvironment{tabular}{\centering\fontsize{8}{5.6}\selectfont}
\usepackage{float}
\usepackage[smallcaps]{glossaries}
\newacronym{adob}{ADOB}{Adaptable Diversity-based Online Boosting}
\newacronym{adwin}{ADWIN}{Adaptive Windowing}
\newacronym{adwinbag}{Adwin OzaBag}{Adwin Online Bagging}
\newacronym{adwinboost}{Adwin OzaBoost}{Adwin Online Boosting}
\newacronym{aue}{AUE2}{Accuracy Updated Ensemble}
\newacronym{bole}{BOLE}{Boosting-like Online Learning Ensemble}
\newacronym{ddd}{DDD}{Diversity for Dealing with Drifts}
\newacronym{ddm}{DDM}{Drift Detection Method}
\newacronym{dde}{DDE}{Drift Detection Ensemble}
\newacronym{dwm}{DWM}{Dynamic Weighted Majority}
\newacronym{ecdd}{ECDD}{{\textsc EWMA} for Concept Drift Detection}
\newacronym{ecpf}{ECPF}{Enhanced Concept Profiling Framework}
\newacronym{eddm}{EDDM}{Early Drift Detection Method}
\newacronym{emzd}{EMZD}{Equal Means Z-Test Drift Detector}
\newacronym{ewma}{EWMA}{Exponentially Weighted Moving Average}
\newacronym{fase}{FASE}{Fast Adaptive Stacking of Ensembles}
\newacronym{fcwm}{FCWM}{Fixed Cumulative Windows Model}
\newacronym{fhddm}{FHDDM}{Fast Hoeffding Drift Detection Method}
\newacronym{flc}{FLC}{Fast and Light Classifier}
\newacronym{fpdd}{FPDD}{Fisher Proportions Drift Detector}
\newacronym{fsdd}{FSDD}{Fisher Squared Drift Detector}
\newacronym{ftdd}{FTDD}{Fisher Test Drift Detector}
\newacronym{hddm}{HDDM}{Drift Detection Methods based on Hoeffding's Bounds}
\newacronym{ht}{HT}{Hoeffding Tree}
\newacronym{knn}{$k$-NN}{$k$-Nearest Neighbors}
\newacronym{mcc}{MCC}{Matthews Correlation Coefficient}
\newacronym{moa}{MOA}{Massive Online Analysis}
\newacronym{nb}{NB}{Naive Bayes}
\newacronym{oabm}{OABM2}{Online AdaBoost-based M1 (OABM1) and M2}
\newacronym{ocboost}{OCBoost}{Online Coordinate Boosting}
\newacronym{onsb}{ONSB}{Online Non-Stationary Boosting}
\newacronym{ozabag}{OzaBag}{Oza and Russell's Online Bagging}
\newacronym{ozaboost}{OzaBoost}{Oza and Russell's Online Boosting}
\newacronym{ozabole}{OzaBole}{OzaBole}
\newacronym{leveraging}{LevBag}{Leveraging Bagging}
\newacronym{pac}{PAC}{Probably Approximately Correct}
\newacronym{pht}{PHT}{Page-Hinkley Test}
\newacronym{pl}{PL}{Paired Learners}
\newacronym{rbf}{RBF}{Radial Basic Function}
\newacronym{rddm}{RDDM}{Reactive Drift Detection Method}
\newacronym{rcd}{RCD}{Recurring Concept Drifts}
\newacronym{seed}{SEED}{SEED Drift Detector}
\newacronym{seqdrift}{SeqDrift}{Sequential Drift}
\newacronym{stepd}{STEPD}{Statistical Test of Equal Proportions}
\newacronym{wma}{WMA}{Weighted Majority Algorithm}
\newacronym{wstd}{WSTD}{Wilcoxon Rank Sum Test Drift Detector}
\newacronym{samknn}{SAM-$k$NN}{Self-adjusting Memory $k$-NN}

\newcolumntype{+}{>{\global\let\currentrowstyle\relax}}
\newcolumntype{^}{>{\currentrowstyle}}

\begin{document}

\begin{frontmatter}
\title{\titulo}

\author[ufpe]{Roberto Souto Maior de Barros\corref{label1}}
\cortext[label1]{Corresponding author}
\ead{roberto@cin.ufpe.br}

\author[ufpe]{Silas Garrido Teixeira de Carvalho Santos}
\ead{sgtcs@cin.ufpe.br}

\author[pucpr]{Jean Paul Barddal}
\ead{jean.barddal@ppgia.pucpr.br}

\address[ufpe]{Centro de Informática, Universidade Federal de Pernambuco, Cidade Universitária, 50740-560, Recife, Brazil}

\address[pucpr]{Programa de P\'{o}s-Gradua\c{c}\~{a}o em Inform\'{a}tica, Pontif\'{i}cia Universidade Cat\'{o}lica do Paran\'{a}, 80215-901, Curitiba, Brazil}

\begin{abstract}
Data streams are often defined as large amounts of data flowing continuously at high speed. 
Moreover, these data are likely subject to changes in data distribution, known as concept drift.
Given all the reasons mentioned above, learning from streams is often online and under restrictions of memory consumption and run-time.
Although many classification algorithms exist, most of the works published in the area use Naive Bayes (NB) and Hoeffding Trees (HT) as base learners in their experiments.
This article proposes an in-depth evaluation of $k$-Nearest Neighbors ($k$-NN) as a candidate for classifying data streams subjected to concept drift. 
It also analyses the complexity in time and the two main parameters of $k$-NN, i.e., the number of nearest neighbors used for predictions ($k$), and window size ($w$).  
We compare different parameter values for $k$-NN and contrast it to NB and HT both with and without a drift detector (RDDM) in many datasets. 
We formulated and answered 10 research questions which led to the conclusion that $k$-NN is a worthy candidate for data stream classification, especially when the run-time constraint is not too restrictive.
\end{abstract}


\end{frontmatter}

\linenumbers

\section{Introduction}

Data stream environments are related to large amounts of data that may be infinite and continuously flow rapidly.
Learning from such stream environments typically requires online methods and is often restricted in memory consumption and run-time.
In addition, the reading and processing of each instance are often limited to single access, an arrangement called single-pass processing. 
Finally, changes in the data distribution are expected, a phenomenon known as concept drift~\cite{gama:2014,barros:2018a,korycki:2021}.

Concept drift can be categorized in several ways, and the most common one is based on the speed of the changes. 
Such drifts are named {\em abrupt} when the changes between concepts are sudden or rapid, and are considered {\em gradual} when the transition from one concept to the next one occurs over a larger number of instances  \cite{gama:2004a,goncalves:2013a,friasblanco:2015,pesaranghader:2016}.

In the real-world, concept drifts occur for several reasons, including attacks and intrusion, mechanical or technical equipment failure, seasonal changes, movement detected from sensors, changes in the characteristic of spam messages in an e-mail, and others.
Consequently, it is essential to detect and adapt these concept drifts swiftly.

Different approaches have been proposed to learn from data streams containing concept drift \cite{gama:2014,barros:2019}.
A common instrumentation to tackle concept drift regards ensembles that may adopt a base learner together with different strategies or weighting functions to estimate the resulting classification. 
Examples of such ensembles include \gls{dwm} \cite{kolter:2007}, \gls{ddd} \cite{minku:2012}, \gls{adob} \cite{santos:2014}, \gls{bole} \cite{barros:2016}, \gls{fase} \cite{friasBlanco:2016}, \gls{oabm} \cite{santos:2020}.
In addition, some ensemble algorithms reuse previous classifiers on recurring concepts, e.g.~\gls{rcd} \cite{goncalves:2013a} and \gls{ecpf} \cite{anderson:2019}. 

Another common approach is based on concept drift detectors, lightweight methods that monitor the prediction results of a base learner and are meant to identify concept drifts in the distribution of the data explicitly. 
Many concept drift detection methods have been proposed in the literature \cite{barros:2018a,goncalves:2014}, with current emphasis on \gls{ddm} \cite{gama:2004a}, 
\gls{rddm} \cite{barros:2017a}, \gls{hddm} \cite{friasblanco:2015} with versions \gls{hddm}$_{A}$ and \gls{hddm}$_{W}$, \gls{fhddm} \cite{pesaranghader:2016}, \gls{wstd} \cite{barros:2018} and \gls{ftdd} \cite{cabral:2018}. 
It is also worth pointing out that concept drift detectors have often been used as auxiliary methods in ensembles \cite{goncalves:2013a,minku:2012,barros:2016,friasBlanco:2016,santos:2020,anderson:2019}, as well as to form ensembles that share a single base classifier \cite{du:2014,maciel:2015}, a comparatively less explored arrangement.

In this work, we target a roughly overlooked approach for classifying data streams: \gls{knn} \cite{aha:1991}.
\gls{knn} is a lazy and instance-based classifier: in the training step, instances are buffered for posterior querying and distance computation in the test step.
When using \gls{knn} in data streams, two issues arise \cite{shaker2012iblstreams,losing2016knn}.
First, how to choose $k$, the number of nearest buffered instances used for prediction.
Intuitively, larger values of $k$ reduce the impact of noisy data, yet, render the class boundaries less distinct.
Second, it is not possible to buffer instances indefinitely, as it jeopardizes both run-time and memory consumption constraints.
This is probably the main reason why \gls{knn} is overlooked in data stream mining, frequently simply dismissed based on the argument that it is too slow, and why bayesian \cite{john:1995} and decision tree \cite{domingos:2000} models are usually preferred, even though they can become very slow too.

This article aims to analyze in-depth the general performance of \gls{knn} applied to data stream environments, surveying its advantages and limitations as well as discussing the reasons for its low popularity in this context.
For this purpose, large-scale experiments were conducted using several artificial dataset generators, configured with both abrupt and gradual drift versions of several sizes, comparing \gls{knn} to the two most commonly used base classifiers in the data stream area --- \gls{nb} \cite{john:1995} and \gls{ht} \cite{domingos:2000,hulten:2001}, and run in the \gls{moa} framework~\cite{bifet:2010a}.
More specifically, these experiments were designed to answer the following research questions:

\begin{itemize}

\item \textbf{RQ1:} What is the impact of the neighborhood ($k$) on the accuracy of \gls{knn}?

\item \textbf{RQ2:} Does the accuracy of \gls{knn} benefit from concept drift detection? If so, how much does it improve with different values of $k$?

\item \textbf{RQ3:} What is the impact of the neighborhood ($k$) on the run-time of \gls{knn}?
\vspace{-1mm}
\item \textbf{RQ4:} How does concept drift detection impact the run-time of \gls{knn} for different values of $k$?
\vspace{-1mm}
\item \textbf{RQ5:} What is the impact of the window size ($w$) on the accuracy of \gls{knn}?
\vspace{-1mm}
\item \textbf{RQ6:} How does drift detection affect the accuracy of \gls{knn} given different $w$ values?
\vspace{-1mm}
\item \textbf{RQ7:} What is the impact of the window size ($w$) on \gls{knn}'s run-time?
\vspace{-1mm}
\item \textbf{RQ8:} How does concept drift detection impact the run-time of \gls{knn} for different values of $w$?
\vspace{-1mm}
\item \textbf{RQ9:} How does \gls{knn} compare to \gls{nb} and \gls{ht} in terms of accuracy? 
\vspace{-1mm}
\item \textbf{RQ10:} How does \gls{knn} compare to \gls{nb} and \gls{ht} regarding run-time?

\end{itemize}

The remainder of the document is organized into six sections. 
Section~\ref{background} reviews the main concepts and approaches involving data stream classification and concept drift.
Section~\ref{setup} shows the configuration of the experiments, also including descriptions of the datasets used in the tests. 
Section \ref{results:k} analyses the accuracy and run-time results of \gls{knn} varying k, evaluates them statistically and answers research questions {RQ1}--{RQ4}.
Section \ref{results:w} addresses the accuracy and run-time results of \gls{knn} varying $w$, evaluates them statistically, and answers research questions {RQ5}--{RQ8}.
Section \ref{comp:nb:ht} compares the best configurations of \gls{knn} with \gls{nb} and \gls{ht} and answers {RQ9} and {RQ10}.
Finally, Section~\ref{conc} concludes this article and proposes future work.

\section{Background} \label{background}

This section is divided into two subsections. 
The first describes the characteristics of an online classification and then reviews the \gls{nb}, \gls{ht}, and \gls{knn} classifiers. 
The second subsection defines concept drift, a common problem in data stream environments, and surveys some of the main approaches that deal with it. 
The contents presented here were selected to cover the topics explored in this article.

\subsection{Online Classification}

Classification is the task that distributes a set of instances into discrete classes according to relations or affinities.
More formally, from a set of $n$ instances in the $(\vec{x}, y)$ form, such that $\vec{x}$ is a $d$-dimensional vector of attributes and $y \in Y$ is the target, a classifier builds a model $f: \vec{x} \rightarrow y$ that is able to predict the unseen targets of $\vec{x}$ values with accuracy.

Data stream classification, or online classification, is a variant of the traditional batch classification.
The main difference between these approaches relies on how data is presented to the learner.
In the batch configuration, the dataset is static and entirely accessible.
In contrast, in streaming environments, instances are not readily available to the classifier for training; instead, they are presented sequentially over time, and the learner must adapt its model according to the arrival of instances from the stream \cite{GAMA:2010}.
Therefore, we denote $\mathcal{S} = [(\vec{x}^{\,t}, y^t)]_{t=0}^{n\rightarrow\infty}$ as a data stream providing instances $(\vec{x}^{\,t}, y^{t})$, each of which arriving at a timestamp $t$.

Learning from data streams shares issues common to traditional classification, such as high dimensionality, class imbalance, generalization to new data, and scalability, but it also suffers from specific issues \cite{bifet:2010,gama:2014,nguyen:2015}. 
First, the classifiers must be able to process instances sequentially, according to their arrival, and discard them right after. 
Though there is no restriction towards buffering instances for a small period of time, such as in the \gls{knn} -- algorithm that we assess in this article -- this should be done cautiously w.r.t.~memory space and processing time. 
The following subsections present three different approaches used in the online classification context.

\subsubsection{Naive Bayes}\label{survey:nb}

\acrfull{nb} is a classifier based on the Bayes theorem. 
Its rationale is to compute class probabilities based on the values of instance features and select the most likely class according to the product of the conditional probabilities between feature values and the class. 
It is referred to as a \textit{naive} classifier since it works under the assumption that all attributes are independent and that all features have the same impact on the outcome. 
Even though both assumptions are unverified in most scenarios \cite{bifet:2018a}, \gls{nb} does surprisingly well for its simplicity in many classification tasks.

Formally, considering that $\vec{x}$ is a $d$-dimensional vector of attributes of a given problem and $y \in Y$ is the associated label, the probability of a class $y$, given the $\vec{x}$ attributes, is defined by Equation \ref{naive_bayes_eq}, where $P(\vec{x})$ and $P(y)$ represent the evidence and the class prior probability, respectively, and $P(\vec{x} \mid y)$ is the likelihood. 
\begin{flalign}
    P(y \mid \vec{x}) = \frac{ P(y) P(\vec{x} \mid y) }{ P(\vec{x}) },
    \label{naive_bayes_eq}
\end{flalign}

The estimation of $P(\vec{x} \mid y)$ can be performed in several ways, such as using the binomial or multinomial distribution for discrete values or the normal distribution for continuous values.

In terms of time complexity, for each training instance, \gls{nb} will need to go through the attributes updating the metrics that will be used for the calculation of $P(\vec{x} \mid y)$ and will have an asymptotic performance of $\Theta(d)$. The frequency of the labels should also be updated to define $P(y)$, this being a constant cost operation.

On the other hand, the cost involved in the classification is related to the calculation of $P(\vec{x} \mid y)$, which will require that each of the attributes associated with the different classes is covered, demanding an $\Theta(c \times d)$ time complexity, where $c$ corresponds to the number of classes. It is important to note that the calculation of $P(\vec{x})$ (Equation \ref{naive_bayes_eq}) can be omitted since it will be the same for all classes. Thus, the probability of a class $y$ given a vector of attributes $\vec{x}$ is defined as follows:
\begin{flalign}
    P(y \mid \vec{x}) \propto P(y) \prod^{d}_{j=1} P(\vec{x}_j \mid y).
\end{flalign}

\subsubsection{Hoeffding Tree}\label{survey:ht}

Decision trees are a popular choice for classification tasks as they are simple, robust, and human-understandable. 
They are learned by recursion and replacing leaves with split nodes that guide class separation and classification. 
The definition of which attribute will be used in a split node is chosen by comparing all available features and choosing the best option according to some heuristic function $J$, e.g., Conditional Entropy, Information Gain, and Symmetrical Uncertainty. 
The splitting process is repeated on top of a set of training examples stored in main memory. As a result, classical decision trees are limited to learning from this specific set of instances and, consequently, are not tailored to evolve over time. 

In streaming scenarios, the assumption that the entire dataset is available for training does {\em not} hold and, thus, the \acrfull{ht} was proposed in \cite{domingos:2000} to learn tree models from streaming data incrementally.
\gls{ht} relaxes this constraint by comparing which feature is the most appropriate, according to $J$, based on a small data sample. 
To determine how big this sample should be, the authors \cite{domingos:2000}  proposed the use of the \textit{Hoeffding bound} \cite{hoeffding:1963}. 
Assuming the heuristic function $J$ with range $R$, the \textit{Hoeffding bound} states that, with probability $(1 - \delta)$, the true mean of $J$ is at least $(\bar{J}-\epsilon)$, where $\epsilon$ is the bound calculated following Equation \ref{eq:hoeffdingBound}.
\begin{equation}
	\epsilon = \sqrt{R^2 \times \ln(1 / \delta) / 2n}
	\label{eq:hoeffdingBound}
\end{equation}
	
Therefore, with high probability, the data distribution observed in a data sample of size $n$ adheres to the population distribution, which is potentially unbounded in data streams. 
Consequently, \gls{ht}s attempt a new split every time a leaf contains $n$ instances.
Assuming that the goal is to maximize $J$, that $X_a$ is the best-ranked feature in terms of $J$, and that $X_b$ is the second best, then a split will be performed on $X_a$ if $\Delta J = J(X_a, Y) - J(X_b; Y) \geq \epsilon$.
As a result of empirical experiments \cite{domingos:2000} noticed that reasonably small values of $n$, e.g., ~$200$, achieve effective results, and the same value has been adopted in the \gls{moa} framework, which has been used in the experiments of this article.

For training, the cost associated with \gls{ht} involves going through the tree and, when finding the leaf node, updating the corresponding statistics. 
Besides, a check should be done to indicate whether a leaf split should occur from time to time. 
All of these operations will result in time complexity of $\mathcal{O}$($l + c \times d \times v$) or simply $\mathcal{O}$($c \times d \times v$) for training, where $l$ corresponds to the height of the tree, $c$ is the number of classes, $d$ represents the number of attributes, and $v$ is the maximum number of values per attribute.

Finally, to classify an instance, the associated cost is basically going through the tree to the leaf node, with time complexity of $\mathcal{O}$($l$). 
However, a common strategy to improve the accuracy of \gls{ht} is to use \gls{nb} learners at the leaves instead of the majority class classifier \cite{bifet:2018a}. 
Thus, it is possible to calculate $P(y \mid \vec{x})$ (details on Subsection \ref{survey:nb}) using the statistics available in the leaf node, though adopting this strategy brings an additional cost, making the complexity become $\mathcal{O}$($l + c \times d$).

\subsubsection{k-Nearest Neighbors}\label{survey:knn}
	
\acrfull{knn} is one of the most fundamental, simple, and widely used classification methods, which can learn complex (non-linear) functions \cite{aha:1991}.
\gls{knn} is a lazy learner since it does not require building a model before its actual use.
It classifies unlabeled instances according to the \textit{closest} previously seen labeled ones stored in a buffer.
The definition of \textit{close} means that a distance measure is used to determine how similar or dissimilar two instances are.

Several approaches compute distances between instances. 
Although there is no optimal distance for all cases, specific scenarios can benefit from choosing the most suitable one.
A recent and in-depth study on this topic can be found at \cite{alfeilat:2019}.
As this type of analysis is outside the scope of this article, we concentrate on the most popular, which is the Euclidean distance \cite{duda:2000}, given by Equation \ref{eq:euclidianDistance}. 
The vectors $\vec{x}_i$ and $\vec{x}_j$ are two arbitrary instances and the summation occurs over all features $X_k \in \mathcal{X}$.
\begin{equation}
    \textrm{dist}_{\textrm{Euclid}}(\vec{x}_i, \vec{x}_j) = 
        \sqrt{
            \sum_{X_k \in \mathcal{X}}{\left(\vec{x}_i[X_k] - \vec{x}_j[X_k]\right)^2}
        }
    \label{eq:euclidianDistance}
\end{equation}

\gls{knn} classifies unlabeled instances according to the majority label in the $k$ closest instances and, thus, picking an appropriate value of $k$ for each application is also significant.
If $k$ is too small, \gls{knn} becomes more prone to over-fitting and tends to misclassify instances in easy situations.
Conversely, bigger values of $k$ may mislead classification when an instance is surrounded by several instances of a different label in fuzzy decision borders.

The classification of data streams using \gls{knn}  requires an additional important trait: dealing with time and memory limitations.
Continuously buffering instances as they arrive is unfeasible since the stream is potentially unbounded.
Therefore, an incremental version of \gls{knn} must \textit{forget} older instances as the stream progresses.
A simple approach for \textit{forgetting} is storing instances in a queue with size $w$.
Again, defining a value for $w$ is nontrivial, and it must be set according to constraints of memory and processing time. 

Given that a queue of size $w$ will store the most recent instances of the stream, the classification process consists of checking the distance between the current instance and the instances in $w$. 
After that, the $k$ instances with the shortest distances are selected. 
A linear implementation of \gls{knn}, which uses a \textit{heap} to keep the $k$ shorter distances as they are calculated, will have time complexity $\mathcal{O}$($w \times (d + log$ $k$)) for classification. 
On the other hand, because it is a lazy algorithm that postpones all processing for the classification period, the training will have constant time complexity, requiring only to store the instances as they arrive.

It is noteworthy highlighting that instance-based approaches such as IBLStreams \cite{shaker2012iblstreams} and \gls{samknn} \cite{losing2016knn} have been proposed for data streams.
IBLStreams provides two mechanisms for self-adjusting the neighborhood hyper-parameter $k$, such that the first accounts for the error rate obtained by ($k-1$), $k$, and ($k+1$) versions of the IBLStreams in the previous 100 instances, while the second tests kernel sizes that are coupled with centroids with 5\% deviation when compared to the current kernel size followed.
However, both mechanisms induce relevant computational overheads because multiple instances of IBLStreams are to be run in parallel.
On the other hand, \gls{samknn} proposes a combination of \gls{knn} with short-term and long-term memories.
Despite this approach not requiring explicitly determining window sizes, it does require the definition of hyper-parameters that determine the size of long-term and short-term memories, as well as the neighborhood size $k$.
In this paper, we focus solely on the original \gls{knn} algorithm and conduct an exploratory study on the impact of the neighborhood size ($k$) and window size ($w$) hyper-parameters on the performance of \gls{knn}.

\subsection{Concept Drift Problem}

Due to the temporal and ephemeral characteristics of data streams, these are expected to change their data distributions, thus giving rise to concept drifts \cite{tsymbal:2004,webb:2018}.
Following the definition in \cite{nguyen:2012}, we denote a concept $C$ as a set of class probabilities and a class-conditional probability density function as stated in $C = \bigcup_{y \in Y}{\{(P[y], P[\vec{x} \mid y])\}}$.

Consequently, given a stream $\mathcal{S}$, instances $(\vec{x}^{\,t}, y^t)$ are generated according to the current concept $C^t$.
If during every instant $t_i$ it follows that $C^{t_i} = C^{t_{i-1}}$, then the concept is stable.
Otherwise, if between any two timestamps $t_i$ and $t_j = t_i + \Delta$ it occurs that $C^{t_i} \neq C^{t_j}$, we have a concept drift.
The cause of concept drifts can neither be determined nor predicted by learning algorithms as the information behind their causes is often unavailable (hidden context), or it is too costly \cite{widmer:1996}.
As a result, learners tailored for data streams must detect and adapt to these changes automatically and autonomously.
    
Concept drifts are often categorized according to their speed.
Given that drift occurs at timestamp  $t_i$ and that it becomes stable once again at a timestamp $t_j = t_i +1$, the drift is said to be abrupt.
Otherwise, it takes more instances between the drift and the moment when the new concept becomes stable, and the drift is called gradual.

It is relatively common to use a concept drift detection method together with a base classifier to learn from data streams. 
In general, the drift detector analyses the prediction results of the base learner and applies a particular decision model to attempt to detect changes in the data distribution. 

As previously explained, many concept drift detection methods have been proposed over the years, and several of those have been previously compared \cite{barros:2018a,goncalves:2014}.
According to extensive experiments \cite{barros:2018a}, the best methods are \gls{rddm} \cite{barros:2017a} and \gls{hddm}$_{A}$ \cite{friasblanco:2015}.
Because \gls{rddm} was the detector chosen for the experiments of this article and it evolved from \gls{ddm} \cite{gama:2004a}, these two detectors are briefly explained below.

\gls{ddm} detects concept drifts in streams by analyzing the error rate and its standard deviation.
For each position ${i}$, \gls{ddm} defines the error rate $p_{i}$ as the probability of making an incorrect prediction and its standard deviation as $s_{i} =\sqrt{p_{i} \times (1-p_{i})~/~i}$. 
Based on the \gls{pac} learning model \cite{mitchell:1997}, the authors of \gls{ddm} argue that, when the distribution of the examples is stationary, the error rate ${p}_{i}$ should decrease as the number of examples ${i}$ increases.
If the error rate increases, \gls{ddm} signals a change in the data distribution, making the current base learner outdated. 

\gls{rddm} \cite{barros:2017a} was proposed to overcome/alleviate a performance loss problem of \gls{ddm} when concepts are very large, caused by decreased sensitivity, requiring too many instances to detect the changes. 
\gls{rddm} adds an explicit mechanism to discard older instances of very long concepts, periodically recalculating the \gls{ddm} statistics responsible for detecting the warning and drift levels.
Besides, it forces concept drift when the number of instances of the warning period reaches a parametrized threshold.
The authors claim \gls{rddm} delivers higher global accuracy than \gls{ddm} in most situations, especially in gradual concept drift datasets, by detecting more drifts and detecting them earlier, despite a slight increase in false positives and memory consumption.

\section{Experimental setting}~\label{setup}
This section provides all the relevant information on the experiments reported in this article. 
\gls{knn} is compared to both \acrfull{nb} and \acrfull{ht} as base learners because they are the most frequently used classifiers in experiments in the area, and their implementations are available in \gls{moa}.

The accuracy was evaluated using Prequential \cite{dawid:1984} with a sliding window of size 1,000 \cite{hidalgo:2019} as its forgetting mechanism, the default in \gls{moa}. 
In this methodology, each arriving instance is used first for testing and subsequently for training. The accuracy is based on the cumulative sum of the sequential errors over time, i.e., the loss function between the predictions and the correct values. 
As can be seen in \cite{gama:2013}, using this metric together with a forgetting mechanism is suitable and recommended for the stream context, especially with the presence of concept drift. Its use provides a fast and accurate metric, mainly because it emphasizes the most recent instances.

Six artificial dataset generators were chosen to build abrupt and gradual concept drift datasets of seven different sizes, i.e., there are seven datasets of each generator, each with 10K, 20K, 50K, 100K, 500K, 1 million, and 2 million instances, respectively. 
There are four concept drifts distributed at regular intervals in all these datasets. Thus, the size of the concepts in each dataset version of the same generator is different, covering different scenarios.
For instance, in the 100K instances datasets, the four concept drifts are in positions 20K, 40K, 60K, and 80K. 
Concept drifts were simulated by joining different concepts. 
In all the gradual concept drift datasets, the changes lasted for 500 instances and were generated by a probability function, available in the \gls{moa} framework.
It is worth pointing out that these are the same datasets tested in previous comparisons \cite{barros:2018a,barros:2019}.

The experiments using the datasets with up to 100K instances were executed 30 times to calculate the accuracy rates of the methods, and the mean results were computed with 95\% confidence intervals.
The experiments were repeated ten times in the datasets containing 500K, 1 million, and 2 million instances.

Finally, the experiments were prepared, and their results were extracted and analyzed using the MOAManager tool~\cite{maciel:2020}, available at \url{https://github.com/brunom4ciel/moamanager/}. 
They were run in a PC equipped with a Core i9-9900 processor, 32GB of DDR4 RAM, and the Ubuntu Desktop 18.04 LTS 64 bits operating system.

\subsection{Datasets}

All the dataset generators selected for the experiments reported in Sections~\ref{results:k} and \ref{results:w} have been previously used in the data stream mining area and are publicly available, either in the \gls{moa} framework, from the \gls{moa} website, or at \textit{https://sites.google.com/site/moaextensions}.

The specific generators are Agrawal, LED, MIXED, Random RBF (RandRBF), SINE, and Waveform (WAVEF).
Note thst Agrawal was used twice: Agrawal$_{1}$ (AGRAW$_1$) is based on its first five functions (F1 to F5), and Agrawal$_{2}$ (AGRAW$_2$) utilizes its remaining functions (F6 to F10), providing very different datasets. 

The Agrawal generator~\cite{agrawal:1993,hidalgo2022} stores information from people willing to receive a loan.
From this data, they belong to group A or group B. 
The attributes are salary, commission, age, education level, make of their car, zip code, the house's value, number of years the house is owned, and the total amount of the loan. 
To perform the classification, the authors proposed ten functions, each with different evaluation forms.
In addition, it is possible to add noise.

The LED generator~\cite{goncalves:2013a,friasblanco:2015,santos:2014} represents the problem of predicting the digit shown by a seven-segment LED display.
It has 24 categorical attributes, 17 of which are irrelevant, and a categorical class, with ten possible values.
Also, each attribute has a 10\% probability of being inverted (noise). 
Concept drifts are simulated by changing the position of the relevant attributes.

MIXED generator~\cite{gama:2004a,goncalves:2013a} has two boolean attributes ($v$ and $w$) and two numeric ones ($x$ and $y$). 
Each instance can be classified as positive or negative. 
They are positive if at least two of the three following conditions are met: $v, w$, and $y < 0.5 + 0.3 sin{(3\pi x)}$. 
Finally, to simulate concept drifts, the labels of the conditions above are reversed.

The RandRBF generator~\cite{maciel:2015,pears:2014} uses $n$ centroids as their centers, labels, and weights are randomly defined, and a Gaussian distribution to determine the values of $m$ attributes. 
The centroid randomly picked also determines the class label of the example.
This creates a normally distributed hyper-sphere of examples surrounding each central point with varying densities, and thus, it is a problem that is hard to learn.
Concept drifts are simulated by changing the centroids' positions over time. 
This dataset generator was used with six classes, 40 attributes, and 50 centroids.

The SINE generator~\cite{gama:2004a,baena-garcia:2006,ross:2012} has two numeric attributes ($x$ and $y$) and two contexts (Sine1 and Sine2). 
In the former, a given instance will be classified as positive if the point {$(x, y)$} is below the curve ${y = sin(x)}$. 
In the latter, the condition $y < 0.5 + 0.3 \times \sin{(3\pi x)}$ must be satisfied. 
Concept drifts can be simulated by reversing the conditions mentioned above, i.e., points below the curves become negative.

The WAVEF generator~\cite{barros:2018a,pears:2014} has three classes and 40 numerical attributes, with the last 19 used to produce noise. 
The goal of the problem is to detect the waveform generated by combining two of three base waves. 
The positions of the attributes representing a particular context are reversed to simulate drifts.

\section{First round evaluation: varying k}\label{results:k}

This section discusses the accuracy and run-time results of \gls{knn} varying its main parameter $k$ from 5 to 50 with a step of 5, presents their statistical evaluations, and then answers research questions \textbf{RQ1}--\textbf{RQ4}.

\subsection{Accuracy analysis} \label{sec:results:k:acc}
Tables~\ref{tab:det-acc-NoDet-Abr1} and \ref{tab:det-acc-NoDet-Abr2} present the accuracy results of the classifiers with no drift detector (split into two parts) and using the datasets with \textit{abrupt} changes.
Similarly, Tables~\ref{tab:det-acc-NoDet-Grad1} and \ref{tab:det-acc-NoDet-Grad2}, included in the appendix, contain the corresponding accuracy results of the classifiers with no drift detector in the datasets configured with {\em gradual} concept drifts. 
In each dataset, in these and all other tables, the best result is shown in \textbf{bold}.

\begin{table} [!t]
\caption{Mean accuracies (\%) of classifiers with no drift detector, in abrupt datasets, with 95\% confidence intervals (Part~1)} 
\vspace{1mm}
\label{tab:det-acc-NoDet-Abr1} 
\begin{adjustbox}{max width=\textwidth} 
\begin{tabular}{cccccccc} 
\toprule 
DS Type & \textbf{NO DETECTOR} & NB & kNN5 & kNN10 & kNN15 & kNN20 & kNN25 \\ 
and Size & DATASET &  HT & kNN30 & kNN35 & kNN40 & kNN45 & kNN50 \\ 
\toprule 
 & \textbf{AGRAW$_{1}$} & 57.12020$\pm$0.18 & 62.07208$\pm$0.24 & 61.19131$\pm$0.19 & 61.53291$\pm$0.21 & 61.80366$\pm$0.19 & 63.38504$\pm$0.20 \\  &  & 58.74598$\pm$0.73 & 64.58653$\pm$0.19 & 65.53644$\pm$0.14 & 66.21039$\pm$0.15 & 66.27367$\pm$0.17 & \textbf{66.46332$\pm$0.18} \\  \\[-1.72mm]
 & \textbf{AGRAW$_{2}$} & 67.57434$\pm$0.30 & 72.98843$\pm$0.17 & 73.43976$\pm$0.18 & 73.51131$\pm$0.24 & 74.15172$\pm$0.22 & 74.81765$\pm$0.27 \\  &  & 69.93518$\pm$0.29 & 76.05491$\pm$0.23 & 76.76452$\pm$0.25 & 77.55732$\pm$0.24 & 77.70291$\pm$0.24 & \textbf{78.04388$\pm$0.24} \\  \\[-1.72mm]
 & \textbf{LED} & 57.01783$\pm$0.25 & 54.62736$\pm$0.28 & 58.07230$\pm$0.31 & 59.65913$\pm$0.31 & 60.42849$\pm$0.34 & 60.88091$\pm$0.35 \\  &  & 55.86908$\pm$0.42 & 61.13414$\pm$0.35 & 61.34416$\pm$0.38 & 61.52509$\pm$0.37 & 61.58309$\pm$0.35 & \textbf{61.59758$\pm$0.31} \\  \\[-1.72mm]
ABRUPT & \textbf{MIXED} & 57.63526$\pm$0.14 & \textbf{77.63321$\pm$0.18} & 77.19439$\pm$0.15 & 76.89132$\pm$0.14 & 76.48700$\pm$0.14 & 76.28882$\pm$0.15 \\ 10K &  & 59.33083$\pm$0.40 & 75.94977$\pm$0.15 & 75.72560$\pm$0.17 & 75.45065$\pm$0.18 & 75.21337$\pm$0.21 & 74.89498$\pm$0.17 \\  \\[-1.72mm]
 & \textbf{RandRBF} & 30.87080$\pm$0.67 & \textbf{37.15521$\pm$0.35} & 36.54318$\pm$0.31 & 35.76562$\pm$0.28 & 34.45845$\pm$0.29 & 33.36792$\pm$0.31 \\  &  & 30.83089$\pm$0.67 & 32.53484$\pm$0.30 & 31.62111$\pm$0.28 & 30.90891$\pm$0.28 & 30.29441$\pm$0.29 & 29.86561$\pm$0.33 \\  \\[-1.72mm]
 & \textbf{SINE} & 56.33381$\pm$0.10 & \textbf{76.94775$\pm$0.19} & 76.81643$\pm$0.19 & 76.85020$\pm$0.22 & 76.53239$\pm$0.22 & 76.42313$\pm$0.19 \\  &  & 59.94026$\pm$0.49 & 76.20929$\pm$0.19 & 76.03047$\pm$0.18 & 75.81166$\pm$0.19 & 75.64065$\pm$0.21 & 75.44514$\pm$0.20 \\  \\[-1.72mm]
 & \textbf{WAVEF} & 76.51171$\pm$0.41 & 79.96834$\pm$0.31 & 80.95003$\pm$0.31 & 81.55781$\pm$0.30 & 81.72707$\pm$0.33 & 81.79629$\pm$0.32 \\  &  & 76.55113$\pm$0.39 & \textbf{81.90777$\pm$0.31} & 81.76758$\pm$0.31 & 81.82349$\pm$0.27 & 81.71726$\pm$0.32 & 81.76333$\pm$0.30 \\ 
\midrule 
 & \textbf{AGRAW$_{1}$} & 57.39686$\pm$0.10 & 63.80668$\pm$0.16 & 62.50295$\pm$0.14 & 62.67883$\pm$0.13 & 62.96523$\pm$0.13 & 64.66674$\pm$0.12 \\  &  & 58.22350$\pm$0.16 & 66.12309$\pm$0.11 & 67.11138$\pm$0.11 & 67.97241$\pm$0.10 & 67.99120$\pm$0.10 & \textbf{68.27175$\pm$0.11} \\  \\[-1.72mm]
 & \textbf{AGRAW$_{2}$} & 68.88348$\pm$0.19 & 77.48921$\pm$0.12 & 77.44370$\pm$0.12 & 77.57815$\pm$0.12 & 77.82935$\pm$0.12 & 78.62348$\pm$0.13 \\  &  & 71.19017$\pm$0.50 & 79.86511$\pm$0.12 & 80.88225$\pm$0.11 & 81.69915$\pm$0.11 & 82.04711$\pm$0.11 & \textbf{82.34710$\pm$0.12} \\  \\[-1.72mm]
 & \textbf{LED} & 57.37123$\pm$0.18 & 57.46840$\pm$0.20 & 61.05612$\pm$0.22 & 62.86969$\pm$0.22 & 63.69979$\pm$0.23 & 64.19162$\pm$0.23 \\  &  & 58.86503$\pm$0.65 & 64.58635$\pm$0.27 & 64.93491$\pm$0.28 & 65.19807$\pm$0.29 & 65.33686$\pm$0.27 & \textbf{65.41008$\pm$0.26} \\  \\[-1.72mm]
ABRUPT & \textbf{MIXED} & 57.98021$\pm$0.07 & \textbf{87.59652$\pm$0.10} & 87.15312$\pm$0.10 & 86.92052$\pm$0.10 & 86.50854$\pm$0.08 & 86.27512$\pm$0.10 \\ 20K &  & 66.18566$\pm$0.41 & 85.95582$\pm$0.10 & 85.68717$\pm$0.10 & 85.44034$\pm$0.11 & 85.16034$\pm$0.11 & 84.89573$\pm$0.11 \\  \\[-1.72mm]
 & \textbf{RandRBF} & 31.24778$\pm$0.65 & \textbf{37.03675$\pm$0.19} & 36.78167$\pm$0.19 & 36.06525$\pm$0.18 & 34.96643$\pm$0.20 & 33.90397$\pm$0.21 \\  &  & 31.22783$\pm$0.65 & 33.01891$\pm$0.19 & 32.18858$\pm$0.19 & 31.53600$\pm$0.22 & 30.89759$\pm$0.24 & 30.39467$\pm$0.27 \\  \\[-1.72mm]
 & \textbf{SINE} & 56.51763$\pm$0.07 & \textbf{84.87702$\pm$0.13} & 84.80284$\pm$0.12 & 84.84782$\pm$0.13 & 84.64193$\pm$0.12 & 84.52240$\pm$0.12 \\  &  & 60.38727$\pm$0.91 & 84.33946$\pm$0.12 & 84.16392$\pm$0.13 & 83.97195$\pm$0.14 & 83.79753$\pm$0.13 & 83.61137$\pm$0.13 \\  \\[-1.72mm]
 & \textbf{WAVEF} & 76.73422$\pm$0.23 & 80.62007$\pm$0.19 & 81.76582$\pm$0.20 & 82.54751$\pm$0.22 & 82.78908$\pm$0.22 & 82.91702$\pm$0.20 \\  &  & 77.99844$\pm$0.24 & 83.12000$\pm$0.17 & 83.03156$\pm$0.18 & \textbf{83.18861$\pm$0.17} & 83.03981$\pm$0.17 & 83.18730$\pm$0.17 \\ 
\midrule 
 & \textbf{AGRAW$_{1}$} & 57.56968$\pm$0.09 & 64.87636$\pm$0.10 & 63.28726$\pm$0.10 & 63.42750$\pm$0.10 & 63.71445$\pm$0.10 & 65.42671$\pm$0.09 \\  &  & 57.18060$\pm$0.31 & 67.02449$\pm$0.08 & 68.08195$\pm$0.08 & 69.04898$\pm$0.08 & 69.06229$\pm$0.08 & \textbf{69.35844$\pm$0.08} \\  \\[-1.72mm]
 & \textbf{AGRAW$_{2}$} & 69.76296$\pm$0.10 & 80.12240$\pm$0.08 & 79.78874$\pm$0.08 & 79.90658$\pm$0.09 & 80.00099$\pm$0.09 & 80.82811$\pm$0.09 \\  &  & 70.65264$\pm$0.51 & 82.04003$\pm$0.08 & 83.23802$\pm$0.07 & 84.08418$\pm$0.07 & 84.54137$\pm$0.08 & \textbf{84.81803$\pm$0.07} \\  \\[-1.72mm]
 & \textbf{LED} & 57.60129$\pm$0.12 & 59.04122$\pm$0.16 & 62.81684$\pm$0.18 & 64.77287$\pm$0.20 & 65.72737$\pm$0.19 & 66.25663$\pm$0.20 \\  &  & 63.20826$\pm$0.42 & 66.69850$\pm$0.21 & 67.08361$\pm$0.20 & 67.40539$\pm$0.21 & 67.61166$\pm$0.21 & \textbf{67.68996$\pm$0.22} \\  \\[-1.72mm]
ABRUPT & \textbf{MIXED} & 58.21799$\pm$0.04 & \textbf{93.56835$\pm$0.06} & 93.17373$\pm$0.05 & 92.92120$\pm$0.06 & 92.60055$\pm$0.05 & 92.34264$\pm$0.06 \\ 50K &  & 60.67344$\pm$0.41 & 92.03866$\pm$0.06 & 91.77327$\pm$0.06 & 91.47447$\pm$0.07 & 91.19925$\pm$0.07 & 90.88293$\pm$0.07 \\  \\[-1.72mm]
 & \textbf{RandRBF} & 31.31736$\pm$0.68 & \textbf{36.93527$\pm$0.14} & 36.78927$\pm$0.12 & 36.14647$\pm$0.12 & 35.08254$\pm$0.13 & 34.10981$\pm$0.15 \\  &  & 33.16774$\pm$0.33 & 33.26210$\pm$0.15 & 32.42540$\pm$0.17 & 31.77017$\pm$0.17 & 31.10598$\pm$0.19 & 30.60311$\pm$0.18 \\  \\[-1.72mm]
 & \textbf{SINE} & 56.60566$\pm$0.03 & 89.66659$\pm$0.08 & 89.62029$\pm$0.07 & \textbf{89.74129$\pm$0.08} & 89.54863$\pm$0.07 & 89.44204$\pm$0.07 \\  &  & 61.04167$\pm$0.82 & 89.23202$\pm$0.06 & 89.07614$\pm$0.07 & 88.84566$\pm$0.07 & 88.68014$\pm$0.07 & 88.46516$\pm$0.07 \\  \\[-1.72mm]
 & \textbf{WAVEF} & 76.82635$\pm$0.13 & 80.96663$\pm$0.14 & 82.16233$\pm$0.14 & 83.01765$\pm$0.13 & 83.44569$\pm$0.12 & 83.58048$\pm$0.12 \\  &  & 78.40423$\pm$0.21 & 83.79040$\pm$0.13 & 83.75321$\pm$0.14 & \textbf{83.92843$\pm$0.13} & 83.78734$\pm$0.13 & 83.92144$\pm$0.13 \\ 
\midrule 
 & \textbf{AGRAW$_{1}$} & 57.61282$\pm$0.06 & 65.22452$\pm$0.08 & 63.56096$\pm$0.07 & 63.66035$\pm$0.08 & 63.92896$\pm$0.07 & 65.64267$\pm$0.06 \\  &  & 58.59351$\pm$0.29 & 67.30407$\pm$0.05 & 68.39934$\pm$0.05 & 69.39997$\pm$0.05 & 69.38493$\pm$0.05 & \textbf{69.68796$\pm$0.05} \\  \\[-1.72mm]
 & \textbf{AGRAW$_{2}$} & 70.08073$\pm$0.07 & 80.95828$\pm$0.07 & 80.49765$\pm$0.05 & 80.64571$\pm$0.06 & 80.65595$\pm$0.04 & 81.53168$\pm$0.05 \\  &  & 71.01165$\pm$0.40 & 82.75624$\pm$0.05 & 84.01806$\pm$0.04 & 84.87308$\pm$0.05 & 85.37617$\pm$0.05 & \textbf{85.65303$\pm$0.05} \\  \\[-1.72mm]
 & \textbf{LED} & 57.90245$\pm$0.08 & 59.65174$\pm$0.13 & 63.49029$\pm$0.15 & 65.51347$\pm$0.15 & 66.50295$\pm$0.14 & 67.02911$\pm$0.15 \\  &  & 64.95586$\pm$0.43 & 67.48530$\pm$0.15 & 67.89315$\pm$0.14 & 68.20152$\pm$0.15 & 68.42365$\pm$0.14 & \textbf{68.53887$\pm$0.15} \\  \\[-1.72mm]
ABRUPT & \textbf{MIXED} & 58.32640$\pm$0.02 & \textbf{95.52986$\pm$0.03} & 95.12888$\pm$0.03 & 94.86831$\pm$0.04 & 94.53941$\pm$0.04 & 94.29309$\pm$0.04 \\ 100K &  & 69.75154$\pm$0.60 & 94.01146$\pm$0.03 & 93.74265$\pm$0.03 & 93.45495$\pm$0.03 & 93.17230$\pm$0.04 & 92.84069$\pm$0.04 \\  \\[-1.72mm]
 & \textbf{RandRBF} & 31.51460$\pm$0.61 & \textbf{36.99664$\pm$0.08} & 36.87841$\pm$0.09 & 36.26443$\pm$0.09 & 35.15248$\pm$0.08 & 34.19674$\pm$0.08 \\  &  & 34.85948$\pm$0.24 & 33.33276$\pm$0.09 & 32.53686$\pm$0.09 & 31.87191$\pm$0.10 & 31.21516$\pm$0.12 & 30.71920$\pm$0.12 \\  \\[-1.72mm]
 & \textbf{SINE} & 56.66589$\pm$0.03 & 91.22266$\pm$0.05 & 91.22932$\pm$0.04 & \textbf{91.37965$\pm$0.04} & 91.15904$\pm$0.05 & 91.06340$\pm$0.04 \\  &  & 66.02926$\pm$0.59 & 90.84745$\pm$0.04 & 90.67319$\pm$0.05 & 90.43821$\pm$0.04 & 90.25799$\pm$0.04 & 90.05248$\pm$0.04 \\  \\[-1.72mm]
 & \textbf{WAVEF} & 76.89031$\pm$0.09 & 81.01234$\pm$0.11 & 82.36969$\pm$0.08 & 83.19111$\pm$0.09 & 83.64096$\pm$0.08 & 83.79574$\pm$0.08 \\  &  & 79.64533$\pm$0.18 & 84.05396$\pm$0.08 & 83.99380$\pm$0.08 & 84.19858$\pm$0.08 & 84.05922$\pm$0.08 & \textbf{84.21782$\pm$0.08} \\ 
\bottomrule 
\end{tabular} 
\end{adjustbox} 
\end{table}

First, we analyze the impact of different values of $k$ w.r.t.~mean accuracy rates. 
Overall, there is no clear trend or consensus on an optimal $k$ value, as it varies in terms of the data stream generator and marginally according to the concept length in the experiment.
For instance, in AGRAW$_1$, AGRAW$_2$, LED, and WAVEF datasets, higher $k$ values usually yielded the most effective accuracy rates, regardless of their length.  
These results are due to the noise trait of these experiments, as 10\% of the instances have their labels changed in the Agrawal datasets, and 10\% of the binary features are swapped in the LED datasets.
Consequently, \gls{knn} improves its results with a larger $k$ because using more neighbors makes the classification more reliable.
We also highlight that, in the Waveform datasets, the accuracy of \gls{knn} increases with $k$ up to $30$. From this point on, all configurations seem to deliver accuracy rates within a single standard deviation.

The opposite behavior was observed in the MIXED, RandRBF, and, to some extent, SINE datasets, where smaller $k$ values achieved the highest accuracy rates.
Nonetheless, the volatility of the accuracy rates obtained across different generators is relevant and should be accounted for.
In RandRBF experiments, we see that the accuracy rates between $k=5$ and $k=50$ vary from 6 to 7\%, due to the potential overlap between classes over time, as this generator creates drifting centroids that may occupy the same region in the feature space.
Consequently, larger values of $k$ make the decision fuzzier, as instances from both classes are considered close to the instances.
The results for MIXED and SINE show smaller accuracy differences across different $k$ values, as these generators are easier to learn and not as volatile.
Also, the accuracy rates for $k \in \{5,10, 15\}$ are usually within the standard deviations observed.

Regarding the results of \gls{knn} in the gradual datasets (reported in Tables \ref{tab:det-acc-NoDet-Grad1} and \ref{tab:det-acc-NoDet-Grad2}), the accuracy rates observed with different $k$ values are similar to those observed in the abrupt scenarios, i.e.,~there is no clear evidence that gradual concept drifts have any significant impact on the accuracy of \gls{knn} concerning the corresponding abrupt scenarios.

\begin{table}[!ht]
\caption{Mean accuracies (\%) of classifiers with no drift detector, in abrupt datasets, with 95\% confidence intervals (Part~2)} 
\vspace{1mm}
\label{tab:det-acc-NoDet-Abr2} 
\begin{adjustbox}{max width=\textwidth} 
\begin{tabular}{cccccccc} 
\toprule 
DS Type & \textbf{NO DETECTOR} & NB & kNN5 & kNN10 & kNN15 & kNN20 & kNN25 \\ 
and Size & DATASET &  HT & kNN30 & kNN35 & kNN40 & kNN45 & kNN50 \\ 
\toprule 
 & \textbf{AGRAW$_{1}$} & 57.69231$\pm$0.05 & 65.50908$\pm$0.07 & 63.77436$\pm$0.06 & 63.85746$\pm$0.07 & 64.11499$\pm$0.06 & 65.85320$\pm$0.06 \\  &  & 59.53245$\pm$0.63 & 67.55882$\pm$0.05 & 68.64230$\pm$0.04 & 69.69198$\pm$0.04 & 69.66275$\pm$0.03 & \textbf{69.98295$\pm$0.04} \\  \\[-1.72mm]
 & \textbf{AGRAW$_{2}$} & 70.42176$\pm$0.05 & 81.73313$\pm$0.05 & 81.15546$\pm$0.06 & 81.31829$\pm$0.05 & 81.26086$\pm$0.05 & 82.17071$\pm$0.05 \\  &  & 72.84391$\pm$0.49 & 83.38166$\pm$0.06 & 84.68610$\pm$0.05 & 85.53365$\pm$0.05 & 86.05187$\pm$0.05 & \textbf{86.31952$\pm$0.05} \\  \\[-1.72mm]
 & \textbf{LED} & 58.14193$\pm$0.07 & 60.03741$\pm$0.12 & 63.95148$\pm$0.12 & 65.96053$\pm$0.11 & 66.97612$\pm$0.11 & 67.55199$\pm$0.13 \\  &  & 67.00682$\pm$0.48 & 68.05009$\pm$0.13 & 68.48258$\pm$0.13 & 68.82370$\pm$0.14 & 69.05436$\pm$0.14 & \textbf{69.19056$\pm$0.13} \\  \\[-1.72mm]
ABRUPT & \textbf{MIXED} & 58.40597$\pm$0.01 & \textbf{97.14026$\pm$0.02} & 96.73072$\pm$0.03 & 96.47924$\pm$0.02 & 96.13575$\pm$0.03 & 95.88576$\pm$0.04 \\ 500K &  & 78.26126$\pm$0.50 & 95.59294$\pm$0.04 & 95.31443$\pm$0.04 & 95.02668$\pm$0.05 & 94.75632$\pm$0.04 & 94.43727$\pm$0.05 \\  \\[-1.72mm]
 & \textbf{RandRBF} & 33.26192$\pm$0.76 & 36.95659$\pm$0.08 & 36.88153$\pm$0.08 & 36.30358$\pm$0.10 & 35.21507$\pm$0.10 & 34.24451$\pm$0.10 \\  &  & \textbf{38.19238$\pm$0.11} & 33.40723$\pm$0.08 & 32.63635$\pm$0.09 & 31.97791$\pm$0.10 & 31.30445$\pm$0.11 & 30.76836$\pm$0.13 \\  \\[-1.72mm]
 & \textbf{SINE} & 56.66710$\pm$0.03 & 92.47998$\pm$0.06 & 92.52178$\pm$0.03 & \textbf{92.65842$\pm$0.04} & 92.45834$\pm$0.04 & 92.35749$\pm$0.04 \\  &  & 76.28811$\pm$0.46 & 92.12264$\pm$0.04 & 91.96429$\pm$0.04 & 91.73104$\pm$0.03 & 91.55919$\pm$0.04 & 91.35723$\pm$0.04 \\  \\[-1.72mm]
 & \textbf{WAVEF} & 76.84963$\pm$0.11 & 81.07676$\pm$0.10 & 82.37848$\pm$0.12 & 83.24047$\pm$0.10 & 83.70060$\pm$0.09 & 83.85490$\pm$0.09 \\  &  & 81.57844$\pm$0.10 & 84.14357$\pm$0.08 & 84.11968$\pm$0.10 & 84.33729$\pm$0.06 & 84.20460$\pm$0.09 & \textbf{84.39534$\pm$0.06} \\ 
\midrule 
 & \textbf{AGRAW$_{1}$} & 57.69537$\pm$0.03 & 65.53135$\pm$0.06 & 63.77890$\pm$0.06 & 63.86685$\pm$0.06 & 64.13347$\pm$0.06 & 65.87793$\pm$0.06 \\  &  & 60.99472$\pm$0.36 & 67.58928$\pm$0.06 & 68.69359$\pm$0.05 & 69.74026$\pm$0.05 & 69.70826$\pm$0.04 & \textbf{70.02047$\pm$0.05} \\  \\[-1.72mm]
 & \textbf{AGRAW$_{2}$} & 70.43582$\pm$0.03 & 81.80239$\pm$0.03 & 81.21614$\pm$0.04 & 81.36623$\pm$0.04 & 81.32004$\pm$0.03 & 82.22983$\pm$0.03 \\  &  & 73.57298$\pm$0.48 & 83.44312$\pm$0.04 & 84.74898$\pm$0.03 & 85.59991$\pm$0.03 & 86.12048$\pm$0.03 & \textbf{86.39081$\pm$0.03} \\  \\[-1.72mm]
 & \textbf{LED} & 58.24221$\pm$0.05 & 60.10004$\pm$0.09 & 64.02587$\pm$0.09 & 66.05732$\pm$0.07 & 67.06246$\pm$0.07 & 67.64482$\pm$0.06 \\  &  & 67.15760$\pm$0.37 & 68.13377$\pm$0.06 & 68.56795$\pm$0.06 & 68.88959$\pm$0.08 & 69.11694$\pm$0.07 & \textbf{69.25782$\pm$0.07} \\  \\[-1.72mm]
ABRUPT & \textbf{MIXED} & 58.42200$\pm$0.01 & \textbf{97.34757$\pm$0.02} & 96.93541$\pm$0.03 & 96.68606$\pm$0.02 & 96.34847$\pm$0.04 & 96.09371$\pm$0.04 \\ 1M &  & 82.25671$\pm$0.43 & 95.79916$\pm$0.03 & 95.53200$\pm$0.03 & 95.24143$\pm$0.04 & 94.95418$\pm$0.03 & 94.64395$\pm$0.04 \\  \\[-1.72mm]
 & \textbf{RandRBF} & 33.16377$\pm$0.50 & 36.90720$\pm$0.05 & 36.83380$\pm$0.05 & 36.27369$\pm$0.04 & 35.17608$\pm$0.06 & 34.21259$\pm$0.06 \\  &  & \textbf{38.75896$\pm$0.09} & 33.38066$\pm$0.05 & 32.60511$\pm$0.06 & 31.96297$\pm$0.06 & 31.29387$\pm$0.06 & 30.77123$\pm$0.08 \\  \\[-1.72mm]
 & \textbf{SINE} & 56.67992$\pm$0.01 & 92.64251$\pm$0.03 & 92.67912$\pm$0.02 & \textbf{92.82875$\pm$0.03} & 92.60895$\pm$0.03 & 92.50899$\pm$0.03 \\  &  & 80.45457$\pm$0.47 & 92.27854$\pm$0.03 & 92.12116$\pm$0.04 & 91.88926$\pm$0.03 & 91.72696$\pm$0.04 & 91.51709$\pm$0.03 \\  \\[-1.72mm]
 & \textbf{WAVEF} & 76.89927$\pm$0.07 & 81.09270$\pm$0.08 & 82.43056$\pm$0.08 & 83.28578$\pm$0.08 & 83.74732$\pm$0.06 & 83.90445$\pm$0.07 \\  &  & 82.32906$\pm$0.13 & 84.18152$\pm$0.08 & 84.15985$\pm$0.08 & 84.37260$\pm$0.07 & 84.24986$\pm$0.09 & \textbf{84.43188$\pm$0.07} \\ 
\midrule 
 & \textbf{AGRAW$_{1}$} & 57.69973$\pm$0.02 & 65.55123$\pm$0.04 & 63.79191$\pm$0.04 & 63.87714$\pm$0.03 & 64.13812$\pm$0.04 & 65.89251$\pm$0.05 \\  &  & 62.48678$\pm$0.18 & 67.60924$\pm$0.05 & 68.72755$\pm$0.04 & 69.77334$\pm$0.04 & 69.75021$\pm$0.04 & \textbf{70.05694$\pm$0.03} \\  \\[-1.72mm]
 & \textbf{AGRAW$_{2}$} & 70.45261$\pm$0.04 & 81.84175$\pm$0.01 & 81.26948$\pm$0.02 & 81.41558$\pm$0.02 & 81.36381$\pm$0.02 & 82.26228$\pm$0.02 \\  &  & 73.78709$\pm$0.27 & 83.47661$\pm$0.03 & 84.79211$\pm$0.03 & 85.64392$\pm$0.02 & 86.16450$\pm$0.03 & \textbf{86.44050$\pm$0.03} \\  \\[-1.72mm]
 & \textbf{LED} & 58.29208$\pm$0.02 & 60.15641$\pm$0.04 & 64.07662$\pm$0.06 & 66.11551$\pm$0.05 & 67.11807$\pm$0.03 & 67.71984$\pm$0.05 \\  &  & 66.70133$\pm$0.27 & 68.20802$\pm$0.06 & 68.64013$\pm$0.07 & 68.97298$\pm$0.08 & 69.19361$\pm$0.08 & \textbf{69.34025$\pm$0.07} \\  \\[-1.72mm]
ABRUPT & \textbf{MIXED} & 58.41316$\pm$0.01 & \textbf{97.43347$\pm$0.01} & 97.02845$\pm$0.01 & 96.77651$\pm$0.02 & 96.44044$\pm$0.02 & 96.18036$\pm$0.02 \\ 2M &  & 85.81355$\pm$0.20 & 95.88783$\pm$0.02 & 95.61440$\pm$0.02 & 95.31986$\pm$0.02 & 95.03042$\pm$0.02 & 94.71729$\pm$0.02 \\  \\[-1.72mm]
 & \textbf{RandRBF} & 33.06870$\pm$0.22 & 36.91998$\pm$0.03 & 36.86287$\pm$0.03 & 36.29838$\pm$0.05 & 35.18788$\pm$0.06 & 34.22267$\pm$0.05 \\  &  & \textbf{39.47914$\pm$0.07} & 33.37684$\pm$0.05 & 32.59580$\pm$0.05 & 31.95864$\pm$0.05 & 31.28713$\pm$0.05 & 30.77411$\pm$0.05 \\  \\[-1.72mm]
 & \textbf{SINE} & 56.68547$\pm$0.01 & 92.72358$\pm$0.02 & 92.75267$\pm$0.01 & \textbf{92.90522$\pm$0.02} & 92.68132$\pm$0.01 & 92.58178$\pm$0.02 \\  &  & 84.18714$\pm$0.55 & 92.35014$\pm$0.02 & 92.19838$\pm$0.02 & 91.96838$\pm$0.02 & 91.80106$\pm$0.02 & 91.58894$\pm$0.02 \\  \\[-1.72mm]
 & \textbf{WAVEF} & 76.92613$\pm$0.03 & 81.11208$\pm$0.05 & 82.45996$\pm$0.05 & 83.29463$\pm$0.04 & 83.73856$\pm$0.05 & 83.88522$\pm$0.05 \\  &  & 82.90690$\pm$0.07 & 84.17064$\pm$0.05 & 84.13947$\pm$0.06 & 84.37035$\pm$0.04 & 84.25641$\pm$0.04 & \textbf{84.43081$\pm$0.04} \\ 
\bottomrule 
\end{tabular} 
\end{adjustbox} 
\end{table}

Tables~\ref{tab:det-acc-RDDM-Abr1} and \ref{tab:det-acc-RDDM-Abr2} are similar to Tables~\ref{tab:det-acc-NoDet-Abr1} and \ref{tab:det-acc-NoDet-Abr2}, respectively, but refer to the results of the classifiers in the {\em abrupt} datasets using \gls{rddm} as drift detector.
Likewise, Tables~\ref{tab:det-acc-RDDM-Grad1} and \ref{tab:det-acc-RDDM-Grad2} are similar to Tables~\ref{tab:det-acc-NoDet-Grad1} and \ref{tab:det-acc-NoDet-Grad2}, respectively, but refer to the results of the classifiers in the {\em gradual} datasets using \gls{rddm} to detect concept drifts.
\gls{rddm} was chosen merely to represent the use of a concept drift detector together with the classifiers mainly because it is one of the state-of-the-art detectors \cite{barros:2018a}.

\begin{table}[!t] 
\caption{Mean accuracies (\%) of classifiers with the RDDM detector, in abrupt datasets, with 95\% confidence intervals (Part~1)} 
\vspace{1mm}
\label{tab:det-acc-RDDM-Abr1} 
\begin{adjustbox}{max width=\textwidth} 
\begin{tabular}{cccccccc} 
\toprule 
DS Type & \textbf{RDDM} & NB & kNN5 & kNN10 & kNN15 & kNN20 & kNN25 \\ 
and Size & DATASET &  HT & kNN30 & kNN35 & kNN40 & kNN45 & kNN50 \\ 
\toprule 
 & \textbf{AGRAW$_{1}$} & 63.55770$\pm$0.28 & 62.49409$\pm$0.26 & 62.61130$\pm$0.26 & 63.39169$\pm$0.35 & 64.32008$\pm$0.30 & 65.32636$\pm$0.20 \\  &  & 64.68765$\pm$0.32 & 66.19828$\pm$0.21 & 66.62102$\pm$0.22 & 67.02582$\pm$0.22 & 66.84856$\pm$0.20 & \textbf{67.03346$\pm$0.20} \\  \\[-1.72mm]
 & \textbf{AGRAW$_{2}$} & 79.62920$\pm$1.01 & 75.43796$\pm$0.30 & 74.98644$\pm$0.23 & 75.94761$\pm$0.32 & 76.66582$\pm$0.32 & 77.80344$\pm$0.40 \\  &  & \textbf{81.57601$\pm$0.88} & 78.97396$\pm$0.28 & 80.03182$\pm$0.36 & 80.49522$\pm$0.28 & 80.82331$\pm$0.32 & 81.12794$\pm$0.31 \\  \\[-1.72mm]
 & \textbf{LED} & \textbf{69.80327$\pm$0.30} & 55.80439$\pm$0.28 & 59.26340$\pm$0.33 & 60.74923$\pm$0.36 & 61.50390$\pm$0.37 & 61.90020$\pm$0.38 \\  &  & 69.78214$\pm$0.31 & 62.05060$\pm$0.41 & 62.09756$\pm$0.43 & 62.22542$\pm$0.42 & 62.13076$\pm$0.39 & 62.03725$\pm$0.38 \\  \\[-1.72mm]
ABRUPT & \textbf{MIXED} & 90.21909$\pm$0.24 & \textbf{95.27387$\pm$0.16} & 94.29205$\pm$0.17 & 93.51365$\pm$0.14 & 92.74100$\pm$0.16 & 92.18349$\pm$0.18 \\ 10K &  & 90.16766$\pm$0.25 & 91.45310$\pm$0.18 & 90.87160$\pm$0.20 & 90.25931$\pm$0.21 & 89.77303$\pm$0.25 & 89.18898$\pm$0.25 \\  \\[-1.72mm]
 & \textbf{RandRBF} & 30.52849$\pm$0.45 & \textbf{36.69680$\pm$0.35} & 35.71990$\pm$0.31 & 34.82579$\pm$0.30 & 33.59107$\pm$0.29 & 32.62478$\pm$0.32 \\  &  & 32.01296$\pm$0.41 & 31.80876$\pm$0.33 & 30.95617$\pm$0.27 & 30.42865$\pm$0.30 & 29.85043$\pm$0.28 & 29.48834$\pm$0.34 \\  \\[-1.72mm]
 & \textbf{SINE} & 86.57963$\pm$0.25 & \textbf{90.52642$\pm$0.16} & 90.20277$\pm$0.18 & 90.14786$\pm$0.17 & 89.72073$\pm$0.17 & 89.44980$\pm$0.17 \\  &  & 87.98444$\pm$0.21 & 89.07562$\pm$0.20 & 88.80814$\pm$0.21 & 88.42099$\pm$0.19 & 88.14065$\pm$0.20 & 87.82981$\pm$0.21 \\  \\[-1.72mm]
 & \textbf{WAVEF} & 79.12226$\pm$0.49 & 79.94701$\pm$0.31 & 80.93873$\pm$0.31 & 81.54583$\pm$0.31 & 81.65007$\pm$0.32 & 81.72529$\pm$0.33 \\  &  & 79.09170$\pm$0.49 & \textbf{81.83742$\pm$0.31} & 81.67058$\pm$0.32 & 81.70516$\pm$0.27 & 81.54260$\pm$0.32 & 81.56632$\pm$0.31 \\ 
\midrule 
 & \textbf{AGRAW$_{1}$} & 64.89027$\pm$0.16 & 63.91970$\pm$0.17 & 63.02182$\pm$0.14 & 63.53720$\pm$0.16 & 64.13146$\pm$0.22 & 65.48782$\pm$0.17 \\  &  & 68.19471$\pm$0.47 & 66.71481$\pm$0.15 & 67.54881$\pm$0.14 & 68.29270$\pm$0.13 & 68.21893$\pm$0.12 & \textbf{68.37070$\pm$0.12} \\  \\[-1.72mm]
 & \textbf{AGRAW$_{2}$} & 83.17975$\pm$0.58 & 78.45847$\pm$0.13 & 78.11071$\pm$0.11 & 78.49263$\pm$0.12 & 78.87740$\pm$0.15 & 80.00171$\pm$0.21 \\  &  & 83.11050$\pm$1.36 & 81.15114$\pm$0.13 & 82.26790$\pm$0.14 & 83.06827$\pm$0.13 & 83.48065$\pm$0.15 & \textbf{83.72279$\pm$0.13} \\  \\[-1.72mm]
 & \textbf{LED} & \textbf{71.73709$\pm$0.17} & 57.98540$\pm$0.19 & 61.61446$\pm$0.23 & 63.39619$\pm$0.23 & 64.21863$\pm$0.23 & 64.61079$\pm$0.25 \\  &  & 71.72671$\pm$0.17 & 64.96785$\pm$0.29 & 65.23374$\pm$0.30 & 65.43257$\pm$0.30 & 65.49370$\pm$0.28 & 65.54158$\pm$0.28 \\  \\[-1.72mm]
ABRUPT & \textbf{MIXED} & 91.02909$\pm$0.15 & \textbf{96.38936$\pm$0.11} & 95.68345$\pm$0.10 & 95.21852$\pm$0.09 & 94.63870$\pm$0.10 & 94.20579$\pm$0.12 \\ 20K &  & 90.66064$\pm$0.15 & 93.69632$\pm$0.12 & 93.24017$\pm$0.13 & 92.82884$\pm$0.13 & 92.43167$\pm$0.15 & 91.94940$\pm$0.13 \\  \\[-1.72mm]
 & \textbf{RandRBF} & 30.49600$\pm$0.43 & \textbf{36.87325$\pm$0.18} & 36.32554$\pm$0.20 & 35.55257$\pm$0.18 & 34.42648$\pm$0.22 & 33.44891$\pm$0.23 \\  &  & 32.30486$\pm$0.39 & 32.56359$\pm$0.20 & 31.78764$\pm$0.21 & 31.15090$\pm$0.25 & 30.55632$\pm$0.27 & 30.17284$\pm$0.29 \\  \\[-1.72mm]
 & \textbf{SINE} & 87.01810$\pm$0.20 & \textbf{91.64419$\pm$0.13} & 91.46534$\pm$0.14 & 91.53199$\pm$0.14 & 91.22076$\pm$0.11 & 91.04457$\pm$0.12 \\  &  & 89.46259$\pm$0.15 & 90.75913$\pm$0.14 & 90.54142$\pm$0.13 & 90.23128$\pm$0.14 & 90.01220$\pm$0.14 & 89.74820$\pm$0.14 \\  \\[-1.72mm]
 & \textbf{WAVEF} & 79.78200$\pm$0.30 & 80.59979$\pm$0.19 & 81.76433$\pm$0.21 & 82.52551$\pm$0.22 & 82.78075$\pm$0.22 & 82.85530$\pm$0.19 \\  &  & 79.63755$\pm$0.27 & 83.08884$\pm$0.17 & 82.98806$\pm$0.19 & \textbf{83.12561$\pm$0.17} & 82.92451$\pm$0.17 & 83.08213$\pm$0.17 \\ 
\midrule 
 & \textbf{AGRAW$_{1}$} & 65.73419$\pm$0.11 & 64.86531$\pm$0.11 & 63.41475$\pm$0.10 & 63.77451$\pm$0.10 & 64.30246$\pm$0.14 & 65.66311$\pm$0.11 \\  &  & \textbf{72.42915$\pm$0.32} & 67.17293$\pm$0.09 & 68.15098$\pm$0.10 & 69.11415$\pm$0.10 & 69.08979$\pm$0.09 & 69.35015$\pm$0.09 \\  \\[-1.72mm]
 & \textbf{AGRAW$_{2}$} & 85.38333$\pm$0.21 & 80.41578$\pm$0.10 & 79.99267$\pm$0.08 & 80.21891$\pm$0.08 & 80.39715$\pm$0.11 & 81.24254$\pm$0.10 \\  &  & \textbf{86.09054$\pm$0.20} & 82.45358$\pm$0.08 & 83.67901$\pm$0.07 & 84.53209$\pm$0.08 & 85.01271$\pm$0.08 & 85.29304$\pm$0.07 \\  \\[-1.72mm]
 & \textbf{LED} & \textbf{72.88781$\pm$0.15} & 59.19590$\pm$0.16 & 63.00111$\pm$0.18 & 64.94460$\pm$0.21 & 65.87737$\pm$0.20 & 66.39327$\pm$0.20 \\  &  & 72.88391$\pm$0.15 & 66.81357$\pm$0.21 & 67.19301$\pm$0.21 & 67.48485$\pm$0.22 & 67.66793$\pm$0.21 & 67.72263$\pm$0.23 \\  \\[-1.72mm]
ABRUPT & \textbf{MIXED} & 91.56782$\pm$0.11 & \textbf{97.04822$\pm$0.05} & 96.52886$\pm$0.05 & 96.19134$\pm$0.06 & 95.77862$\pm$0.06 & 95.45618$\pm$0.08 \\ 50K &  & 91.59759$\pm$0.15 & 95.08059$\pm$0.07 & 94.76080$\pm$0.07 & 94.38774$\pm$0.08 & 94.05739$\pm$0.09 & 93.67819$\pm$0.08 \\  \\[-1.72mm]
 & \textbf{RandRBF} & 30.72685$\pm$0.37 & \textbf{36.86124$\pm$0.13} & 36.65105$\pm$0.12 & 35.97598$\pm$0.13 & 34.85450$\pm$0.14 & 33.91131$\pm$0.15 \\  &  & 32.39952$\pm$0.29 & 33.08784$\pm$0.15 & 32.24374$\pm$0.17 & 31.61529$\pm$0.17 & 30.95200$\pm$0.19 & 30.48051$\pm$0.18 \\  \\[-1.72mm]
 & \textbf{SINE} & 87.22099$\pm$0.14 & 92.29431$\pm$0.08 & 92.22389$\pm$0.07 & \textbf{92.34442$\pm$0.07} & 92.11943$\pm$0.07 & 91.98204$\pm$0.06 \\  &  & 91.19445$\pm$0.14 & 91.73976$\pm$0.06 & 91.55407$\pm$0.07 & 91.29612$\pm$0.07 & 91.10594$\pm$0.07 & 90.86956$\pm$0.07 \\  \\[-1.72mm]
 & \textbf{WAVEF} & 80.15606$\pm$0.14 & 80.93873$\pm$0.14 & 82.14333$\pm$0.14 & 82.99579$\pm$0.12 & 83.41691$\pm$0.12 & 83.55055$\pm$0.12 \\  &  & 79.93768$\pm$0.16 & 83.77413$\pm$0.13 & 83.72505$\pm$0.14 & \textbf{83.89574$\pm$0.13} & 83.74374$\pm$0.13 & 83.87684$\pm$0.13 \\ 
\midrule 
 & \textbf{AGRAW$_{1}$} & 66.08369$\pm$0.08 & 65.20932$\pm$0.08 & 63.61951$\pm$0.08 & 63.81497$\pm$0.08 & 64.17847$\pm$0.08 & 65.75935$\pm$0.06 \\  &  & \textbf{74.85082$\pm$0.30} & 67.37946$\pm$0.06 & 68.42481$\pm$0.06 & 69.41076$\pm$0.05 & 69.39171$\pm$0.06 & 69.68465$\pm$0.06 \\  \\[-1.72mm]
 & \textbf{AGRAW$_{2}$} & 86.13011$\pm$0.05 & 81.14963$\pm$0.07 & 80.62725$\pm$0.05 & 80.85991$\pm$0.05 & 80.91333$\pm$0.05 & 81.87815$\pm$0.07 \\  &  & \textbf{87.16559$\pm$0.20} & 83.03014$\pm$0.06 & 84.28383$\pm$0.05 & 85.14383$\pm$0.06 & 85.65367$\pm$0.05 & 85.92654$\pm$0.05 \\  \\[-1.72mm]
 & \textbf{LED} & \textbf{73.39285$\pm$0.12} & 59.69480$\pm$0.14 & 63.53923$\pm$0.15 & 65.57214$\pm$0.14 & 66.55632$\pm$0.14 & 67.06381$\pm$0.15 \\  &  & 73.39090$\pm$0.12 & 67.50856$\pm$0.14 & 67.91412$\pm$0.14 & 68.20807$\pm$0.14 & 68.41363$\pm$0.14 & 68.51605$\pm$0.15 \\  \\[-1.72mm]
ABRUPT & \textbf{MIXED} & 91.78499$\pm$0.06 & \textbf{97.25106$\pm$0.03} & 96.79068$\pm$0.04 & 96.48841$\pm$0.04 & 96.12191$\pm$0.04 & 95.83576$\pm$0.04 \\ 100K &  & 92.45704$\pm$0.22 & 95.51403$\pm$0.04 & 95.21095$\pm$0.04 & 94.88088$\pm$0.04 & 94.57427$\pm$0.05 & 94.21022$\pm$0.04 \\  \\[-1.72mm]
 & \textbf{RandRBF} & 31.15903$\pm$0.29 & \textbf{36.96087$\pm$0.08} & 36.81254$\pm$0.09 & 36.17606$\pm$0.10 & 35.05425$\pm$0.09 & 34.11311$\pm$0.08 \\  &  & 32.80067$\pm$0.20 & 33.25313$\pm$0.09 & 32.44936$\pm$0.09 & 31.77364$\pm$0.11 & 31.12639$\pm$0.13 & 30.64200$\pm$0.12 \\  \\[-1.72mm]
 & \textbf{SINE} & 87.30729$\pm$0.11 & 92.48300$\pm$0.05 & 92.48532$\pm$0.04 & \textbf{92.62770$\pm$0.04} & 92.40867$\pm$0.05 & 92.30780$\pm$0.04 \\  &  & 92.40649$\pm$0.11 & 92.07009$\pm$0.04 & 91.88522$\pm$0.05 & 91.63315$\pm$0.05 & 91.43602$\pm$0.05 & 91.22003$\pm$0.05 \\  \\[-1.72mm]
 & \textbf{WAVEF} & 80.24963$\pm$0.11 & 80.98614$\pm$0.11 & 82.33706$\pm$0.09 & 83.16378$\pm$0.08 & 83.61458$\pm$0.08 & 83.76944$\pm$0.08 \\  &  & 79.96686$\pm$0.14 & 84.02079$\pm$0.08 & 83.95200$\pm$0.08 & 84.15581$\pm$0.08 & 84.02395$\pm$0.08 & \textbf{84.16632$\pm$0.07} \\ 
\bottomrule 
\end{tabular} 
\end{adjustbox} 
\end{table}

\begin{table}[t!]
\caption{Mean accuracies (\%) of classifiers with the RDDM detector, in abrupt datasets, with 95\% confidence intervals (Part~2)} 
\vspace{1mm}
\label{tab:det-acc-RDDM-Abr2} 
\begin{adjustbox}{max width=\textwidth} 
\begin{tabular}{cccccccc} 
\toprule 
DS Type & \textbf{RDDM} & NB & kNN5 & kNN10 & kNN15 & kNN20 & kNN25 \\ 
and Size & DATASET &  HT & kNN30 & kNN35 & kNN40 & kNN45 & kNN50 \\ 
\toprule 
 & \textbf{AGRAW$_{1}$} & 66.39112$\pm$0.07 & 65.46962$\pm$0.07 & 63.77711$\pm$0.07 & 63.92386$\pm$0.08 & 64.24169$\pm$0.09 & 65.88199$\pm$0.06 \\  &  & \textbf{78.60897$\pm$0.60} & 67.55721$\pm$0.05 & 68.62438$\pm$0.05 & 69.65001$\pm$0.04 & 69.60935$\pm$0.04 & 69.93041$\pm$0.04 \\  \\[-1.72mm]
 & \textbf{AGRAW$_{2}$} & 86.78076$\pm$0.07 & 81.73898$\pm$0.07 & 81.16466$\pm$0.07 & 81.35959$\pm$0.05 & 81.37199$\pm$0.07 & 82.32073$\pm$0.05 \\  &  & \textbf{88.79054$\pm$0.16} & 83.45874$\pm$0.06 & 84.73354$\pm$0.05 & 85.57388$\pm$0.05 & 86.08662$\pm$0.06 & 86.34783$\pm$0.06 \\  \\[-1.72mm]
 & \textbf{LED} & \textbf{73.74739$\pm$0.10} & 59.98041$\pm$0.12 & 63.87124$\pm$0.11 & 65.86407$\pm$0.12 & 66.87990$\pm$0.09 & 67.42750$\pm$0.12 \\  &  & 73.58075$\pm$0.10 & 67.94017$\pm$0.12 & 68.37546$\pm$0.11 & 68.69844$\pm$0.14 & 68.91406$\pm$0.14 & 69.04886$\pm$0.11 \\  \\[-1.72mm]
ABRUPT & \textbf{MIXED} & 92.00990$\pm$0.03 & \textbf{97.46634$\pm$0.03} & 97.02906$\pm$0.03 & 96.75130$\pm$0.03 & 96.40191$\pm$0.04 & 96.13914$\pm$0.04 \\ 500K &  & 94.22220$\pm$0.16 & 95.83534$\pm$0.04 & 95.53814$\pm$0.06 & 95.23082$\pm$0.05 & 94.96805$\pm$0.05 & 94.62935$\pm$0.05 \\  \\[-1.72mm]
 & \textbf{RandRBF} & 32.12894$\pm$0.30 & \textbf{36.93057$\pm$0.07} & 36.82781$\pm$0.09 & 36.22820$\pm$0.09 & 35.13631$\pm$0.10 & 34.16133$\pm$0.10 \\  &  & 33.73271$\pm$0.27 & 33.33361$\pm$0.08 & 32.56721$\pm$0.09 & 31.89815$\pm$0.09 & 31.22790$\pm$0.11 & 30.70736$\pm$0.12 \\  \\[-1.72mm]
 & \textbf{SINE} & 87.39919$\pm$0.07 & 92.69182$\pm$0.05 & 92.72034$\pm$0.04 & 92.86000$\pm$0.03 & 92.65892$\pm$0.04 & 92.55453$\pm$0.05 \\  &  & \textbf{94.88845$\pm$0.34} & 92.31158$\pm$0.04 & 92.14897$\pm$0.03 & 91.91482$\pm$0.03 & 91.73175$\pm$0.04 & 91.53567$\pm$0.03 \\  \\[-1.72mm]
 & \textbf{WAVEF} & 80.37370$\pm$0.13 & 81.04898$\pm$0.10 & 82.34828$\pm$0.12 & 83.19295$\pm$0.10 & 83.66582$\pm$0.10 & 83.80580$\pm$0.09 \\  &  & 80.17058$\pm$0.16 & 84.08563$\pm$0.08 & 84.05562$\pm$0.10 & 84.26746$\pm$0.06 & 84.15136$\pm$0.09 & \textbf{84.32668$\pm$0.07} \\ 
\midrule 
 & \textbf{AGRAW$_{1}$} & 66.49110$\pm$0.06 & 65.47772$\pm$0.06 & 63.76221$\pm$0.06 & 63.91539$\pm$0.07 & 64.23946$\pm$0.07 & 65.89862$\pm$0.07 \\  &  & \textbf{79.25155$\pm$1.25} & 67.58066$\pm$0.06 & 68.65980$\pm$0.06 & 69.68422$\pm$0.05 & 69.65325$\pm$0.05 & 69.95313$\pm$0.05 \\  \\[-1.72mm]
 & \textbf{AGRAW$_{2}$} & 86.86471$\pm$0.02 & 81.78537$\pm$0.03 & 81.20044$\pm$0.04 & 81.38513$\pm$0.04 & 81.40010$\pm$0.02 & 82.33852$\pm$0.03 \\  &  & \textbf{88.82691$\pm$0.30} & 83.48685$\pm$0.04 & 84.76882$\pm$0.04 & 85.60375$\pm$0.03 & 86.11869$\pm$0.03 & 86.38644$\pm$0.03 \\  \\[-1.72mm]
 & \textbf{LED} & \textbf{73.81737$\pm$0.07} & 60.02663$\pm$0.09 & 63.94975$\pm$0.09 & 65.94463$\pm$0.07 & 66.95385$\pm$0.06 & 67.51940$\pm$0.06 \\  &  & 73.62051$\pm$0.07 & 68.00124$\pm$0.06 & 68.43275$\pm$0.06 & 68.73545$\pm$0.09 & 68.95022$\pm$0.08 & 69.08168$\pm$0.06 \\  \\[-1.72mm]
ABRUPT & \textbf{MIXED} & 92.04268$\pm$0.04 & \textbf{97.48098$\pm$0.03} & 97.05465$\pm$0.03 & 96.79480$\pm$0.03 & 96.44180$\pm$0.03 & 96.18081$\pm$0.04 \\ 1M &  & 94.61614$\pm$0.27 & 95.87476$\pm$0.03 & 95.59789$\pm$0.04 & 95.28812$\pm$0.04 & 95.00628$\pm$0.03 & 94.67608$\pm$0.04 \\  \\[-1.72mm]
 & \textbf{RandRBF} & 32.16257$\pm$0.15 & \textbf{36.87590$\pm$0.05} & 36.78764$\pm$0.05 & 36.21073$\pm$0.05 & 35.10686$\pm$0.07 & 34.14139$\pm$0.06 \\  &  & 34.27013$\pm$0.33 & 33.30873$\pm$0.05 & 32.53405$\pm$0.06 & 31.90182$\pm$0.07 & 31.22605$\pm$0.07 & 30.71078$\pm$0.07 \\  \\[-1.72mm]
 & \textbf{SINE} & 87.44476$\pm$0.04 & 92.72212$\pm$0.02 & 92.75038$\pm$0.02 & 92.89699$\pm$0.03 & 92.67261$\pm$0.03 & 92.57623$\pm$0.03 \\  &  & \textbf{95.43600$\pm$0.22} & 92.33621$\pm$0.04 & 92.17468$\pm$0.04 & 91.93959$\pm$0.03 & 91.78204$\pm$0.05 & 91.56210$\pm$0.04 \\  \\[-1.72mm]
 & \textbf{WAVEF} & 80.41061$\pm$0.08 & 81.06376$\pm$0.08 & 82.40150$\pm$0.08 & 83.24971$\pm$0.08 & 83.71811$\pm$0.06 & 83.86625$\pm$0.07 \\  &  & 80.25358$\pm$0.14 & 84.13009$\pm$0.08 & 84.09875$\pm$0.07 & 84.30571$\pm$0.07 & 84.18932$\pm$0.09 & \textbf{84.36072$\pm$0.07} \\ 
\midrule 
 & \textbf{AGRAW$_{1}$} & 66.52017$\pm$0.04 & 65.49585$\pm$0.05 & 63.77541$\pm$0.04 & 63.91231$\pm$0.04 & 64.21571$\pm$0.05 & 65.91331$\pm$0.05 \\  &  & \textbf{79.34231$\pm$0.52} & 67.59820$\pm$0.04 & 68.69274$\pm$0.04 & 69.71833$\pm$0.04 & 69.69168$\pm$0.03 & 69.98437$\pm$0.04 \\  \\[-1.72mm]
 & \textbf{AGRAW$_{2}$} & 86.90362$\pm$0.02 & 81.81433$\pm$0.02 & 81.25080$\pm$0.02 & 81.42728$\pm$0.02 & 81.42790$\pm$0.02 & 82.35862$\pm$0.03 \\  &  & \textbf{89.07504$\pm$0.05} & 83.50878$\pm$0.03 & 84.79772$\pm$0.03 & 85.63396$\pm$0.02 & 86.14775$\pm$0.02 & 86.41993$\pm$0.03 \\  \\[-1.72mm]
 & \textbf{LED} & \textbf{73.87352$\pm$0.04} & 60.07145$\pm$0.04 & 63.98207$\pm$0.04 & 66.01640$\pm$0.06 & 66.98957$\pm$0.04 & 67.58627$\pm$0.06 \\  &  & 73.65803$\pm$0.05 & 68.07519$\pm$0.08 & 68.51200$\pm$0.08 & 68.82408$\pm$0.07 & 69.02414$\pm$0.09 & 69.18845$\pm$0.09 \\  \\[-1.72mm]
ABRUPT & \textbf{MIXED} & 92.01368$\pm$0.03 & \textbf{97.48189$\pm$0.02} & 97.06164$\pm$0.01 & 96.80543$\pm$0.02 & 96.45154$\pm$0.02 & 96.18294$\pm$0.03 \\ 2M &  & 94.53737$\pm$0.16 & 95.88780$\pm$0.02 & 95.59262$\pm$0.02 & 95.29578$\pm$0.02 & 95.00413$\pm$0.02 & 94.68558$\pm$0.03 \\  \\[-1.72mm]
 & \textbf{RandRBF} & 32.12830$\pm$0.15 & \textbf{36.88893$\pm$0.03} & 36.81797$\pm$0.03 & 36.23695$\pm$0.05 & 35.12659$\pm$0.06 & 34.16027$\pm$0.05 \\  &  & 34.26177$\pm$0.20 & 33.31917$\pm$0.04 & 32.52965$\pm$0.05 & 31.89809$\pm$0.05 & 31.23422$\pm$0.05 & 30.70892$\pm$0.05 \\  \\[-1.72mm]
 & \textbf{SINE} & 87.46986$\pm$0.04 & 92.73990$\pm$0.02 & 92.76533$\pm$0.01 & 92.91070$\pm$0.02 & 92.68469$\pm$0.01 & 92.58144$\pm$0.02 \\  &  & \textbf{95.67616$\pm$0.12} & 92.34294$\pm$0.02 & 92.19062$\pm$0.02 & 91.95929$\pm$0.01 & 91.78635$\pm$0.02 & 91.57562$\pm$0.02 \\  \\[-1.72mm]
 & \textbf{WAVEF} & 80.46233$\pm$0.04 & 81.08339$\pm$0.05 & 82.42630$\pm$0.05 & 83.26176$\pm$0.04 & 83.70134$\pm$0.05 & 83.84840$\pm$0.05 \\  &  & 80.36639$\pm$0.09 & 84.12662$\pm$0.05 & 84.09146$\pm$0.05 & 84.30634$\pm$0.05 & 84.19323$\pm$0.04 & \textbf{84.35618$\pm$0.04} \\ 
\bottomrule 
\end{tabular} 
\end{adjustbox} 
\end{table}

Hereafter, we focus on analyzing the impact of drift detection with different values of $k$.
First, we compare the accuracy rates in Tables \ref{tab:det-acc-NoDet-Abr1} and \ref{tab:det-acc-NoDet-Abr2} (no drift detector) to those of Tables \ref{tab:det-acc-RDDM-Abr1} and \ref{tab:det-acc-RDDM-Abr2} (with RDDM).
These results depict that \gls{knn} benefits from \gls{rddm} in most of the experiments.
More specifically, in average, accounting all stream lengths, \gls{knn} improves by 0.06\% in AGRAW$_1$, 0.74\% in AGRAW$_2$, 0.04\% in LED, 4.59\% in MIXED, and 3.50\% in SINE.
On the other hand, accuracy drops of 0.11\% and 0.06\% were observed in RandRBF and WAVEF, respectively.
Even though these results do not depict significant differences between the use or not of \gls{rddm}, we highlight the differences observed in the smaller experiments, i.e., those with 10, 20, 50, and 100 thousand instances datasets.
In such scenarios, RDDM improved the accuracy rates by 0.16\% in AGRAW$_1$, 1.30\% in AGRAW$_2$, 0.19\% in LED, 7.90\% in MIXED, and 6.04\% in SINE, while it impacted the accuracies negatively by 0.18\% in RandRBF and 0.05\% in WAVEF.

Nonetheless, these results also reveal that the choice of $k$ is more homogeneous, with positive and negative trends between $k$ and accuracy much clearer than before.
For instance, in AGRAW$_1$, AGRAW$_2$, LED, and WAVEF datasets, higher $k$ values returned higher accuracy rates, and $k=50$ achieved the best accuracy results in all stream lengths.
On the other hand, the results in MIXED, RandRBF, and SINE showed a negative trend between $k$ and accuracy: $k=5$ achieved the best accuracy rates, except for the 10 thousand instance scenario, where the trend between $k$ and accuracy is not entirely clear.

The results obtained in the gradual dataset experiments are introduced in Tables \ref{tab:det-acc-NoDet-Grad1} and \ref{tab:det-acc-NoDet-Grad2} provided in the appendix.
The behavior observed in these results shows that \gls{knn} benefits little from its association with \gls{rddm}.
More specifically, the accuracy gains are only observed in the AGRAW$_2$ (0.44\%), MIXED (2.51\%), and SINE (1.98\%) experiments.
Despite that, \gls{rddm} larger improvements were observed in the shorter dataset experiments, those with up to 100 thousand instances, with gains of 0.78\% in AGRAW$_2$, 4.34\% in MIXED, and 3.43\% in SINE.

Complementing the analysis of the results, different views of the accuracy results in Tables \ref{tab:det-acc-NoDet-Abr1} to \ref{tab:det-acc-RDDM-Abr2} and \ref{tab:det-acc-NoDet-Grad1} to \ref{tab:det-acc-RDDM-Grad2} were compared using the $F_F$ statistic \cite{demsar:2006}. 
The null hypothesis states that all methods are statistically equal. However, when rejected, it is necessary to use a posthoc test to discover in what method(s) there are statistical differences.
We used the Nemenyi posthoc test to compare each method against all the others.
The results are presented using graphics in which the critical difference (CD) is represented by bars and methods connected by a bar are statistically similar. 
Observe the graphics also include the values of the CD and the ranks of the methods.

\begin{figure}
\vspace{+3mm}
\begin{center}
\includegraphics[width=0.8\linewidth]{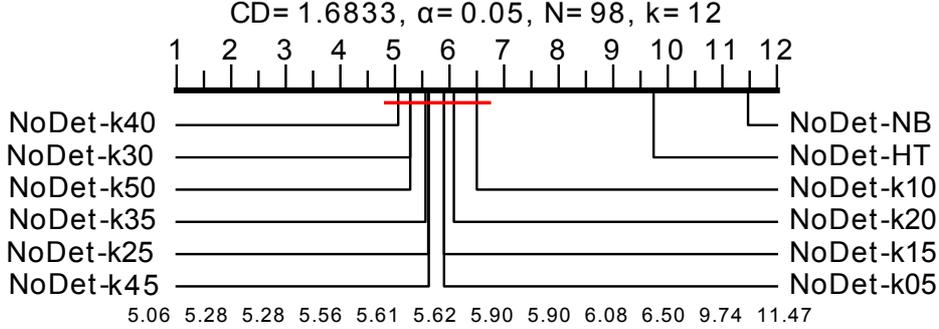}
\vspace{-10mm}
\caption{Comparison results using the Nemenyi test of classifiers without detector in all datasets with 95\% confidence.}\label{fig:Acc-NoDet-Tudo-Nem}
\end{center}
\vspace{-3mm}
\end{figure}

Figure~\ref{fig:Acc-NoDet-Tudo-Nem} evaluates the accuracy results of the classifiers aggregating {\em all} the tests executed without a concept drift detector, presented in Tables~\ref{tab:det-acc-NoDet-Abr1}, \ref{tab:det-acc-NoDet-Abr2}, \ref{tab:det-acc-NoDet-Grad1}, and \ref{tab:det-acc-NoDet-Grad2}, where the numbers after $k$ in the names of the methods is the number of neighbors. 
Even though the results of the experiments in the {\em abrupt} (Tables~\ref{tab:det-acc-NoDet-Abr1} and \ref{tab:det-acc-NoDet-Abr2}) and {\em gradual} datasets (Tables~\ref{tab:det-acc-NoDet-Grad1} and \ref{tab:det-acc-NoDet-Grad2}) were also evaluated separately, they are omitted because they are very much alike these of Figure~\ref{fig:Acc-NoDet-Tudo-Nem}.

According to the ranks, all the versions of \gls{knn} were significantly better than both \gls{ht} and \gls{nb}. 
Even though they were all statistically equivalent, versions with higher values of $k$ were generally ranked better than those with lower values of $k$, despite several exceptions. 
It is also worth pointing out that in this aggregated view, \gls{ht} was statistically superior to \gls{nb}, but this was {\em not} the case when the {\em abrupt} and {\em gradual} datasets were evaluated independently.

\begin{figure}
\begin{center}
\includegraphics[width=0.8\linewidth]{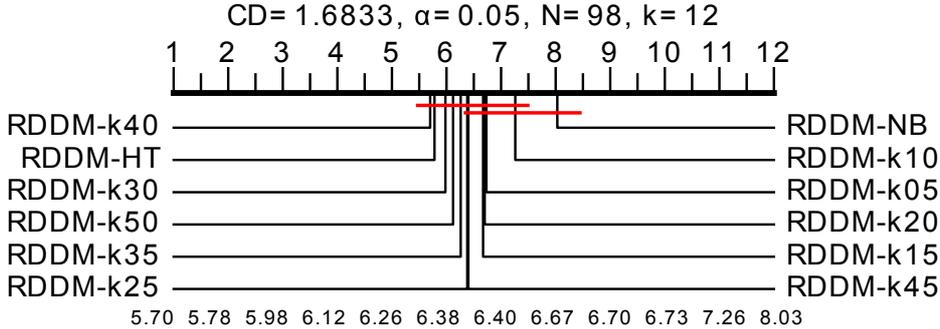}
\vspace{-10mm}
\caption{Comparison results using the Nemenyi test of classifiers with the RDDM detector in all datasets with 95\% confidence.}\label{fig:Acc-RDDM-Tudo-Nem}
\end{center}
\vspace{-3mm}
\end{figure}

Figure~\ref{fig:Acc-RDDM-Tudo-Nem} represents the evaluation of the accuracy results of the classifiers aggregating {\em all} the tests executed using \gls{rddm} as drift detector, which is similar to the aggregation carried out for the tests with no drift detector and represented in Figure~\ref{fig:Acc-NoDet-Tudo-Nem}. 
Again, {\em abrupt} (Tables~\ref{tab:det-acc-RDDM-Abr1} and \ref{tab:det-acc-RDDM-Abr2}) and {\em gradual} datasets (Tables~\ref{tab:det-acc-RDDM-Grad1} and \ref{tab:det-acc-RDDM-Grad2}) were also compared separately and their results were similar.

These results show \gls{knn} with higher values of $k$ and \gls{ht} in the top ranks whereas \gls{nb} and \gls{knn} with lower values of $k$ are ranked lower, but only \gls{nb} was statistically inferior to the best five ranked methods.
Nevertheless, it is essential to add that, in the {\em abrupt} datasets, there was no statistical difference at all and, in the {\em gradual} datasets, only \gls{knn} with $k=40$ was significantly better than \gls{nb}.

\subsection{Run-time analysis}

Changing the focus of the analysis to the run-time, we refer to Tables \ref{tab:det-time-NoDet-Abr1}, \ref{tab:det-time-NoDet-Abr2}, \ref{tab:det-time-NoDet-Grad1}, and \ref{tab:det-time-NoDet-Grad2} presented in the appendix, containing the results in the datasets with \textit{abrupt} (the first two) and \textit{gradual} (the last two) concept drifts. 
The results, in general, show that, as $k$ increases, there is also an increase in run-time in most cases.
This pattern can be seen, especially in larger datasets, e.g., 500K instances or more.

Despite this, the increases in run-time are not very significant, allowing larger $k$ values to be used without significant problems.
This information can be confirmed mainly by checking the time complexity of the linear \gls{knn} version used in this article (Subsection \ref{survey:knn}).
The value of $k$ has a very modest influence over time, especially when compared to other variables such as windows size ($w$) and the number of attributes ($d$).

Another more general observation that stands out is that, in some cases, the increase of $k$ in \gls{knn} causes the run-time to be reduced, mostly when such increase is small.
This fact can be explained by analyzing how the closest neighbors are defined.
As the distances are calculated, they are inserted in a \textit{heap} of size $k$. 
However, the insert operation (with logarithmic cost) will only be performed if the current distance is less than the largest distance already contained in the \textit{heap}, which can be accessed in constant time.
Due to this characteristic, \gls{knn} with a larger $k$ may perform fewer insertions than \gls{knn} with a smaller $k$, influencing its run-time.

From the perspective of time complexity, the asymptotic behavior of \gls{knn} -- $\mathcal{O}$($w \times (d + log$ $k$)) -- reflects, among other things, its upper bound.
Therefore, it is clear that small increases in $k$, even in this scenario, will have little effect on the run-time.
On the other hand, it is also important to note that significant increases in $k$ will most likely result in an increase in run-time since the cost of the insertion operation ($log$ $k$) will have a much higher weight.

Comparing the variation of $k$ in datasets with abrupt and gradual concept drifts, no significant changes were observed in the run-time in these scenarios.
In both, the performance ranking of the methods remains identical, demonstrating that the type of concept drift has no direct influence on the run-time.

\begin{figure}
\begin{center}
\includegraphics[width=0.8\linewidth]{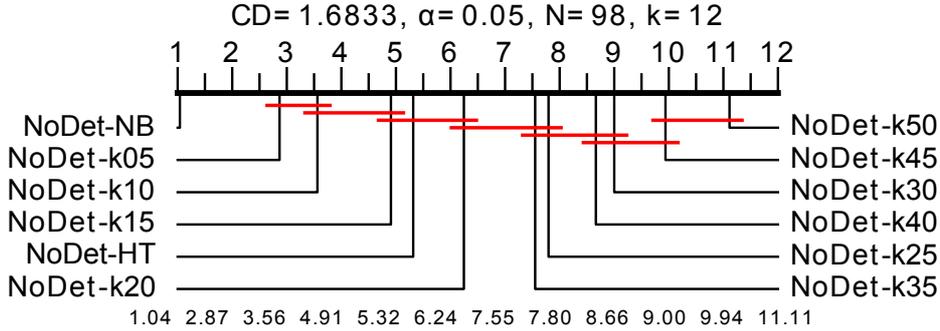}
\vspace{-10mm}
\caption{Comparison of run-times of classifiers with no detector in all datasets using the Nemenyi test with 95\% confidence.}\label{fig:Time-NoDet-Tudo-Nem}
\end{center}
\vspace{-3mm}
\end{figure}

Proceeding to the statistical evaluation, Figure \ref{fig:Time-NoDet-Tudo-Nem} summarizes the overall performance of the different versions of \gls{knn}. 
As expected, versions with $k$ = 50 and $k$ = 5 consumed the highest and lowest run-times, respectively, and versions with minor differences in the value of $k$ are usually statistically equivalent.
However, there were exceptions in the performance of some versions, i.e., the results did not follow the increase of $k$ strictly, reflecting the asymptotic variations between the best and the worst case in the different entries, as previously discussed.

Continuing the run-time analysis, Tables \ref{tab:det-time-RDDM-Abr1}, \ref{tab:det-time-RDDM-Abr2}, \ref{tab:det-time-RDDM-Grad1}, and \ref{tab:det-time-RDDM-Grad2} show the results of the experiments including a concept drift detector with the classifiers.
One of the characteristics that generally make the classifiers consume more run-time in such a scenario is related to a feature called warning state \cite{barros:2018a}.
When there are indications that a concept drift may have occurred, the detector emits a signal for an alternative classifier to be trained in parallel.
Subsequently, the main classifier will be removed and replaced by the alternative one if the change is confirmed.

Nevertheless, in the case of \gls{knn}, training a classifier in parallel does not affect the execution time as much as it affects other classifiers such as \gls{nb} and \gls{ht}.
As seen in Subsection \ref{survey:knn}, the \gls{knn} time complexity of training is constant, leaving most of the overhead for the classification period.
Thus, because the alternative classifier will {\em not} be tested, only trained, it causes little impact in the global run-time.

Despite the advantage of training a classifier in parallel, leaving all the processing to the classification period, as \gls{knn} does, also has disadvantages.
As can be seen in \cite{lu:2019}, one of the most common strategies to detect concept drifts is to analyze the errors of a classifier.
Therefore, using detectors with such characteristics, which is the case of \gls{rddm}, is likely to cause a significant increase in run-time since the classification might be carried out more frequently along the stream. 
This impact can be clearly seen by looking at the run-time tables without a detector (Tables \ref{tab:det-time-NoDet-Abr1}--\ref{tab:det-time-NoDet-Grad2}) and those with a detector (Tables \ref{tab:det-time-RDDM-Abr1}--\ref{tab:det-time-RDDM-Grad2}).

\begin{figure}
\vspace{+1mm}
\begin{center}
\includegraphics[width=0.8\linewidth]{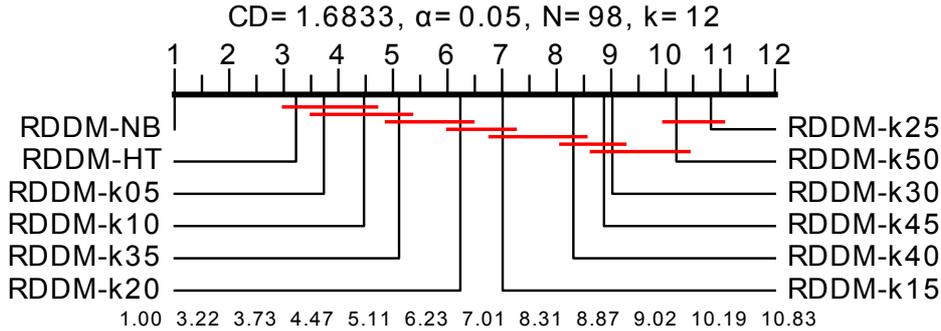}
\vspace{-10mm}
\caption{Comparison of run-times of classifiers with RDDM in all datasets using the Nemenyi test with 95\% confidence.}\label{fig:Time-RDDM-Tudo-Nem}
\end{center}
\vspace{-5mm}
\end{figure}

Figure \ref{fig:Time-RDDM-Tudo-Nem} captures the statistical evaluation of the general performance of the methods with a drift detector.
It is possible to note that, in many cases, increasing the $k$ also causes an increase in the run-time consumption, resulting in a higher classification cost.
However, it is necessary to point out that the main factor responsible for increasing run-time consumption with a detector is directly related to the classification operation, which tends to be performed much more often than without a detector.

\subsection{Answers to RQ1--RQ4}\label{rq1to4}

After analyzing the results of the first round of experiments, which tested ten different values for $k$, we now proceed to answer the first four research questions objectively.
The description of {\bfseries RQ1} was: {\em What is the impact of the neighborhood ($k$) on the accuracy of \gls{knn}?}

Based on the reported experiments, the answer to {\bfseries RQ1} is: 
the value of $k$ usually impacts the results of \gls{knn}, often moderately improving its results with larger neighborhoods. 
However, in some dataset generators, the lower values of $k$ deliver the highest accuracies.
This trend is mostly unaffected by the type of concept drift, with limited variation due to the length of the concepts in the datasets. 
When considering the complete set of datasets, the values 40, 30, and 50 of $k$ were the ones that achieved the best ranks, a conclusion which was clearly captured in Figure~\ref{fig:Acc-NoDet-Tudo-Nem}.

Proceeding to answer {\bfseries RQ2}, its description can be repeated: {\em Does the accuracy of \gls{knn} benefit from concept drift detection? If so, how much does it improve with different values of $k$?}

Based on the experiments, the answer to {\bfseries RQ2} is: \gls{knn} does benefit from its association with drift detection. 
Nonetheless, the addition of \gls{rddm} to the process yields an average increase of 1.25\% in abrupt and 0.64\% in gradual drifting streams.
Focusing on different values of $k$, the positive impact of \gls{rddm} depicts different trends in the abrupt and gradual datasets.
More specifically, in the abrupt experiments, the accuracy improvements decrease monotonically with the increase of $k$, as $k=5$ yields an average accuracy improvement of 1.26\% and $k=50$ achieved 1.07\%.
On the other hand, in the gradual drifts experiments, the best improvements in accuracy rates were achieved with $k=20$ (0.74\%), and smaller accuracy gains were observed with both the decrease and increase of $k$.

Now analyzing \textbf{RQ3}, its description was: {\em What is the impact of the neighborhood ($k$) on the run-time of \gls{knn}?}
The answer is: as expected, when the $k$ increases, the run-time also increases.
However, this overhead is not so significant and generally does not cause major problems for the classifier.
Besides, in some situations where the increase in $k$ is small, the run-time can even be reduced if the best case of the algorithm is accessed more often.

Finally, the description of \textbf{RQ4} was: {\em How does concept drift detection impact the run-time of \gls{knn} for different values of $k$?}
According to the reported experiments, the use of a drift detector together with \gls{knn} considerably increases the run-time since the classification operation will be performed much more frequently, mostly due to the warning periods.
Despite not having a strong influence on the time increase, using larger values for $k$ normally makes the cost of the classification operation higher.
Thus, it is sensible to attribute a small portion of this increase to using a larger value of $k$.

\section{Second Round of Experiments: varying $w$}\label{results:w}

This section analyses the accuracy and run-time of \gls{knn} considering changes in the window size $w$ and then answers \textbf{RQ5}--\textbf{RQ8}.
In these analyses, $w$ assumes values in the $\{600, 800, 1000, 1200, 1400, 1600\}$ set, and $k$ was set to 40 according to the best mean results of the previous round.
We point out that experiments with $k \in \{30, 50\}$ were also executed and their results are given in Tables \ref{tab:det-acc-NoDet-Abr30} to \ref{tab:det-time-RDDM-Grad50} in the appendix.

\subsection{Accuracy analysis} \label{sec:results:w:acc}

Tables \ref{tab:det-acc-NoDet-Abr40} and \ref{tab:det-acc-NoDet-Grad40} depict the accuracy rates obtained with different $w$ values in the experiments with abrupt and gradual datasets, respectively, without a drift detector.
Regarding the abrupt dataset experiments (Table \ref{tab:det-acc-NoDet-Abr40}), we note that, for experiments with up to 50 thousand instances, in AGRAW$_1$, AGRAW$_2$, MIXED, and SINE datasets, smaller values of $w$ usually result in higher accuracy rates across different stream lengths.
More specifically, we indicate that, in the 20K datasets, $w=600$ achieved the best accuracy rates in AGRAW$_2$ (82.8884\%), MIXED (87.4343\%), and SINE (85.6631\%), whereas $w=800$ achieved the best result in AGRAW$_1$ (68.2965\%).

\begin{table}[t!] 
\caption{Mean accuracies (\%) of classifiers with no drift detector, in abrupt datasets, with 95\% confidence intervals (k=40)} 
\vspace{1mm}
\label{tab:det-acc-NoDet-Abr40} 
\begin{adjustbox}{max width=\textwidth} 
\begin{tabular}{cccccccccc} 
\toprule 
DS Type  & \textbf{NO DET.} & \multicolumn{2}{c}{\textbf{CLASSIFIER}} & \multicolumn{6}{l}{kNN --- with k = 40, varying the size of the window (w)} \\ 
and Size & DATASET & NB & HT & w=600 & w=800 & w=1000 & w=1200 & w=1400 & w=1600 \\ 
\toprule 
 & \textbf{AGRAW$_{1}$} & 57.1202$\pm$0.18 & 58.7460$\pm$0.73 & 66.5146$\pm$0.15 & \textbf{66.8792$\pm$0.13} & 66.2104$\pm$0.15 & 65.0142$\pm$0.18 & 63.5866$\pm$0.17 & 62.3676$\pm$0.20 \\
 & \textbf{AGRAW$_{2}$} & 67.5743$\pm$0.30 & 69.9352$\pm$0.29 & \textbf{79.8761$\pm$0.20} & 78.9106$\pm$0.24 & 77.5573$\pm$0.24 & 75.9520$\pm$0.24 & 74.3497$\pm$0.28 & 72.9260$\pm$0.29 \\
ABRUPT &  \textbf{LED}  & 57.0178$\pm$0.25 & 55.8691$\pm$0.42 & 60.2659$\pm$0.36 & 61.4241$\pm$0.34 & \textbf{61.5251$\pm$0.37} & 61.2537$\pm$0.35 & 60.8386$\pm$0.32 & 60.3921$\pm$0.32 \\
10K  &  \textbf{MIXED}  & 57.6353$\pm$0.14 & 59.3308$\pm$0.40 & \textbf{81.3191$\pm$0.18} & 78.6476$\pm$0.18 & 75.4507$\pm$0.18 & 72.0042$\pm$0.17 & 68.4585$\pm$0.16 & 64.8057$\pm$0.15 \\
 &   \textbf{RandRBF}   & 30.8708$\pm$0.67 & 30.8309$\pm$0.67 & 28.5237$\pm$0.30 & 29.9102$\pm$0.31 & 30.9089$\pm$0.28 & 31.6503$\pm$0.33 & 32.2637$\pm$0.32 & \textbf{32.8696$\pm$0.36} \\
 &      \textbf{SINE}   & 56.3338$\pm$0.10 & 59.9403$\pm$0.49 & \textbf{80.7285$\pm$0.18} & 78.5016$\pm$0.19 & 75.8117$\pm$0.19 & 73.0607$\pm$0.20 & 70.1891$\pm$0.18 & 67.2150$\pm$0.18 \\
 &    \textbf{WAVEF}   & 76.5117$\pm$0.41 & 76.5511$\pm$0.39 & 81.8460$\pm$0.27 & 81.8656$\pm$0.30 & 81.8235$\pm$0.27 & 81.8781$\pm$0.29 & \textbf{81.9658$\pm$0.27} & 81.8991$\pm$0.29 \\
\midrule
 & \textbf{AGRAW$_{1}$} & 57.3969$\pm$0.10 & 58.2235$\pm$0.16 & 67.4417$\pm$0.12 & \textbf{68.2965$\pm$0.10} & 67.9724$\pm$0.10 & 66.9240$\pm$0.11 & 65.6328$\pm$0.13 & 64.3870$\pm$0.13 \\
 & \textbf{AGRAW$_{2}$} & 68.8835$\pm$0.19 & 71.1902$\pm$0.50 & \textbf{82.8884$\pm$0.09} & 82.5590$\pm$0.10 & 81.6992$\pm$0.11 & 80.5343$\pm$0.13 & 79.2715$\pm$0.14 & 78.2002$\pm$0.14 \\
ABRUPT &  \textbf{LED}  & 57.3712$\pm$0.18 & 58.8650$\pm$0.65 & 62.7336$\pm$0.26 & 64.4537$\pm$0.26 & 65.1981$\pm$0.29 & 65.4704$\pm$0.23 & 65.6067$\pm$0.20 & \textbf{65.6267$\pm$0.22} \\
20K  &  \textbf{MIXED}  & 57.9802$\pm$0.07 & 66.1857$\pm$0.41 & \textbf{87.4343$\pm$0.12} & 86.6792$\pm$0.11 & 85.4403$\pm$0.11 & 83.9489$\pm$0.09 & 82.3409$\pm$0.11 & 80.6850$\pm$0.12 \\
 &   \textbf{RandRBF}   & 31.2478$\pm$0.65 & 31.2278$\pm$0.65 & 28.7839$\pm$0.31 & 30.4493$\pm$0.26 & 31.5360$\pm$0.22 & 32.3641$\pm$0.22 & 33.0486$\pm$0.17 & \textbf{33.7019$\pm$0.20} \\
 &      \textbf{SINE}   & 56.5176$\pm$0.07 & 60.3873$\pm$0.91 & \textbf{85.6631$\pm$0.13} & 84.9721$\pm$0.13 & 83.9720$\pm$0.14 & 82.7985$\pm$0.14 & 81.4991$\pm$0.12 & 80.1634$\pm$0.12 \\
 &    \textbf{WAVEF}   & 76.7342$\pm$0.23 & 77.9984$\pm$0.24 & 82.8203$\pm$0.16 & 83.0046$\pm$0.17 & 83.1886$\pm$0.17 & 83.2646$\pm$0.17 & 83.3845$\pm$0.17 & \textbf{83.3900$\pm$0.16} \\
\midrule
 & \textbf{AGRAW$_{1}$} & 57.5697$\pm$0.09 & 57.1806$\pm$0.31 & 68.1026$\pm$0.10 & \textbf{69.1809$\pm$0.08} & 69.0490$\pm$0.08 & 68.1004$\pm$0.09 & 66.8269$\pm$0.09 & 65.6582$\pm$0.09 \\
 & \textbf{AGRAW$_{2}$} & 69.7630$\pm$0.10 & 70.6526$\pm$0.51 & \textbf{84.6563$\pm$0.07} & 84.6446$\pm$0.05 & 84.0842$\pm$0.07 & 83.1297$\pm$0.08 & 82.0740$\pm$0.09 & 81.3069$\pm$0.09 \\
ABRUPT &  \textbf{LED}  & 57.6013$\pm$0.12 & 63.2083$\pm$0.42 & 64.1928$\pm$0.19 & 66.2519$\pm$0.25 & 67.4054$\pm$0.21 & 67.9666$\pm$0.22 & 68.4388$\pm$0.19 & \textbf{68.7525$\pm$0.20} \\
50K  &  \textbf{MIXED}  & 58.2180$\pm$0.04 & 60.6734$\pm$0.41 & 91.1437$\pm$0.08 & \textbf{91.5325$\pm$0.08} & 91.4745$\pm$0.07 & 91.1929$\pm$0.06 & 90.7185$\pm$0.06 & 90.2035$\pm$0.06 \\
 &   \textbf{RandRBF}   & 31.3174$\pm$0.68 & 33.1677$\pm$0.33 & 28.9400$\pm$0.21 & 30.6071$\pm$0.20 & 31.7702$\pm$0.17 & 32.6868$\pm$0.16 & 33.4306$\pm$0.15 & \textbf{34.0899$\pm$0.15} \\
 &      \textbf{SINE}   & 56.6057$\pm$0.03 & 61.0417$\pm$0.82 & 88.5818$\pm$0.09 & \textbf{88.8511$\pm$0.07} & 88.8457$\pm$0.07 & 88.7021$\pm$0.06 & 88.3759$\pm$0.05 & 88.0122$\pm$0.06 \\
 &    \textbf{WAVEF}   & 76.8264$\pm$0.13 & 78.4042$\pm$0.21 & 83.4236$\pm$0.13 & 83.7224$\pm$0.14 & 83.9284$\pm$0.13 & 84.0307$\pm$0.13 & 84.1512$\pm$0.11 & \textbf{84.1810$\pm$0.11} \\
\midrule
 & \textbf{AGRAW$_{1}$} & 57.6128$\pm$0.06 & 58.5935$\pm$0.29 & 68.2899$\pm$0.06 & \textbf{69.4491$\pm$0.06} & 69.4000$\pm$0.05 & 68.4715$\pm$0.05 & 67.2387$\pm$0.06 & 66.0876$\pm$0.07 \\
 & \textbf{AGRAW$_{2}$} & 70.0807$\pm$0.07 & 71.0117$\pm$0.40 & 85.2159$\pm$0.06 & \textbf{85.3237$\pm$0.05} & 84.8731$\pm$0.05 & 83.9790$\pm$0.05 & 83.0220$\pm$0.06 & 82.3252$\pm$0.06 \\
ABRUPT &  \textbf{LED}  & 57.9025$\pm$0.08 & 64.9559$\pm$0.43 & 64.8478$\pm$0.14 & 66.9803$\pm$0.17 & 68.2015$\pm$0.15 & 68.9448$\pm$0.13 & 69.5100$\pm$0.13 & \textbf{69.9279$\pm$0.13} \\
100K &  \textbf{MIXED}  & 58.3264$\pm$0.02 & 69.7515$\pm$0.60 & 92.3188$\pm$0.05 & 93.1275$\pm$0.04 & 93.4550$\pm$0.03 & \textbf{93.5383$\pm$0.04} & 93.4730$\pm$0.03 & 93.3480$\pm$0.04 \\
 &   \textbf{RandRBF}   & 31.5146$\pm$0.61 & \textbf{34.8595$\pm$0.24} & 28.9971$\pm$0.16 & 30.6712$\pm$0.14 & 31.8719$\pm$0.10 & 32.8014$\pm$0.08 & 33.5342$\pm$0.10 & 34.1948$\pm$0.10 \\
 &      \textbf{SINE}   & 56.6659$\pm$0.03 & 66.0293$\pm$0.59 & 89.5238$\pm$0.06 & 90.1228$\pm$0.05 & 90.4382$\pm$0.04 & 90.6177$\pm$0.05 & \textbf{90.6539$\pm$0.04} & 90.6244$\pm$0.03 \\
 &    \textbf{WAVEF}   & 76.8903$\pm$0.09 & 79.6453$\pm$0.18 & 83.6321$\pm$0.08 & 83.9497$\pm$0.09 & 84.1986$\pm$0.08 & 84.3259$\pm$0.09 & 84.4375$\pm$0.08 & \textbf{84.4906$\pm$0.08} \\
\midrule
 & \textbf{AGRAW$_{1}$} & 57.6923$\pm$0.05 & 59.5325$\pm$0.63 & 68.5046$\pm$0.04 & \textbf{69.6976$\pm$0.04} & 69.6920$\pm$0.04 & 68.8048$\pm$0.05 & 67.5575$\pm$0.05 & 66.4036$\pm$0.05 \\
 & \textbf{AGRAW$_{2}$} & 70.4218$\pm$0.05 & 72.8439$\pm$0.49 & 85.7130$\pm$0.05 & \textbf{85.8886$\pm$0.04} & 85.5337$\pm$0.05 & 84.7105$\pm$0.05 & 83.7962$\pm$0.06 & 83.2020$\pm$0.06 \\
ABRUPT &  \textbf{LED}  & 58.1419$\pm$0.07 & 67.0068$\pm$0.48 & 65.2739$\pm$0.09 & 67.4905$\pm$0.13 & 68.8237$\pm$0.14 & 69.6349$\pm$0.15 & 70.2244$\pm$0.14 & \textbf{70.7214$\pm$0.12} \\
500K &  \textbf{MIXED}  & 58.4060$\pm$0.01 & 78.2613$\pm$0.50 & 93.2834$\pm$0.05 & 94.3992$\pm$0.05 & 95.0267$\pm$0.05 & 95.4170$\pm$0.05 & 95.6880$\pm$0.04 & \textbf{95.8638$\pm$0.05} \\
 &   \textbf{RandRBF}   & 33.2619$\pm$0.76 & \textbf{38.1924$\pm$0.11} & 28.9076$\pm$0.11 & 30.6917$\pm$0.12 & 31.9779$\pm$0.10 & 32.9257$\pm$0.07 & 33.6545$\pm$0.07 & 34.2843$\pm$0.10 \\
 &      \textbf{SINE}   & 56.6671$\pm$0.03 & 76.2881$\pm$0.46 & 90.3044$\pm$0.06 & 91.1569$\pm$0.04 & 91.7310$\pm$0.03 & 92.1387$\pm$0.04 & 92.4502$\pm$0.03 & \textbf{92.7019$\pm$0.03} \\
 &    \textbf{WAVEF}   & 76.8496$\pm$0.11 & 81.5784$\pm$0.10 & 83.7591$\pm$0.07 & 84.0612$\pm$0.08 & 84.3373$\pm$0.06 & 84.4853$\pm$0.09 & 84.5703$\pm$0.07 & \textbf{84.6628$\pm$0.09} \\
\midrule
 & \textbf{AGRAW$_{1}$} & 57.6954$\pm$0.03 & 60.9947$\pm$0.36 & 68.5080$\pm$0.05 & 69.7197$\pm$0.04 & \textbf{69.7403$\pm$0.05} & 68.8492$\pm$0.05 & 67.5785$\pm$0.05 & 66.4352$\pm$0.04 \\
 & \textbf{AGRAW$_{2}$} & 70.4358$\pm$0.03 & 73.5730$\pm$0.48 & 85.7632$\pm$0.03 & \textbf{85.9548$\pm$0.03} & 85.5999$\pm$0.03 & 84.7803$\pm$0.03 & 83.8747$\pm$0.03 & 83.2871$\pm$0.04 \\
ABRUPT &  \textbf{LED}  & 58.2422$\pm$0.05 & 67.1576$\pm$0.37 & 65.3357$\pm$0.07 & 67.5638$\pm$0.08 & 68.8896$\pm$0.08 & 69.7557$\pm$0.08 & 70.3544$\pm$0.07 & \textbf{70.8686$\pm$0.07} \\
1M  &   \textbf{MIXED}  & 58.4220$\pm$0.01 & 82.2567$\pm$0.43 & 93.4202$\pm$0.05 & 94.5671$\pm$0.05 & 95.2414$\pm$0.04 & 95.6725$\pm$0.04 & 95.9791$\pm$0.04 & \textbf{96.1934$\pm$0.04} \\
 &   \textbf{RandRBF}   & 33.1638$\pm$0.50 & \textbf{38.7590$\pm$0.09} & 28.8861$\pm$0.08 & 30.6777$\pm$0.07 & 31.9630$\pm$0.06 & 32.9056$\pm$0.04 & 33.6294$\pm$0.05 & 34.2661$\pm$0.06 \\
 &      \textbf{SINE}   & 56.6799$\pm$0.01 & 80.4546$\pm$0.47 & 90.4104$\pm$0.04 & 91.2982$\pm$0.04 & 91.8893$\pm$0.03 & 92.3315$\pm$0.03 & 92.6864$\pm$0.03 & \textbf{92.9632$\pm$0.03} \\
 &    \textbf{WAVEF}   & 76.8993$\pm$0.07 & 82.3291$\pm$0.13 & 83.7926$\pm$0.07 & 84.1202$\pm$0.07 & 84.3726$\pm$0.07 & 84.5284$\pm$0.08 & 84.6273$\pm$0.07 & \textbf{84.7106$\pm$0.07} \\
\midrule
 & \textbf{AGRAW$_{1}$} & 57.6997$\pm$0.02 & 62.4868$\pm$0.18 & 68.5253$\pm$0.03 & 69.7525$\pm$0.03 & \textbf{69.7733$\pm$0.04} & 68.8863$\pm$0.04 & 67.6216$\pm$0.04 & 66.4654$\pm$0.04 \\
 & \textbf{AGRAW$_{2}$} & 70.4526$\pm$0.04 & 73.7871$\pm$0.27 & 85.7929$\pm$0.02 & \textbf{85.9977$\pm$0.02} & 85.6439$\pm$0.02 & 84.8267$\pm$0.02 & 83.9199$\pm$0.02 & 83.3335$\pm$0.03 \\
ABRUPT &  \textbf{LED}  & 58.2921$\pm$0.02 & 66.7013$\pm$0.27 & 65.4039$\pm$0.05 & 67.6228$\pm$0.06 & 68.9730$\pm$0.08 & 69.8054$\pm$0.05 & 70.4247$\pm$0.06 & \textbf{70.9541$\pm$0.05} \\
2M  &   \textbf{MIXED}  & 58.4132$\pm$0.01 & 85.8136$\pm$0.20 & 93.4632$\pm$0.02 & 94.6292$\pm$0.02 & 95.3199$\pm$0.02 & 95.7731$\pm$0.02 & 96.1015$\pm$0.02 & \textbf{96.3400$\pm$0.02} \\
 &   \textbf{RandRBF}   & 33.0687$\pm$0.22 & \textbf{39.4791$\pm$0.07} & 28.9163$\pm$0.03 & 30.6881$\pm$0.05 & 31.9586$\pm$0.05 & 32.8977$\pm$0.05 & 33.6339$\pm$0.04 & 34.2722$\pm$0.05 \\
 &      \textbf{SINE}   & 56.6855$\pm$0.01 & 84.1871$\pm$0.55 & 90.4566$\pm$0.03 & 91.3537$\pm$0.02 & 91.9684$\pm$0.02 & 92.4264$\pm$0.01 & 92.7908$\pm$0.02 & \textbf{93.0810$\pm$0.01} \\
 &    \textbf{WAVEF}   & 76.9261$\pm$0.03 & 82.9069$\pm$0.07 & 83.7852$\pm$0.04 & 84.1323$\pm$0.05 & 84.3704$\pm$0.04 & 84.5158$\pm$0.05 & 84.6226$\pm$0.05 & \textbf{84.7132$\pm$0.05} \\
\bottomrule 
\end{tabular} 
\end{adjustbox} 
\end{table}

On the other hand, in the LED, RandRBF, and WAVEF experiments, higher values of $w$ normally enhance the results in all dataset sizes. The difference between $w \in \{1200, 1400, 1600\}$ often falls within the confidence interval and, except for the 10K datasets, the best configuration for these three generators was always $w=1600$.

Next, focusing on the experiments using datasets with greater lengths, i.e., those with 100 thousand instances or more, we point out that larger window sizes also tend to induce better accuracy results in the MIXED and SINE datasets, in addition to the LED, RandRBF, and WAVEF datasets, again with different values of $w$ frequently falling within the confidence interval.
More specifically, in the datasets configured with 500 thousand instances or more, the $w=1600$ configuration reached the very best accuracy rates in all these generators.

At this point, it is also worthy highlighting the volatility of the results obtained with different values of $w$ in each generator, i.e., the difference between the results obtained with the best and worst-performing $w$ values.
In these experiments with abrupt datasets, we observed that the volatility was more significant in RandRBF (5.10\%), MIXED (4.86\%), and LED (4.33\%), followed by SINE (4.08\%), AGRAW$_2$ (3.71\%), AGRAW$_1$ (3.60\%), and WAVEF (0.72\%).
These results show that the window size severely impacts \gls{knn}'s accuracy, and thus, a tuning process would be advised. However, it is not trivial in data stream scenarios, where no assumptions about the window size can be made \textit{a priori}.

The results obtained with different $w$ values in the gradual datasets, given in Table \ref{tab:det-acc-NoDet-Grad40}, depict a slightly different behavior when compared to those of the abrupt experiments.
Specifically, in the AGRAW$_1$ and AGRAW$_2$ datasets, the optimum window size observed was $w=800$ with mean accuracy rates of 68.93\% and 84.01\%, respectively, in contrast to $w=600$ in the abrupt experiments. 
A similar increase also happened in the MIXED and SINE datasets. 
Nevertheless, $w=1600$ again yielded the best accuracy rates in the larger datasets, except for AGRAW$_1$ and AGRAW$_2$.
The mean accuracy observed with $w=1600$ was 68.15\% in LED, 88.18\% in MIXED, 33.96\% in RandRBF, 83.37\% in SINE, and 84.01\% in WAVEF.
Furthermore, we observe that the results obtained with $w \geq 1200$ also commonly fall within the confidence interval.
Finally, the volatility results observed in the gradual experiments were similar to those of the abrupt streams, i.e., the differences were also larger in RandRBF (5.12\%), MIXED (4.66\%), LED (4.41\%), SINE (3.89\%), AGRAW$_2$ (3.56\%), AGRAW$_1$ (3.55\%), and WAVEF (0.73\%).

\begin{table}[h!] 
\caption{Mean accuracies (\%) of classifiers with no drift detector, in gradual datasets, with 95\% confidence intervals (k=40)} 
\vspace{1mm}
\label{tab:det-acc-NoDet-Grad40} 
\begin{adjustbox}{max width=\textwidth} 
\begin{tabular}{cccccccccc} 
\toprule 
DS Type  & \textbf{NO DET.} & \multicolumn{2}{c}{\textbf{CLASSIFIER}} & \multicolumn{6}{l}{kNN --- with k = 40, varying the size of the window (w)} \\ 
and Size & DATASET & NB & HT & w=600 & w=800 & w=1000 & w=1200 & w=1400 & w=1600 \\ 
\toprule 
 & \textbf{AGRAW$_{1}$} & 57.1569$\pm$0.19 & 58.6905$\pm$0.71 & 66.0978$\pm$0.13 & \textbf{66.6016$\pm$0.16} & 66.0814$\pm$0.16 & 64.8984$\pm$0.18 & 63.4974$\pm$0.18 & 62.2898$\pm$0.19 \\
 & \textbf{AGRAW$_{2}$} & 67.5442$\pm$0.28 & 69.4492$\pm$0.43 & \textbf{78.9693$\pm$0.19} & 78.2501$\pm$0.21 & 77.0588$\pm$0.24 & 75.5371$\pm$0.22 & 74.0051$\pm$0.26 & 72.6835$\pm$0.28 \\
GRAD. &   \textbf{LED}  & 57.0161$\pm$0.25 & 55.8402$\pm$0.41 & 59.8822$\pm$0.37 & 61.0943$\pm$0.31 & \textbf{61.3595$\pm$0.37} & 61.1227$\pm$0.34 & 60.7657$\pm$0.31 & 60.2972$\pm$0.32 \\
10K  &  \textbf{MIXED}  & 57.6343$\pm$0.12 & 59.4987$\pm$0.38 & \textbf{80.3050$\pm$0.17} & 78.1008$\pm$0.17 & 75.1357$\pm$0.18 & 71.8357$\pm$0.17 & 68.3453$\pm$0.15 & 64.7379$\pm$0.15 \\
 &   \textbf{RandRBF}   & 30.9061$\pm$0.68 & 30.8662$\pm$0.68 & 28.4359$\pm$0.32 & 29.8386$\pm$0.32 & 30.8588$\pm$0.29 & 31.5995$\pm$0.31 & 32.2805$\pm$0.34 & \textbf{32.8786$\pm$0.35} \\
 &      \textbf{SINE}   & 56.3515$\pm$0.10 & 59.2024$\pm$0.28 & \textbf{79.6857$\pm$0.17} & 77.8549$\pm$0.19 & 75.4714$\pm$0.20 & 72.8101$\pm$0.18 & 70.0557$\pm$0.18 & 67.1004$\pm$0.19 \\
 &    \textbf{WAVEF}   & 76.4807$\pm$0.41 & 76.5205$\pm$0.38 & 81.7829$\pm$0.25 & 81.8770$\pm$0.28 & 81.9253$\pm$0.28 & 81.8975$\pm$0.27 & \textbf{81.9366$\pm$0.27} & 81.9324$\pm$0.27 \\
\midrule
 & \textbf{AGRAW$_{1}$} & 57.4003$\pm$0.10 & 58.1591$\pm$0.17 & 67.3005$\pm$0.11 & \textbf{68.1797$\pm$0.10} & 67.9019$\pm$0.11 & 66.8971$\pm$0.11 & 65.6182$\pm$0.12 & 64.3821$\pm$0.12 \\
 & \textbf{AGRAW$_{2}$} & 68.8390$\pm$0.19 & 70.8517$\pm$0.52 & \textbf{82.4641$\pm$0.09} & 82.2535$\pm$0.10 & 81.4614$\pm$0.10 & 80.3116$\pm$0.13 & 79.1276$\pm$0.14 & 78.0539$\pm$0.14 \\
GRAD. &   \textbf{LED}  & 57.4441$\pm$0.18 & 58.8483$\pm$0.71 & 62.4867$\pm$0.25 & 64.2620$\pm$0.27 & 65.0942$\pm$0.28 & 65.3828$\pm$0.22 & 65.5580$\pm$0.20 & \textbf{65.5898$\pm$0.20} \\
20K  &  \textbf{MIXED}  & 58.1845$\pm$0.06 & 65.0502$\pm$0.47 & \textbf{86.9392$\pm$0.12} & 86.4248$\pm$0.11 & 85.3030$\pm$0.10 & 83.8612$\pm$0.09 & 82.3028$\pm$0.10 & 80.6462$\pm$0.12 \\
 &   \textbf{RandRBF}   & 31.2656$\pm$0.64 & 31.2457$\pm$0.64 & 28.7431$\pm$0.29 & 30.3579$\pm$0.27 & 31.5094$\pm$0.21 & 32.3343$\pm$0.20 & 33.0189$\pm$0.17 & \textbf{33.7029$\pm$0.19} \\
 &      \textbf{SINE}   & 56.6529$\pm$0.07 & 59.7488$\pm$0.63 & \textbf{85.1510$\pm$0.13} & 84.6850$\pm$0.13 & 83.7775$\pm$0.14 & 82.6758$\pm$0.13 & 81.4201$\pm$0.11 & 80.1273$\pm$0.11 \\
 &    \textbf{WAVEF}   & 76.7177$\pm$0.23 & 77.9071$\pm$0.24 & 82.8113$\pm$0.16 & 83.0098$\pm$0.19 & 83.1972$\pm$0.18 & 83.2794$\pm$0.17 & 83.3600$\pm$0.17 & \textbf{83.4021$\pm$0.15} \\
\midrule
 & \textbf{AGRAW$_{1}$} & 57.5690$\pm$0.09 & 57.1829$\pm$0.30 & 68.0434$\pm$0.10 & \textbf{69.1545$\pm$0.08} & 69.0264$\pm$0.08 & 68.0757$\pm$0.10 & 66.8159$\pm$0.09 & 65.6511$\pm$0.09 \\
 & \textbf{AGRAW$_{2}$} & 69.7512$\pm$0.10 & 70.5800$\pm$0.49 & 84.5004$\pm$0.07 & \textbf{84.5279$\pm$0.05} & 84.0017$\pm$0.07 & 83.0628$\pm$0.08 & 82.0236$\pm$0.09 & 81.2592$\pm$0.09 \\
GRAD. &   \textbf{LED}  & 57.6150$\pm$0.12 & 63.4192$\pm$0.36 & 64.1174$\pm$0.19 & 66.1922$\pm$0.25 & 67.3566$\pm$0.21 & 67.9523$\pm$0.22 & 68.4069$\pm$0.19 & \textbf{68.7273$\pm$0.20} \\
50K  &  \textbf{MIXED}  & 58.2482$\pm$0.04 & 60.5272$\pm$0.40 & 90.9410$\pm$0.08 & \textbf{91.4210$\pm$0.08} & 91.4136$\pm$0.07 & 91.1526$\pm$0.05 & 90.6990$\pm$0.05 & 90.1910$\pm$0.06 \\
 &   \textbf{RandRBF}   & 31.3117$\pm$0.68 & 33.1653$\pm$0.33 & 28.9257$\pm$0.21 & 30.5928$\pm$0.20 & 31.7654$\pm$0.17 & 32.6689$\pm$0.16 & 33.4197$\pm$0.15 & \textbf{34.0898$\pm$0.15} \\
 &      \textbf{SINE}   & 56.6173$\pm$0.03 & 60.7900$\pm$0.76 & 88.3829$\pm$0.09 & 88.7385$\pm$0.07 & \textbf{88.7757$\pm$0.07} & 88.6515$\pm$0.06 & 88.3446$\pm$0.05 & 87.9905$\pm$0.06 \\
 &    \textbf{WAVEF}   & 76.8259$\pm$0.13 & 78.4224$\pm$0.21 & 83.4216$\pm$0.13 & 83.7253$\pm$0.14 & 83.9257$\pm$0.13 & 84.0451$\pm$0.13 & 84.1594$\pm$0.11 & \textbf{84.1889$\pm$0.11} \\
\midrule
 & \textbf{AGRAW$_{1}$} & 57.6107$\pm$0.06 & 58.5971$\pm$0.28 & 68.2629$\pm$0.06 & \textbf{69.4351$\pm$0.05} & 69.3851$\pm$0.05 & 68.4680$\pm$0.05 & 67.2352$\pm$0.06 & 66.0818$\pm$0.07 \\
 & \textbf{AGRAW$_{2}$} & 70.0836$\pm$0.07 & 70.9982$\pm$0.39 & 85.1378$\pm$0.06 & \textbf{85.2661$\pm$0.05} & 84.8339$\pm$0.05 & 83.9487$\pm$0.05 & 82.9934$\pm$0.06 & 82.3088$\pm$0.06 \\
GRAD. &   \textbf{LED}  & 57.8975$\pm$0.08 & 64.8055$\pm$0.46 & 64.8099$\pm$0.13 & 66.9419$\pm$0.17 & 68.1863$\pm$0.15 & 68.9369$\pm$0.13 & 69.5016$\pm$0.13 & \textbf{69.9246$\pm$0.13} \\
100K &  \textbf{MIXED}  & 58.3175$\pm$0.02 & 65.1336$\pm$0.97 & 92.2303$\pm$0.05 & 93.0766$\pm$0.04 & 93.4294$\pm$0.03 & \textbf{93.5202$\pm$0.04} & 93.4619$\pm$0.04 & 93.3396$\pm$0.04 \\
 &   \textbf{RandRBF}   & 31.5258$\pm$0.60 & \textbf{34.8533$\pm$0.20} & 28.9873$\pm$0.16 & 30.6627$\pm$0.14 & 31.8673$\pm$0.10 & 32.7947$\pm$0.08 & 33.5350$\pm$0.10 & 34.1963$\pm$0.10 \\
 &      \textbf{SINE}   & 56.6591$\pm$0.03 & 65.3034$\pm$0.54 & 89.4307$\pm$0.06 & 90.0722$\pm$0.05 & 90.4109$\pm$0.05 & 90.6017$\pm$0.05 & \textbf{90.6406$\pm$0.04} & 90.6149$\pm$0.03 \\
 &    \textbf{WAVEF}   & 76.8900$\pm$0.09 & 79.6396$\pm$0.18 & 83.6285$\pm$0.08 & 83.9522$\pm$0.09 & 84.1935$\pm$0.08 & 84.3289$\pm$0.08 & 84.4367$\pm$0.08 & \textbf{84.4935$\pm$0.08} \\
\midrule
 & \textbf{AGRAW$_{1}$} & 57.6923$\pm$0.05 & 59.5312$\pm$0.63 & 68.4974$\pm$0.03 & \textbf{69.6941$\pm$0.04} & 69.6902$\pm$0.04 & 68.8028$\pm$0.05 & 67.5572$\pm$0.05 & 66.4038$\pm$0.05 \\
 & \textbf{AGRAW$_{2}$} & 70.4198$\pm$0.05 & 72.8476$\pm$0.49 & 85.6956$\pm$0.05 & \textbf{85.8759$\pm$0.04} & 85.5247$\pm$0.05 & 84.7014$\pm$0.05 & 83.7892$\pm$0.06 & 83.1977$\pm$0.06 \\
GRAD. &   \textbf{LED}  & 58.1423$\pm$0.07 & 67.1004$\pm$0.48 & 65.2675$\pm$0.09 & 67.4850$\pm$0.13 & 68.8202$\pm$0.14 & 69.6335$\pm$0.15 & 70.2218$\pm$0.14 & \textbf{70.7204$\pm$0.12} \\
500K &  \textbf{MIXED}  & 58.4046$\pm$0.01 & 76.4938$\pm$0.69 & 93.2667$\pm$0.05 & 94.3913$\pm$0.05 & 95.0249$\pm$0.05 & 95.4152$\pm$0.05 & 95.6872$\pm$0.04 & \textbf{95.8629$\pm$0.05} \\
 &   \textbf{RandRBF}   & 33.2553$\pm$0.76 & \textbf{38.1916$\pm$0.10} & 28.9056$\pm$0.11 & 30.6906$\pm$0.13 & 31.9796$\pm$0.10 & 32.9252$\pm$0.07 & 33.6545$\pm$0.07 & 34.2840$\pm$0.10 \\
 &      \textbf{SINE}   & 56.6644$\pm$0.03 & 75.3324$\pm$0.49 & 90.2878$\pm$0.06 & 91.1461$\pm$0.04 & 91.7250$\pm$0.03 & 92.1354$\pm$0.04 & 92.4478$\pm$0.03 & \textbf{92.7019$\pm$0.03} \\
 &    \textbf{WAVEF}   & 76.8494$\pm$0.11 & 81.5784$\pm$0.10 & 83.7580$\pm$0.07 & 84.0613$\pm$0.08 & 84.3384$\pm$0.07 & 84.4864$\pm$0.09 & 84.5699$\pm$0.07 & \textbf{84.6635$\pm$0.09} \\
\midrule
 & \textbf{AGRAW$_{1}$} & 57.6955$\pm$0.03 & 60.9908$\pm$0.36 & 68.5047$\pm$0.05 & 69.7179$\pm$0.04 & \textbf{69.7397$\pm$0.05} & 68.8479$\pm$0.05 & 67.5782$\pm$0.05 & 66.4347$\pm$0.04 \\
 & \textbf{AGRAW$_{2}$} & 70.4358$\pm$0.03 & 73.5680$\pm$0.48 & 85.7551$\pm$0.03 & \textbf{85.9482$\pm$0.03} & 85.5950$\pm$0.03 & 84.7762$\pm$0.03 & 83.8716$\pm$0.03 & 83.2849$\pm$0.04 \\
GRAD. &   \textbf{LED}  & 58.2426$\pm$0.05 & 67.1881$\pm$0.40 & 65.3296$\pm$0.07 & 67.5613$\pm$0.08 & 68.8896$\pm$0.08 & 69.7555$\pm$0.08 & 70.3531$\pm$0.07 & \textbf{70.8680$\pm$0.07} \\
1M  &   \textbf{MIXED}  & 58.4247$\pm$0.01 & 81.1282$\pm$0.50 & 93.4093$\pm$0.05 & 94.5619$\pm$0.05 & 95.2385$\pm$0.04 & 95.6710$\pm$0.04 & 95.9777$\pm$0.04 & \textbf{96.1925$\pm$0.04} \\
 &   \textbf{RandRBF}   & 33.1624$\pm$0.51 & \textbf{38.7585$\pm$0.09} & 28.8861$\pm$0.08 & 30.6771$\pm$0.07 & 31.9630$\pm$0.06 & 32.9052$\pm$0.04 & 33.6296$\pm$0.05 & 34.2665$\pm$0.06 \\
 &      \textbf{SINE}   & 56.6816$\pm$0.01 & 79.9245$\pm$0.44 & 90.4005$\pm$0.04 & 91.2922$\pm$0.04 & 91.8853$\pm$0.03 & 92.3293$\pm$0.03 & 92.6844$\pm$0.03 & \textbf{92.9618$\pm$0.03} \\
 &    \textbf{WAVEF}   & 76.8992$\pm$0.07 & 82.3262$\pm$0.14 & 83.7924$\pm$0.07 & 84.1202$\pm$0.07 & 84.3737$\pm$0.07 & 84.5290$\pm$0.08 & 84.6270$\pm$0.07 & \textbf{84.7113$\pm$0.07} \\
\midrule
 & \textbf{AGRAW$_{1}$} & 57.6996$\pm$0.02 & 62.4834$\pm$0.18 & 68.5239$\pm$0.03 & 69.7512$\pm$0.03 & \textbf{69.7732$\pm$0.04} & 68.8861$\pm$0.04 & 67.6213$\pm$0.04 & 66.4655$\pm$0.04 \\
 & \textbf{AGRAW$_{2}$} & 70.4528$\pm$0.04 & 73.7906$\pm$0.27 & 85.7884$\pm$0.02 & \textbf{85.9946$\pm$0.02} & 85.6417$\pm$0.02 & 84.8250$\pm$0.02 & 83.9190$\pm$0.02 & 83.3328$\pm$0.03 \\
GRAD. &   \textbf{LED}  & 58.2927$\pm$0.02 & 66.7024$\pm$0.28 & 65.4010$\pm$0.05 & 67.6205$\pm$0.06 & 68.9715$\pm$0.08 & 69.8046$\pm$0.05 & 70.4238$\pm$0.06 & \textbf{70.9536$\pm$0.05} \\
2M  &   \textbf{MIXED}  & 58.4159$\pm$0.01 & 84.6890$\pm$0.51 & 93.4579$\pm$0.02 & 94.6264$\pm$0.02 & 95.3183$\pm$0.02 & 95.7721$\pm$0.02 & 96.1009$\pm$0.02 & \textbf{96.3400$\pm$0.02} \\
 &   \textbf{RandRBF}   & 33.0691$\pm$0.22 & \textbf{39.4803$\pm$0.07} & 28.9159$\pm$0.03 & 30.6879$\pm$0.05 & 31.9586$\pm$0.05 & 32.8981$\pm$0.05 & 33.6345$\pm$0.04 & 34.2728$\pm$0.05 \\
 &      \textbf{SINE}   & 56.6870$\pm$0.01 & 83.5899$\pm$0.58 & 90.4511$\pm$0.03 & 91.3503$\pm$0.02 & 91.9666$\pm$0.02 & 92.4253$\pm$0.01 & 92.7901$\pm$0.02 & \textbf{93.0809$\pm$0.01} \\
 &    \textbf{WAVEF}   & 76.9261$\pm$0.03 & 82.9082$\pm$0.07 & 83.7849$\pm$0.04 & 84.1320$\pm$0.05 & 84.3703$\pm$0.04 & 84.5159$\pm$0.05 & 84.6219$\pm$0.05 & \textbf{84.7138$\pm$0.05} \\
\bottomrule 
\end{tabular} 
\end{adjustbox} 
\end{table}

Tables \ref{tab:det-acc-RDDM-Abr40} and \ref{tab:det-acc-RDDM-Grad40} report the accuracy rates obtained with different $w$ values when \gls{knn} is associated with \gls{rddm}.
We point out that trends between $w$ and accuracy rates are observed in all stream lengths in the abrupt experiments. 
In the AGRAW$_1$ and AGRAW$_2$ abrupt datasets, smaller values of $w$ induce, on average, the best mean accuracy rates when $w=800$, 69.01\% and 84.69\%, respectively, with $w=600$, commonly falling within the confidence interval.
On the other hand, larger values of $w$ induce higher accuracies in the other generators:
$w=1600$ had the best mean accuracy rates in LED (68.66\%), MIXED (94.91\%), RandRBF (33.55\%), SINE (92.00\%), and WAVEF (83.96\%).

\begin{table}[t!] 
\caption{Mean accuracies (\%) of classifiers with the RDDM detector, in  abrupt datasets, with 95\% confidence intervals (k=40)} 
\vspace{1mm}
\label{tab:det-acc-RDDM-Abr40} 
\begin{adjustbox}{max width=\textwidth} 
\begin{tabular}{cccccccccc} 
\toprule 
DS Type  & \textbf{RDDM} & \multicolumn{2}{c}{\textbf{CLASSIFIER}} & \multicolumn{6}{l}{kNN --- with k = 40, varying the size of the window (w)} \\ 
and Size & DATASET & NB & HT & w=600 & w=800 & w=1000 & w=1200 & w=1400 & w=1600 \\ 
\toprule 
 & \textbf{AGRAW$_{1}$} & 63.5577$\pm$0.28 & 64.6877$\pm$0.32 & 66.5580$\pm$0.21 & \textbf{67.1307$\pm$0.19} & 67.0258$\pm$0.22 & 66.6889$\pm$0.20 & 66.3868$\pm$0.21 & 66.3454$\pm$0.19 \\
 & \textbf{AGRAW$_{2}$} & 79.6292$\pm$1.01 & \textbf{81.5760$\pm$0.88} & 81.1649$\pm$0.19 & 81.0532$\pm$0.26 & 80.4952$\pm$0.28 & 79.8649$\pm$0.38 & 79.3032$\pm$0.45 & 78.8525$\pm$0.55 \\
ABRUPT &  \textbf{LED}  & \textbf{69.8033$\pm$0.30} & 69.7821$\pm$0.31 & 60.0839$\pm$0.37 & 61.5288$\pm$0.37 & 62.2254$\pm$0.42 & 62.4327$\pm$0.37 & 62.5349$\pm$0.37 & 62.7074$\pm$0.38 \\
10K  &  \textbf{MIXED}  & 90.2191$\pm$0.24 & 90.1677$\pm$0.25 & 89.0881$\pm$0.20 & 89.9103$\pm$0.21 & 90.2593$\pm$0.21 & 90.4284$\pm$0.19 & 90.5154$\pm$0.20 & \textbf{90.6003$\pm$0.20} \\
 &   \textbf{RandRBF}   & 30.5285$\pm$0.45 & \textbf{32.0130$\pm$0.41} & 28.3941$\pm$0.32 & 29.5232$\pm$0.33 & 30.4287$\pm$0.30 & 31.0105$\pm$0.32 & 31.4613$\pm$0.31 & 31.8940$\pm$0.35 \\
 &      \textbf{SINE}   & 86.5796$\pm$0.25 & 87.9844$\pm$0.21 & 87.5918$\pm$0.21 & 88.1595$\pm$0.20 & 88.4210$\pm$0.19 & 88.6128$\pm$0.20 & 88.7200$\pm$0.20 & \textbf{88.8010$\pm$0.20} \\
 &    \textbf{WAVEF}   & 79.1223$\pm$0.49 & 79.0917$\pm$0.49 & 81.6600$\pm$0.28 & 81.6797$\pm$0.31 & 81.7052$\pm$0.27 & 81.7760$\pm$0.29 & \textbf{81.8621$\pm$0.28} & 81.8520$\pm$0.31 \\
\midrule
 & \textbf{AGRAW$_{1}$} & 64.8903$\pm$0.16 & 68.1947$\pm$0.47 & 67.3906$\pm$0.13 & \textbf{68.3625$\pm$0.12} & 68.2927$\pm$0.13 & 67.6418$\pm$0.14 & 66.8956$\pm$0.14 & 66.3280$\pm$0.19 \\
 & \textbf{AGRAW$_{2}$} & 83.1798$\pm$0.58 & 83.1105$\pm$1.36 & 83.4166$\pm$0.10 & \textbf{83.5124$\pm$0.11} & 83.0683$\pm$0.13 & 82.2583$\pm$0.15 & 81.4971$\pm$0.20 & 80.8361$\pm$0.22 \\
ABRUPT &  \textbf{LED}  & \textbf{71.7371$\pm$0.17} & 71.7267$\pm$0.17 & 62.6087$\pm$0.28 & 64.4735$\pm$0.27 & 65.4326$\pm$0.30 & 65.9566$\pm$0.26 & 66.3614$\pm$0.24 & 66.6008$\pm$0.25 \\
20K  &  \textbf{MIXED}  & 91.0291$\pm$0.15 & 90.6606$\pm$0.15 & 91.2479$\pm$0.11 & 92.2515$\pm$0.12 & 92.8288$\pm$0.13 & 93.1657$\pm$0.12 & 93.3761$\pm$0.13 & \textbf{93.5649$\pm$0.14} \\
 &   \textbf{RandRBF}   & 30.4960$\pm$0.43 & 32.3049$\pm$0.39 & 28.7179$\pm$0.30 & 30.2488$\pm$0.28 & 31.1509$\pm$0.25 & 31.8077$\pm$0.23 & 32.3395$\pm$0.17 & \textbf{32.7963$\pm$0.20} \\
 &      \textbf{SINE}   & 87.0181$\pm$0.20 & 89.4626$\pm$0.15 & 89.0446$\pm$0.14 & 89.7690$\pm$0.13 & 90.2313$\pm$0.14 & 90.5625$\pm$0.14 & 90.7966$\pm$0.14 & \textbf{90.9688$\pm$0.13} \\
 &    \textbf{WAVEF}   & 79.7820$\pm$0.30 & 79.6376$\pm$0.27 & 82.7927$\pm$0.17 & 82.9451$\pm$0.17 & 83.1256$\pm$0.17 & 83.2151$\pm$0.17 & 83.3264$\pm$0.18 & \textbf{83.3689$\pm$0.16} \\
\midrule
 & \textbf{AGRAW$_{1}$} & 65.7342$\pm$0.11 & \textbf{72.4292$\pm$0.32} & 68.0633$\pm$0.10 & 69.1442$\pm$0.09 & 69.1142$\pm$0.10 & 68.2989$\pm$0.11 & 67.2247$\pm$0.12 & 66.3517$\pm$0.13 \\
 & \textbf{AGRAW$_{2}$} & 85.3833$\pm$0.21 & \textbf{86.0905$\pm$0.20} & 84.7950$\pm$0.07 & 84.9267$\pm$0.05 & 84.5321$\pm$0.08 & 83.7336$\pm$0.08 & 82.8497$\pm$0.11 & 82.4643$\pm$0.10 \\
ABRUPT &  \textbf{LED}  & \textbf{72.8878$\pm$0.15} & 72.8839$\pm$0.15 & 64.1554$\pm$0.19 & 66.2638$\pm$0.25 & 67.4849$\pm$0.22 & 68.1713$\pm$0.22 & 68.7251$\pm$0.21 & 69.1592$\pm$0.20 \\
50K  &  \textbf{MIXED}  & 91.5678$\pm$0.11 & 91.5976$\pm$0.15 & 92.6157$\pm$0.08 & 93.7312$\pm$0.08 & 94.3877$\pm$0.08 & 94.8288$\pm$0.07 & 95.1037$\pm$0.07 & \textbf{95.3264$\pm$0.08} \\
 &   \textbf{RandRBF}   & 30.7269$\pm$0.37 & 32.3995$\pm$0.29 & 28.9050$\pm$0.21 & 30.4971$\pm$0.19 & 31.6153$\pm$0.17 & 32.4800$\pm$0.17 & 33.1336$\pm$0.16 & \textbf{33.7147$\pm$0.16} \\
 &      \textbf{SINE}   & 87.2210$\pm$0.14 & 91.1945$\pm$0.14 & 89.8615$\pm$0.08 & 90.7122$\pm$0.07 & 91.2961$\pm$0.07 & 91.7384$\pm$0.07 & 92.0344$\pm$0.06 & \textbf{92.2682$\pm$0.06} \\
 &    \textbf{WAVEF}   & 80.1561$\pm$0.14 & 79.9377$\pm$0.16 & 83.4003$\pm$0.13 & 83.6892$\pm$0.14 & 83.8957$\pm$0.13 & 84.0001$\pm$0.13 & 84.1312$\pm$0.11 & \textbf{84.1627$\pm$0.12} \\
\midrule
 & \textbf{AGRAW$_{1}$} & 66.0837$\pm$0.08 & \textbf{74.8508$\pm$0.30} & 68.2657$\pm$0.06 & 69.4323$\pm$0.06 & 69.4108$\pm$0.05 & 68.5426$\pm$0.06 & 67.4145$\pm$0.07 & 66.4479$\pm$0.09 \\
 & \textbf{AGRAW$_{2}$} & 86.1301$\pm$0.05 & \textbf{87.1656$\pm$0.20} & 85.3222$\pm$0.05 & 85.5109$\pm$0.05 & 85.1438$\pm$0.06 & 84.3241$\pm$0.05 & 83.5378$\pm$0.08 & 83.1897$\pm$0.08 \\
ABRUPT &  \textbf{LED}  & \textbf{73.3929$\pm$0.12} & 73.3909$\pm$0.12 & 64.7987$\pm$0.13 & 66.9546$\pm$0.17 & 68.2081$\pm$0.14 & 69.0020$\pm$0.13 & 69.6307$\pm$0.13 & 70.0833$\pm$0.13 \\
100K &  \textbf{MIXED}  & 91.7850$\pm$0.06 & 92.4570$\pm$0.22 & 93.0365$\pm$0.04 & 94.1915$\pm$0.05 & 94.8809$\pm$0.04 & 95.3336$\pm$0.04 & 95.6356$\pm$0.04 & \textbf{95.8913$\pm$0.05} \\
 &   \textbf{RandRBF}   & 31.1590$\pm$0.29 & 32.8007$\pm$0.20 & 28.9599$\pm$0.15 & 30.6086$\pm$0.13 & 31.7736$\pm$0.11 & 32.6859$\pm$0.08 & 33.3913$\pm$0.10 & \textbf{34.0021$\pm$0.10} \\
 &      \textbf{SINE}   & 87.3073$\pm$0.11 & 92.4065$\pm$0.11 & 90.1356$\pm$0.06 & 91.0057$\pm$0.06 & 91.6332$\pm$0.05 & 92.1133$\pm$0.05 & 92.4443$\pm$0.04 & \textbf{92.7200$\pm$0.04} \\
 &    \textbf{WAVEF}   & 80.2496$\pm$0.11 & 79.9669$\pm$0.14 & 83.5807$\pm$0.08 & 83.8981$\pm$0.09 & 84.1558$\pm$0.08 & 84.2737$\pm$0.09 & 84.3839$\pm$0.09 & \textbf{84.4466$\pm$0.08} \\
\midrule
 & \textbf{AGRAW$_{1}$} & 66.3911$\pm$0.07 & \textbf{78.6090$\pm$0.60} & 68.4729$\pm$0.04 & 69.6481$\pm$0.04 & 69.6500$\pm$0.04 & 68.7885$\pm$0.06 & 67.5886$\pm$0.05 & 66.5210$\pm$0.04 \\
 & \textbf{AGRAW$_{2}$} & 86.7808$\pm$0.07 & \textbf{88.7905$\pm$0.16} & 85.7111$\pm$0.05 & 85.9061$\pm$0.05 & 85.5739$\pm$0.05 & 84.7680$\pm$0.04 & 83.9519$\pm$0.07 & 83.7069$\pm$0.13 \\
ABRUPT &  \textbf{LED}  & \textbf{73.7474$\pm$0.10} & 73.5808$\pm$0.10 & 65.1762$\pm$0.09 & 67.3630$\pm$0.14 & 68.6984$\pm$0.14 & 69.5252$\pm$0.16 & 70.0933$\pm$0.16 & 70.5889$\pm$0.12 \\
500K &  \textbf{MIXED}  & 92.0099$\pm$0.03 & 94.2222$\pm$0.16 & 93.3612$\pm$0.05 & 94.5453$\pm$0.06 & 95.2308$\pm$0.05 & 95.7010$\pm$0.07 & 96.0444$\pm$0.06 & \textbf{96.2771$\pm$0.06} \\
 &   \textbf{RandRBF}   & 32.1289$\pm$0.30 & 33.7327$\pm$0.27 & 28.8795$\pm$0.11 & 30.6469$\pm$0.12 & 31.8982$\pm$0.09 & 32.8279$\pm$0.08 & 33.5474$\pm$0.08 & \textbf{34.1626$\pm$0.10} \\
 &      \textbf{SINE}   & 87.3992$\pm$0.07 & \textbf{94.8885$\pm$0.34} & 90.3819$\pm$0.06 & 91.2826$\pm$0.04 & 91.9148$\pm$0.03 & 92.3752$\pm$0.04 & 92.7410$\pm$0.03 & 93.0490$\pm$0.04 \\
 &    \textbf{WAVEF}   & 80.3737$\pm$0.13 & 80.1706$\pm$0.16 & 83.7046$\pm$0.07 & 84.0010$\pm$0.08 & 84.2675$\pm$0.06 & 84.4268$\pm$0.09 & 84.5099$\pm$0.06 & \textbf{84.5950$\pm$0.08} \\
\midrule
 & \textbf{AGRAW$_{1}$} & 66.4911$\pm$0.06 & \textbf{79.2516$\pm$1.25} & 68.4707$\pm$0.05 & 69.6581$\pm$0.04 & 69.6842$\pm$0.05 & 68.8184$\pm$0.05 & 67.5832$\pm$0.05 & 66.5312$\pm$0.05 \\
 & \textbf{AGRAW$_{2}$} & 86.8647$\pm$0.02 & \textbf{88.8269$\pm$0.30} & 85.7464$\pm$0.03 & 85.9465$\pm$0.03 & 85.6038$\pm$0.03 & 84.8128$\pm$0.03 & 83.9976$\pm$0.03 & 83.7542$\pm$0.10 \\
ABRUPT &  \textbf{LED}  & \textbf{73.8174$\pm$0.07} & 73.6205$\pm$0.07 & 65.2348$\pm$0.07 & 67.4290$\pm$0.09 & 68.7355$\pm$0.09 & 69.6128$\pm$0.09 & 70.1838$\pm$0.09 & 70.7077$\pm$0.08 \\
1M  &   \textbf{MIXED}  & 92.0427$\pm$0.04 & 94.6161$\pm$0.27 & 93.4248$\pm$0.05 & 94.5997$\pm$0.05 & 95.2881$\pm$0.04 & 95.7596$\pm$0.03 & 96.1016$\pm$0.04 & \textbf{96.3437$\pm$0.04} \\
 &   \textbf{RandRBF}   & 32.1626$\pm$0.15 & \textbf{34.2701$\pm$0.33} & 28.8536$\pm$0.08 & 30.6353$\pm$0.07 & 31.9018$\pm$0.07 & 32.8187$\pm$0.05 & 33.5300$\pm$0.05 & 34.1526$\pm$0.06 \\
 &      \textbf{SINE}   & 87.4448$\pm$0.04 & \textbf{95.4360$\pm$0.22} & 90.4235$\pm$0.05 & 91.3273$\pm$0.04 & 91.9396$\pm$0.03 & 92.4035$\pm$0.04 & 92.7744$\pm$0.03 & 93.0902$\pm$0.04 \\
 &    \textbf{WAVEF}   & 80.4106$\pm$0.08 & 80.2536$\pm$0.14 & 83.7439$\pm$0.08 & 84.0563$\pm$0.07 & 84.3057$\pm$0.07 & 84.4595$\pm$0.08 & 84.5716$\pm$0.07 & \textbf{84.6452$\pm$0.07} \\
\midrule
 & \textbf{AGRAW$_{1}$} & 66.5202$\pm$0.04 & \textbf{79.3423$\pm$0.52} & 68.4835$\pm$0.03 & 69.7016$\pm$0.03 & 69.7183$\pm$0.04 & 68.8500$\pm$0.04 & 67.6267$\pm$0.04 & 66.5466$\pm$0.04 \\
 & \textbf{AGRAW$_{2}$} & 86.9036$\pm$0.02 & \textbf{89.0750$\pm$0.05} & 85.7713$\pm$0.02 & 85.9781$\pm$0.02 & 85.6340$\pm$0.02 & 84.8403$\pm$0.03 & 84.0221$\pm$0.03 & 83.7629$\pm$0.08 \\
ABRUPT &  \textbf{LED}  & \textbf{73.8735$\pm$0.04} & 73.6580$\pm$0.05 & 65.2947$\pm$0.07 & 67.4958$\pm$0.06 & 68.8241$\pm$0.07 & 69.6528$\pm$0.06 & 70.2599$\pm$0.08 & 70.7791$\pm$0.07 \\
2M  &   \textbf{MIXED}  & 92.0137$\pm$0.03 & 94.5374$\pm$0.16 & 93.4316$\pm$0.02 & 94.6047$\pm$0.02 & 95.2958$\pm$0.02 & 95.7662$\pm$0.02 & 96.1042$\pm$0.02 & \textbf{96.3500$\pm$0.02} \\
 &   \textbf{RandRBF}   & 32.1283$\pm$0.15 & \textbf{34.2618$\pm$0.20} & 28.8890$\pm$0.03 & 30.6449$\pm$0.05 & 31.8981$\pm$0.05 & 32.8152$\pm$0.06 & 33.5471$\pm$0.05 & 34.1635$\pm$0.06 \\
 &      \textbf{SINE}   & 87.4699$\pm$0.04 & \textbf{95.6762$\pm$0.12} & 90.4348$\pm$0.03 & 91.3343$\pm$0.03 & 91.9593$\pm$0.01 & 92.4246$\pm$0.01 & 92.7927$\pm$0.02 & 93.0957$\pm$0.01 \\
 &    \textbf{WAVEF}   & 80.4623$\pm$0.04 & 80.3664$\pm$0.09 & 83.7333$\pm$0.04 & 84.0793$\pm$0.05 & 84.3063$\pm$0.05 & 84.4542$\pm$0.06 & 84.5630$\pm$0.05 & \textbf{84.6395$\pm$0.06} \\
\bottomrule 
\end{tabular} 
\end{adjustbox} 
\end{table}

Comparing these results to those without drift detection in the abrupt datasets,
the use of \gls{rddm} was beneficial in MIXED and SINE datasets, improving the mean accuracy by 4.39\% and 3.96\%, respectively.
In the other generators, the impact was marginal, yielding accuracy improvements of 0.01\% in AGRAW$_1$, 0.41\% in AGRAW$_2$, and 0.49\% in LED, with decreased mean accuracies in the RandRBF (-0.40\%) and WAVEF (-0.04\%) datasets.

\begin{table}[t!] 
\caption{Mean accuracies (\%) of classifiers with the RDDM detector, in  gradual datasets, with 95\% confidence intervals (k=40)} 
\vspace{1mm}
\label{tab:det-acc-RDDM-Grad40} 
\begin{adjustbox}{max width=\textwidth} 
\begin{tabular}{cccccccccc} 
\toprule 
DS Type  & \textbf{RDDM} & \multicolumn{2}{c}{\textbf{CLASSIFIER}} & \multicolumn{6}{l}{kNN --- with k = 40, varying the size of the window (w)} \\ 
and Size & DATASET & NB & HT & w=600 & w=800 & w=1000 & w=1200 & w=1400 & w=1600 \\ 
\toprule
 & \textbf{AGRAW$_{1}$} & 62.0529$\pm$0.29 & 62.9242$\pm$0.28 & 65.9535$\pm$0.13 & \textbf{66.4526$\pm$0.18} & 66.2380$\pm$0.17 & 65.7595$\pm$0.16 & 65.4789$\pm$0.21 & 65.3699$\pm$0.23 \\
 & \textbf{AGRAW$_{2}$} & 74.2929$\pm$1.50 & 78.6547$\pm$1.02 & \textbf{79.4167$\pm$0.19} & 79.3120$\pm$0.22 & 78.8540$\pm$0.27 & 78.1824$\pm$0.28 & 77.6017$\pm$0.36 & 76.9177$\pm$0.42 \\
GRAD. &   \textbf{LED}  & \textbf{67.8506$\pm$0.30} & 67.8066$\pm$0.30 & 59.5685$\pm$0.36 & 60.8370$\pm$0.31 & 61.2801$\pm$0.38 & 61.2704$\pm$0.36 & 61.1501$\pm$0.38 & 61.3195$\pm$0.38 \\
10K  &  \textbf{MIXED}  & 83.8911$\pm$0.30 & 83.6951$\pm$0.32 & 83.7983$\pm$0.21 & 84.5068$\pm$0.21 & 84.8584$\pm$0.20 & 84.9218$\pm$0.18 & 84.9615$\pm$0.16 & \textbf{85.0175$\pm$0.19} \\
 &   \textbf{RandRBF}   & 30.3931$\pm$0.46 & \textbf{31.9290$\pm$0.40} & 28.2744$\pm$0.34 & 29.5421$\pm$0.31 & 30.4091$\pm$0.28 & 30.9701$\pm$0.31 & 31.5879$\pm$0.32 & 31.9016$\pm$0.33 \\
 &      \textbf{SINE}   & 81.8545$\pm$0.19 & 82.6644$\pm$0.20 & 82.2284$\pm$0.20 & 82.6633$\pm$0.20 & 82.8464$\pm$0.19 & 82.9898$\pm$0.17 & 83.0939$\pm$0.19 & \textbf{83.1385$\pm$0.20} \\
 &    \textbf{WAVEF}   & 78.4615$\pm$0.38 & 78.4238$\pm$0.38 & 81.6769$\pm$0.26 & 81.7966$\pm$0.28 & \textbf{81.8526$\pm$0.28} & 81.8265$\pm$0.25 & 81.8369$\pm$0.29 & 81.8178$\pm$0.29 \\
\midrule
 & \textbf{AGRAW$_{1}$} & 63.9811$\pm$0.13 & 66.8450$\pm$0.42 & 67.1249$\pm$0.12 & \textbf{68.0788$\pm$0.11} & 67.9908$\pm$0.13 & 67.3352$\pm$0.13 & 66.5441$\pm$0.14 & 65.8772$\pm$0.19 \\
 & \textbf{AGRAW$_{2}$} & 80.7887$\pm$0.95 & 82.6374$\pm$1.00 & 82.6887$\pm$0.10 & \textbf{82.8050$\pm$0.12} & 82.3863$\pm$0.12 & 81.5760$\pm$0.15 & 80.7650$\pm$0.17 & 80.1687$\pm$0.19 \\
GRAD. &   \textbf{LED}  & \textbf{70.6636$\pm$0.19} & 70.6609$\pm$0.20 & 62.1669$\pm$0.26 & 63.9369$\pm$0.30 & 64.9307$\pm$0.32 & 65.3513$\pm$0.27 & 65.7627$\pm$0.25 & 66.0927$\pm$0.25 \\
20K  &  \textbf{MIXED}  & 88.0093$\pm$0.18 & 87.5316$\pm$0.19 & 88.6662$\pm$0.12 & 89.5977$\pm$0.11 & 90.1020$\pm$0.11 & 90.3747$\pm$0.10 & 90.5672$\pm$0.13 & \textbf{90.7404$\pm$0.13} \\
 &   \textbf{RandRBF}   & 30.5298$\pm$0.49 & 32.1857$\pm$0.37 & 28.6542$\pm$0.30 & 30.1673$\pm$0.29 & 31.1504$\pm$0.22 & 31.7983$\pm$0.22 & 32.3534$\pm$0.17 & \textbf{32.8435$\pm$0.20} \\
 &      \textbf{SINE}   & 84.9811$\pm$0.16 & 86.8286$\pm$0.14 & 86.4588$\pm$0.15 & 87.1223$\pm$0.15 & 87.5382$\pm$0.13 & 87.8467$\pm$0.14 & 88.0263$\pm$0.13 & \textbf{88.2063$\pm$0.11} \\
 &    \textbf{WAVEF}   & 79.2217$\pm$0.32 & 79.1046$\pm$0.30 & 82.7925$\pm$0.16 & 82.9737$\pm$0.19 & 83.1522$\pm$0.18 & 83.2284$\pm$0.17 & 83.2939$\pm$0.16 & \textbf{83.3447$\pm$0.15} \\
\midrule
 & \textbf{AGRAW$_{1}$} & 65.3786$\pm$0.11 & \textbf{71.4328$\pm$0.32} & 68.0050$\pm$0.10 & 69.1206$\pm$0.09 & 69.0441$\pm$0.10 & 68.2060$\pm$0.10 & 67.1078$\pm$0.10 & 66.2657$\pm$0.12 \\
 & \textbf{AGRAW$_{2}$} & 84.7709$\pm$0.23 & \textbf{85.6930$\pm$0.23} & 84.5615$\pm$0.07 & 84.7147$\pm$0.06 & 84.3132$\pm$0.08 & 83.5285$\pm$0.08 & 82.6877$\pm$0.11 & 82.1707$\pm$0.12 \\
GRAD. &   \textbf{LED}  & \textbf{72.6254$\pm$0.16} & 72.6200$\pm$0.16 & 64.0380$\pm$0.20 & 66.0806$\pm$0.25 & 67.3131$\pm$0.22 & 67.9727$\pm$0.23 & 68.5119$\pm$0.21 & 68.9592$\pm$0.22 \\
50K  &  \textbf{MIXED}  & 90.4973$\pm$0.10 & 90.7385$\pm$0.10 & 91.6956$\pm$0.07 & 92.7697$\pm$0.08 & 93.4107$\pm$0.07 & 93.8288$\pm$0.06 & 94.1006$\pm$0.06 & \textbf{94.3055$\pm$0.07} \\
 &   \textbf{RandRBF}   & 30.8146$\pm$0.36 & 32.3772$\pm$0.29 & 28.8952$\pm$0.21 & 30.4700$\pm$0.19 & 31.6155$\pm$0.16 & 32.4485$\pm$0.17 & 33.1489$\pm$0.16 & \textbf{33.7053$\pm$0.16} \\
 &      \textbf{SINE}   & 86.7823$\pm$0.11 & 90.3403$\pm$0.10 & 88.9535$\pm$0.09 & 89.7705$\pm$0.07 & 90.3315$\pm$0.07 & 90.7585$\pm$0.06 & 91.0441$\pm$0.06 & \textbf{91.2728$\pm$0.06} \\
 &    \textbf{WAVEF}   & 79.9501$\pm$0.14 & 79.7088$\pm$0.15 & 83.3821$\pm$0.14 & 83.6939$\pm$0.14 & 83.8886$\pm$0.12 & 84.0075$\pm$0.13 & 84.1266$\pm$0.11 & \textbf{84.1595$\pm$0.11} \\
\midrule
 & \textbf{AGRAW$_{1}$} & 65.9170$\pm$0.08 & \textbf{74.4321$\pm$0.35} & 68.2292$\pm$0.06 & 69.4105$\pm$0.05 & 69.3951$\pm$0.05 & 68.5202$\pm$0.05 & 67.3828$\pm$0.06 & 66.3757$\pm$0.07 \\
 & \textbf{AGRAW$_{2}$} & 85.7921$\pm$0.06 & \textbf{86.9698$\pm$0.19} & 85.1806$\pm$0.06 & 85.3633$\pm$0.05 & 84.9942$\pm$0.06 & 84.1744$\pm$0.05 & 83.3086$\pm$0.08 & 82.8201$\pm$0.11 \\
GRAD. &   \textbf{LED}  & \textbf{73.2979$\pm$0.12} & 73.2963$\pm$0.12 & 64.7244$\pm$0.12 & 66.8877$\pm$0.16 & 68.1596$\pm$0.14 & 68.9571$\pm$0.12 & 69.5590$\pm$0.13 & 70.0353$\pm$0.14 \\
100K &  \textbf{MIXED}  & 91.2882$\pm$0.07 & 92.3687$\pm$0.09 & 92.6141$\pm$0.05 & 93.7481$\pm$0.04 & 94.4327$\pm$0.04 & 94.8718$\pm$0.03 & 95.1655$\pm$0.04 & \textbf{95.4179$\pm$0.04} \\
 &   \textbf{RandRBF}   & 31.2347$\pm$0.32 & 32.8390$\pm$0.20 & 28.9523$\pm$0.15 & 30.6085$\pm$0.13 & 31.7843$\pm$0.10 & 32.6854$\pm$0.09 & 33.3988$\pm$0.10 & \textbf{34.0018$\pm$0.10} \\
 &      \textbf{SINE}   & 87.1573$\pm$0.09 & 91.9854$\pm$0.09 & 89.7195$\pm$0.06 & 90.5730$\pm$0.06 & 91.1959$\pm$0.05 & 91.6733$\pm$0.05 & 92.0030$\pm$0.04 & \textbf{92.2744$\pm$0.04} \\
 &    \textbf{WAVEF}   & 80.1337$\pm$0.12 & 79.8119$\pm$0.14 & 83.5759$\pm$0.08 & 83.8892$\pm$0.09 & 84.1520$\pm$0.08 & 84.2631$\pm$0.08 & 84.3712$\pm$0.08 & \textbf{84.4458$\pm$0.08} \\
\midrule
 & \textbf{AGRAW$_{1}$} & 66.3627$\pm$0.05 & \textbf{78.1204$\pm$0.97} & 68.4620$\pm$0.04 & 69.6387$\pm$0.04 & 69.6415$\pm$0.04 & 68.7873$\pm$0.05 & 67.5755$\pm$0.05 & 66.4971$\pm$0.06 \\
 & \textbf{AGRAW$_{2}$} & 86.6685$\pm$0.08 & \textbf{88.7404$\pm$0.20} & 85.6761$\pm$0.05 & 85.8748$\pm$0.04 & 85.5452$\pm$0.05 & 84.7423$\pm$0.04 & 83.9326$\pm$0.08 & 83.6401$\pm$0.12 \\
GRAD. &   \textbf{LED}  & \textbf{73.7194$\pm$0.10} & 73.5666$\pm$0.10 & 65.1719$\pm$0.10 & 67.3267$\pm$0.14 & 68.6930$\pm$0.15 & 69.5160$\pm$0.16 & 70.0844$\pm$0.14 & 70.5927$\pm$0.12 \\
500K &  \textbf{MIXED}  & 91.9064$\pm$0.03 & 94.0055$\pm$0.16 & 93.2873$\pm$0.05 & 94.4550$\pm$0.06 & 95.1380$\pm$0.05 & 95.6023$\pm$0.07 & 95.9367$\pm$0.07 & \textbf{96.1795$\pm$0.06} \\
 &   \textbf{RandRBF}   & 32.1382$\pm$0.30 & 33.9795$\pm$0.33 & 28.8762$\pm$0.11 & 30.6384$\pm$0.11 & 31.9050$\pm$0.08 & 32.8314$\pm$0.07 & 33.5418$\pm$0.07 & \textbf{34.1613$\pm$0.10} \\
 &      \textbf{SINE}   & 87.3896$\pm$0.06 & \textbf{94.8852$\pm$0.37} & 90.3028$\pm$0.06 & 91.1958$\pm$0.04 & 91.8166$\pm$0.03 & 92.2822$\pm$0.03 & 92.6491$\pm$0.04 & 92.9538$\pm$0.04 \\
 &    \textbf{WAVEF}   & 80.3325$\pm$0.16 & 80.1808$\pm$0.23 & 83.7033$\pm$0.08 & 84.0057$\pm$0.08 & 84.2744$\pm$0.06 & 84.4214$\pm$0.09 & 84.5056$\pm$0.05 & \textbf{84.5917$\pm$0.09} \\
\midrule
 & \textbf{AGRAW$_{1}$} & 66.4404$\pm$0.06 & \textbf{78.9543$\pm$0.69} & 68.4625$\pm$0.05 & 69.6584$\pm$0.05 & 69.6799$\pm$0.05 & 68.8083$\pm$0.05 & 67.5885$\pm$0.04 & 66.5405$\pm$0.05 \\
 & \textbf{AGRAW$_{2}$} & 86.8139$\pm$0.04 & \textbf{88.8079$\pm$0.22} & 85.7296$\pm$0.02 & 85.9266$\pm$0.03 & 85.5852$\pm$0.03 & 84.7923$\pm$0.03 & 83.9757$\pm$0.04 & 83.6798$\pm$0.12 \\
GRAD. &   \textbf{LED}  & \textbf{73.7930$\pm$0.06} & 73.6111$\pm$0.07 & 65.2150$\pm$0.07 & 67.4254$\pm$0.09 & 68.7352$\pm$0.08 & 69.6016$\pm$0.09 & 70.1805$\pm$0.09 & 70.6990$\pm$0.08 \\
1M  &   \textbf{MIXED}  & 91.9926$\pm$0.04 & 94.3767$\pm$0.28 & 93.3777$\pm$0.04 & 94.5572$\pm$0.05 & 95.2424$\pm$0.04 & 95.7050$\pm$0.04 & 96.0485$\pm$0.04 & \textbf{96.2892$\pm$0.04} \\
 &   \textbf{RandRBF}   & 32.0952$\pm$0.18 & \textbf{34.2234$\pm$0.35} & 28.8565$\pm$0.08 & 30.6366$\pm$0.07 & 31.8997$\pm$0.07 & 32.8175$\pm$0.05 & 33.5382$\pm$0.05 & 34.1613$\pm$0.06 \\
 &      \textbf{SINE}   & 87.4518$\pm$0.05 & \textbf{95.4554$\pm$0.34} & 90.3766$\pm$0.05 & 91.2760$\pm$0.04 & 91.8907$\pm$0.03 & 92.3560$\pm$0.03 & 92.7313$\pm$0.03 & 93.0397$\pm$0.04 \\
 &    \textbf{WAVEF}   & 80.3965$\pm$0.07 & 80.3126$\pm$0.16 & 83.7449$\pm$0.08 & 84.0564$\pm$0.07 & 84.3087$\pm$0.07 & 84.4603$\pm$0.08 & 84.5764$\pm$0.07 & \textbf{84.6390$\pm$0.07} \\
\midrule
 & \textbf{AGRAW$_{1}$} & 66.5098$\pm$0.04 & \textbf{79.5348$\pm$0.69} & 68.4834$\pm$0.04 & 69.7001$\pm$0.03 & 69.7140$\pm$0.04 & 68.8516$\pm$0.04 & 67.6256$\pm$0.04 & 66.5625$\pm$0.04 \\
 & \textbf{AGRAW$_{2}$} & 86.8776$\pm$0.02 & \textbf{89.1013$\pm$0.07} & 85.7594$\pm$0.02 & 85.9723$\pm$0.03 & 85.6274$\pm$0.02 & 84.8323$\pm$0.03 & 84.0195$\pm$0.03 & 83.7543$\pm$0.06 \\
GRAD. &   \textbf{LED}  & \textbf{73.8586$\pm$0.05} & 73.6469$\pm$0.05 & 65.2935$\pm$0.07 & 67.4867$\pm$0.06 & 68.8312$\pm$0.08 & 69.6431$\pm$0.05 & 70.2493$\pm$0.08 & 70.7769$\pm$0.07 \\
2M  &   \textbf{MIXED}  & 91.9840$\pm$0.03 & 94.5502$\pm$0.19 & 93.4082$\pm$0.02 & 94.5849$\pm$0.02 & 95.2719$\pm$0.02 & 95.7409$\pm$0.02 & 96.0740$\pm$0.03 & \textbf{96.3276$\pm$0.02} \\
 &   \textbf{RandRBF}   & 32.1311$\pm$0.12 & \textbf{34.3354$\pm$0.17} & 28.8895$\pm$0.03 & 30.6460$\pm$0.05 & 31.8985$\pm$0.05 & 32.8141$\pm$0.06 & 33.5483$\pm$0.05 & 34.1572$\pm$0.05 \\
 &      \textbf{SINE}   & 87.4597$\pm$0.03 & \textbf{95.7088$\pm$0.13} & 90.4156$\pm$0.03 & 91.3107$\pm$0.03 & 91.9316$\pm$0.02 & 92.3982$\pm$0.01 & 92.7666$\pm$0.02 & 93.0682$\pm$0.01 \\
 &    \textbf{WAVEF}   & 80.4472$\pm$0.05 & 80.3168$\pm$0.12 & 83.7325$\pm$0.04 & 84.0787$\pm$0.05 & 84.3056$\pm$0.05 & 84.4500$\pm$0.06 & 84.5644$\pm$0.05 & \textbf{84.6389$\pm$0.06} \\
\bottomrule 
\end{tabular} 
\end{adjustbox} 
\end{table}

Regarding the volatility of the results in the abrupt datasets, the impact of changing $w$ with \gls{rddm} is similar to what has been observed without drift detection.
The mean differences between the best and worst-performing $w$ were: 2.58\% in AGRAW$_1$, 2.34\% in AGRAW$_2$, 4.75\% in LED, 2.59\% in MIXED, 4.76\% in RandRBF, 2.30\% in SINE, and 0.72\% in WAVEF.

Conducting similar analyses on the gradual experiments, the same trends between $w$ and accuracy hold.
More specifically, the accuracy peaks with $w=800$ for AGRAW$_1$ and AGRAW$_2$, and with $w=1600$ in the other generators, thus matching the behavior mentioned above.
The best mean results per generator were: 68.87\% in AGRAW$_1$, 84.28\% in AGRAW$_2$, 68.35\% in LED, 93.47\% in MIXED, 33.56\% in RandRBF, 90.56\% in SINE, and 83.95\% in WAVEF.

Comparing these results to those without \gls{rddm} in the gradual datasets shows that the impact of using \gls{rddm} is slightly worse than in the abrupt datasets. 
In particular, there are larger accuracy drops in AGRAW$_1$ with $w \geq 1400$. 
The mean differences were -0.06\% in AGRAW$_1$, 0.23\% in AGRAW$_2$, 0.20\% in LED, 3.09\% in MIXED, -0.39\% in RandRBF, 2.69\% in SINE, and -0.07\% in WAVEF.

Despite the differences in the behavior of \gls{knn} with abrupt and gradual datasets, the volatility rates in the gradual datasets were quite similar to those in the abrupt datasets: 2.60\% in AGRAW$_1$, 2.41\% in AGRAW$_2$, 4.61\% in LED, 2.49\% in MIXED, 4.79\% in RandRBF, 2.21\% in SINE, and 0.72\% in WAVEF.

\begin{figure}[!t]
\vspace{+2mm}
\centering
\includegraphics[width=0.8\linewidth]{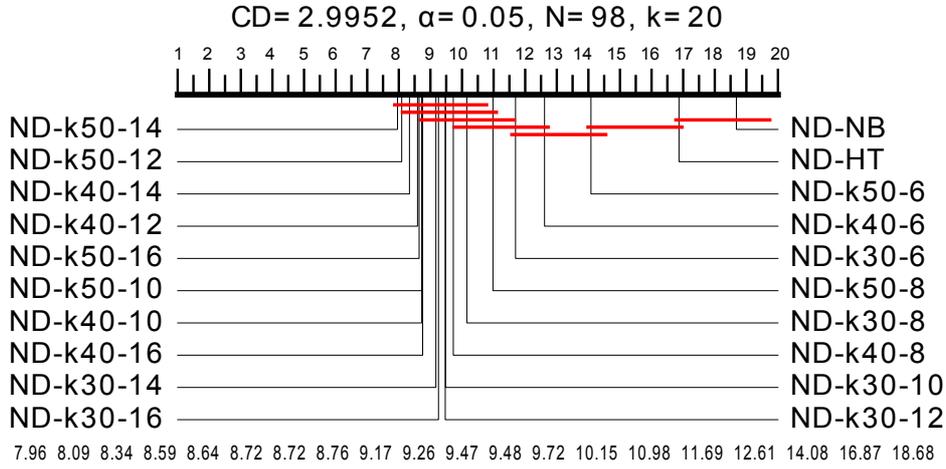}
\vspace{-15mm}
\caption{Comparison of accuracy results of classifiers with no drift detector in all datasets with 95\% confidence.}
\label{fig:Acc-w-NoDet-Acc-All}
\vspace{-1mm}
\end{figure}

As in the previous section, the analysis of the different $w$ values is complemented with an evaluation using the $F_F$ statistic and the Nemenyi posthoc test \cite{demsar:2006}. 
Figure \ref{fig:Acc-w-NoDet-Acc-All} compares different \gls{knn} configurations including the best three performing neighborhood sizes ($k \in \{30, 40, 50\}$) and window sizes $w \in \{600, 800, 1000, 1200, 1400, 1600\}$ with no drift detector in terms of accuracy, aggregating the abrupt and gradual dataset results.
In the names of the methods, ND stands for {\em No Detector}, the number after $k$ is the number of neighbors, and the last number is the window size $w$ divided by 100.
The results obtained solely for abrupt and gradual experiments were omitted as the rankings, and statistical differences match those observed with their combination.

According to the ranks, we observe that \gls{knn} configurations are the best performers. 
More specifically, we point out that, although $k=50$ with $w=1400$ is the best ranked parametrization, many combinations are statistically similar, including $w \in \{1000, 1200, 1400, 1600\}$ regardless of $k$ and $w=800$ with $k \in \{30, 40\}$.
Therefore, these results corroborate the findings discussed earlier in this section, thus reporting that larger values of $w$ yielded the best accuracy results across the conducted experiments, especially on those with a larger number of instances and longer concepts, which are important as data streams are potentially unbounded.
Furthermore, we also observe that most configurations of \gls{knn} were statistically superior to \gls{ht} and \gls{nb}.
A more in-depth analysis and comparison between \gls{knn} and these two classifiers is presented in Section \ref{comp:nb:ht}.

\begin{figure}
\centering
\includegraphics[width=0.8\linewidth]{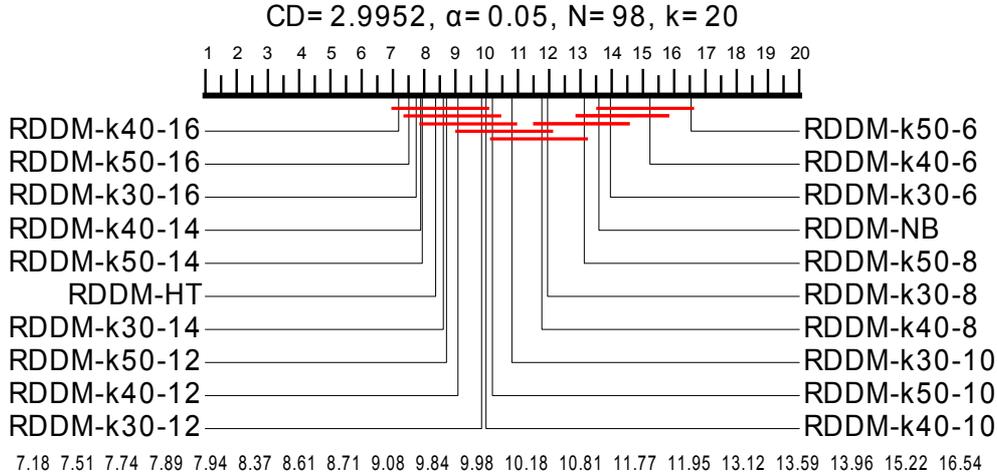}
\vspace{-15mm}
\caption{Comparison of accuracy results of classifiers with the RDDM detector in all datasets with 95\% confidence.}
\label{fig:Acc-w-RDDM-Acc-All}
\end{figure}

Finally, Figure \ref{fig:Acc-w-RDDM-Acc-All} captures the accuracy statistical results of the  \gls{knn} configurations using \gls{rddm} as drift detector in all the experiments that were depicted in Tables \ref{tab:det-acc-RDDM-Abr40}, and \ref{tab:det-acc-RDDM-Grad40}.
Even though small differences between abrupt, gradual, and all datasets exist, they occur in the worst-ranked configurations. 
In this comparison, we observe that $k=40$ with $w=1600$ is the best performing \gls{knn} configuration. 
Nonetheless, this configuration is statistically similar to all configurations using $w \in \{1200, 1400, 1600\}$ and $w=1000$ with $k \in \{40, 50\}$, as well as \gls{ht}.

\subsection{Run-time analysis}

This subsection now analyzes the run-time varying the window size $w$ of \gls{knn}.
The results using \gls{rddm} as drift detector and \gls{knn} with $k=40$ in abrupt and gradual datasets are presented in Tables \ref{tab:det-time-RDDM-Abr40} and \ref{tab:det-time-RDDM-Grad40}, respectively. 
The corresponding results using $k \in \{30, 50\}$ as well as those without the presence of a concept drift detector are all included in the appendix as Tables \ref{tab:det-time-NoDet-Abr30}--\ref{tab:det-time-RDDM-Grad50}.
In general, it is possible to note that, with few exceptions, a larger $w$ demands more run-time, regardless of the dataset used.
The reader is reminded that, as discussed in Subsection \ref{survey:knn}, $w$ is the variable with the most significant influence on the time complexity of \gls{knn}.
Therefore, its increase can compromise its performance and should be carefully assessed.

In addition to $w$, another variable that influences the run-time of \gls{knn} is the number of attributes of the dataset ($d$).
This characteristic can be confirmed by observing the performance of \gls{knn} in the LED, RandRBF, and Waveform datasets, which have more attributes than the others.
The explanation for this increase is related to the way \gls{knn} finds the closest neighbors.
In order to calculate the distance between two different instances and check the level of proximity between them, all attributes need to be covered.
Thus, the overhead will be directly proportional to the number of attributes that the dataset has.

\begin{table}[t!] 
\caption{Mean run-time (sec) of classifiers with the RDDM detector, in abrupt datasets, with 95\% confidence intervals (k=40)} 
\vspace{1mm}
\label{tab:det-time-RDDM-Abr40} 
\begin{adjustbox}{max width=\textwidth} 
\begin{tabular}{cccccccccc} 
\toprule
DS Type  & \textbf{RDDM} & \multicolumn{2}{c}{\textbf{CLASSIFIER}} & \multicolumn{6}{l}{kNN --- with k = 40, varying the size of the window (w)} \\ 
and Size & DATASET & NB & HT & w=600 & w=800 & w=1000 & w=1200 & w=1400 & w=1600 \\ 
\toprule
 & \textbf{AGRAW$_{1}$} & \textbf{0.319$\pm$0.02} & 0.518$\pm$0.03 & 2.880$\pm$0.10 & 3.341$\pm$0.12 & 3.793$\pm$0.13 & 3.962$\pm$0.14 & 4.056$\pm$0.13 & 4.520$\pm$0.15 \\
 & \textbf{AGRAW$_{2}$} & \textbf{0.297$\pm$0.02} & 0.577$\pm$0.03 & 3.007$\pm$0.07 & 3.661$\pm$0.08 & 4.125$\pm$0.12 & 4.389$\pm$0.17 & 4.718$\pm$0.19 & 5.314$\pm$0.27 \\
ABRUPT & \textbf{LED} & \textbf{0.825$\pm$0.03} & 1.088$\pm$0.04 & 4.967$\pm$0.08 & 6.084$\pm$0.11 & 6.716$\pm$0.13 & 7.405$\pm$0.13 & 7.550$\pm$0.15 & 8.326$\pm$0.17 \\
10K & \textbf{MIXED} & \textbf{0.159$\pm$0.01} & 0.277$\pm$0.02 & 2.057$\pm$0.09 & 2.696$\pm$0.07 & 2.927$\pm$0.07 & 3.137$\pm$0.09 & 3.198$\pm$0.09 & 3.608$\pm$0.11 \\
 & \textbf{RandRBF} & \textbf{1.464$\pm$0.06} & 2.355$\pm$0.07 & 4.641$\pm$0.09 & 5.674$\pm$0.18 & 6.462$\pm$0.21 & 7.079$\pm$0.24 & 7.327$\pm$0.20 & 8.914$\pm$0.29 \\
 & \textbf{SINE} & \textbf{0.181$\pm$0.01} & 0.487$\pm$0.04 & 2.111$\pm$0.05 & 2.760$\pm$0.07 & 3.056$\pm$0.10 & 3.194$\pm$0.10 & 3.302$\pm$0.12 & 3.817$\pm$0.09 \\
 & \textbf{WAVEF} & \textbf{0.658$\pm$0.03} & 1.104$\pm$0.05 & 3.469$\pm$0.10 & 4.373$\pm$0.13 & 5.021$\pm$0.15 & 5.461$\pm$0.15 & 5.710$\pm$0.15 & 6.422$\pm$0.14 \\
\midrule 
 & \textbf{AGRAW$_{1}$} & \textbf{0.598$\pm$0.03} & 1.630$\pm$0.11 & 6.159$\pm$0.13 & 7.484$\pm$0.16 & 8.352$\pm$0.21 & 9.368$\pm$0.21 & 9.880$\pm$0.22 & 11.146$\pm$0.31 \\
 & \textbf{AGRAW$_{2}$} & \textbf{0.579$\pm$0.03} & 2.446$\pm$0.88 & 6.153$\pm$0.15 & 7.609$\pm$0.13 & 8.720$\pm$0.15 & 9.908$\pm$0.16 & 10.673$\pm$0.24 & 12.180$\pm$0.28 \\
ABRUPT & \textbf{LED} & \textbf{1.622$\pm$0.04} & 2.108$\pm$0.06 & 10.558$\pm$0.19 & 13.017$\pm$0.24 & 14.698$\pm$0.29 & 17.243$\pm$0.31 & 18.252$\pm$0.32 & 21.074$\pm$0.34 \\
20K & \textbf{MIXED} & \textbf{0.250$\pm$0.02} & 0.548$\pm$0.04 & 4.460$\pm$0.15 & 5.903$\pm$0.10 & 6.689$\pm$0.12 & 7.483$\pm$0.17 & 8.029$\pm$0.18 & 9.208$\pm$0.17 \\
 & \textbf{RandRBF} & \textbf{2.973$\pm$0.10} & 4.540$\pm$0.08 & 9.509$\pm$0.18 & 11.768$\pm$0.21 & 13.686$\pm$0.34 & 15.377$\pm$0.23 & 16.516$\pm$0.28 & 19.303$\pm$0.30 \\
 & \textbf{SINE} & \textbf{0.314$\pm$0.02} & 1.160$\pm$0.06 & 4.662$\pm$0.10 & 6.076$\pm$0.11 & 6.876$\pm$0.12 & 7.664$\pm$0.17 & 8.258$\pm$0.20 & 9.445$\pm$0.18 \\
 & \textbf{WAVEF} & \textbf{1.278$\pm$0.03} & 2.137$\pm$0.07 & 7.110$\pm$0.18 & 9.418$\pm$0.16 & 10.676$\pm$0.17 & 12.427$\pm$0.19 & 13.476$\pm$0.23 & 15.542$\pm$0.33 \\
\midrule 
 & \textbf{AGRAW$_{1}$} & \textbf{1.368$\pm$0.06} & 9.993$\pm$0.89 & 16.144$\pm$0.20 & 19.438$\pm$0.24 & 22.545$\pm$0.31 & 25.397$\pm$0.26 & 27.543$\pm$0.37 & 32.190$\pm$0.52 \\
 & \textbf{AGRAW$_{2}$} & \textbf{1.403$\pm$0.04} & 7.477$\pm$0.87 & 15.912$\pm$0.20 & 19.490$\pm$0.25 & 22.289$\pm$0.31 & 26.203$\pm$0.33 & 27.865$\pm$0.38 & 32.602$\pm$0.38 \\
ABRUPT & \textbf{LED} & \textbf{4.146$\pm$0.13} & 5.162$\pm$0.09 & 27.886$\pm$0.23 & 34.156$\pm$0.36 & 40.061$\pm$0.48 & 46.347$\pm$0.35 & 50.433$\pm$0.46 & 59.610$\pm$0.50 \\
50K & \textbf{MIXED} & \textbf{0.677$\pm$0.03} & 2.261$\pm$0.25 & 11.664$\pm$0.18 & 15.743$\pm$0.20 & 18.029$\pm$0.24 & 20.552$\pm$0.26 & 22.698$\pm$0.24 & 26.288$\pm$0.33 \\
 & \textbf{RandRBF} & \textbf{7.322$\pm$0.27} & 11.263$\pm$0.14 & 24.774$\pm$0.20 & 30.782$\pm$0.41 & 36.740$\pm$0.42 & 42.367$\pm$0.42 & 45.962$\pm$0.46 & 53.377$\pm$0.59 \\
 & \textbf{SINE} & \textbf{0.782$\pm$0.04} & 4.006$\pm$0.13 & 11.836$\pm$0.20 & 15.931$\pm$0.25 & 18.774$\pm$0.25 & 21.013$\pm$0.22 & 23.570$\pm$0.33 & 27.179$\pm$0.35 \\
 & \textbf{WAVEF} & \textbf{3.113$\pm$0.10} & 7.093$\pm$0.66 & 18.638$\pm$0.20 & 24.313$\pm$0.28 & 28.510$\pm$0.39 & 33.014$\pm$0.36 & 36.513$\pm$0.52 & 42.707$\pm$0.56 \\
\midrule 
 & \textbf{AGRAW$_{1}$} & \textbf{2.792$\pm$0.06} & 31.819$\pm$1.38 & 32.343$\pm$0.27 & 39.843$\pm$0.33 & 45.094$\pm$0.38 & 52.874$\pm$0.37 & 58.307$\pm$0.67 & 65.797$\pm$0.84 \\
 & \textbf{AGRAW$_{2}$} & \textbf{2.796$\pm$0.05} & 27.968$\pm$3.02 & 32.088$\pm$0.36 & 39.674$\pm$0.42 & 45.586$\pm$0.37 & 53.411$\pm$0.32 & 59.061$\pm$0.45 & 65.851$\pm$0.59 \\
ABRUPT & \textbf{LED} & \textbf{8.312$\pm$0.17} & 10.357$\pm$0.13 & 56.792$\pm$0.47 & 69.884$\pm$0.48 & 82.122$\pm$0.62 & 95.334$\pm$0.72 & 106.616$\pm$0.82 & 121.433$\pm$0.83 \\
100K & \textbf{MIXED} & \textbf{1.346$\pm$0.04} & 6.402$\pm$0.61 & 23.926$\pm$0.27 & 32.157$\pm$0.36 & 36.626$\pm$0.40 & 42.484$\pm$0.29 & 48.188$\pm$0.42 & 55.305$\pm$0.46 \\
 & \textbf{RandRBF} & \textbf{15.107$\pm$0.23} & 22.811$\pm$0.44 & 51.316$\pm$0.48 & 61.870$\pm$0.55 & 73.769$\pm$0.67 & 83.268$\pm$0.73 & 95.655$\pm$0.98 & 112.761$\pm$1.12 \\
 & \textbf{SINE} & \textbf{1.604$\pm$0.05} & 11.262$\pm$0.25 & 24.736$\pm$0.19 & 32.890$\pm$0.24 & 38.155$\pm$0.29 & 43.624$\pm$0.45 & 50.124$\pm$0.42 & 56.303$\pm$0.39 \\
 & \textbf{WAVEF} & \textbf{6.330$\pm$0.19} & 15.213$\pm$1.34 & 38.062$\pm$0.37 & 49.034$\pm$0.31 & 57.946$\pm$0.40 & 66.461$\pm$0.61 & 76.216$\pm$0.67 & 87.126$\pm$0.73 \\
\midrule 
 & \textbf{AGRAW$_{1}$} & \textbf{14.385$\pm$0.45} & 267.279$\pm$10.18 & 164.869$\pm$2.44 & 200.422$\pm$1.91 & 231.193$\pm$2.78 & 265.770$\pm$3.26 & 299.854$\pm$2.43 & 331.553$\pm$2.09 \\
 & \textbf{AGRAW$_{2}$} & \textbf{14.183$\pm$0.42} & 310.263$\pm$37.71 & 164.909$\pm$2.57 & 199.408$\pm$2.42 & 231.368$\pm$2.92 & 263.368$\pm$2.55 & 298.249$\pm$2.35 & 331.999$\pm$4.71 \\
ABRUPT & \textbf{LED} & \textbf{42.866$\pm$0.60} & 51.234$\pm$0.54 & 285.640$\pm$3.13 & 348.651$\pm$2.52 & 417.390$\pm$6.64 & 493.874$\pm$3.07 & 541.566$\pm$5.00 & 616.839$\pm$7.23 \\
500K & \textbf{MIXED} & \textbf{6.935$\pm$0.18} & 71.627$\pm$4.78 & 123.294$\pm$1.12 & 164.465$\pm$1.52 & 189.926$\pm$1.24 & 214.894$\pm$2.28 & 248.745$\pm$2.27 & 282.639$\pm$1.34 \\
 & \textbf{RandRBF} & \textbf{76.283$\pm$1.47} & 177.338$\pm$26.44 & 256.571$\pm$2.68 & 317.422$\pm$2.52 & 377.497$\pm$5.01 & 429.389$\pm$3.83 & 486.341$\pm$7.89 & 558.656$\pm$9.44 \\
 & \textbf{SINE} & \textbf{7.959$\pm$0.26} & 116.363$\pm$5.74 & 126.306$\pm$0.87 & 167.044$\pm$1.41 & 195.455$\pm$0.97 & 222.403$\pm$1.79 & 254.497$\pm$2.64 & 291.261$\pm$1.35 \\
 & \textbf{WAVEF} & \textbf{32.869$\pm$0.56} & 115.112$\pm$17.91 & 190.545$\pm$1.52 & 250.322$\pm$1.35 & 294.453$\pm$2.45 & 343.101$\pm$4.56 & 380.811$\pm$3.26 & 441.420$\pm$6.06 \\
\midrule 
 & \textbf{AGRAW$_{1}$} & \textbf{29.151$\pm$0.53} & 613.609$\pm$12.21 & 332.662$\pm$2.21 & 404.782$\pm$1.91 & 466.950$\pm$3.46 & 536.966$\pm$3.79 & 610.414$\pm$7.25 & 679.691$\pm$6.83 \\
 & \textbf{AGRAW$_{2}$} & \textbf{28.907$\pm$0.44} & 731.622$\pm$69.98 & 331.842$\pm$2.65 & 406.986$\pm$2.15 & 469.093$\pm$3.45 & 533.388$\pm$2.16 & 602.527$\pm$7.74 & 657.011$\pm$10.13 \\
ABRUPT & \textbf{LED} & \textbf{83.341$\pm$2.42} & 104.589$\pm$0.94 & 555.282$\pm$3.63 & 698.148$\pm$3.44 & 825.924$\pm$6.01 & 845.836$\pm$120.57 & 1086.192$\pm$8.69 & 1222.745$\pm$9.76 \\
1M & \textbf{MIXED} & \textbf{13.748$\pm$0.20} & 173.155$\pm$19.26 & 245.027$\pm$1.45 & 330.236$\pm$2.08 & 380.118$\pm$2.48 & 433.574$\pm$3.19 & 506.311$\pm$6.16 & 564.576$\pm$3.61 \\
 & \textbf{RandRBF} & \textbf{152.738$\pm$4.13} & 479.969$\pm$71.13 & 505.495$\pm$3.65 & 636.589$\pm$4.26 & 735.234$\pm$9.34 & 718.289$\pm$113.36 & 1002.490$\pm$15.09 & 1112.521$\pm$12.18 \\
 & \textbf{SINE} & \textbf{16.264$\pm$0.31} & 270.707$\pm$12.79 & 248.902$\pm$1.05 & 334.615$\pm$1.77 & 384.699$\pm$2.32 & 449.010$\pm$3.32 & 521.292$\pm$6.18 & 581.554$\pm$3.81 \\
 & \textbf{WAVEF} & \textbf{64.648$\pm$2.13} & 256.825$\pm$35.23 & 378.840$\pm$3.09 & 500.388$\pm$2.25 & 583.706$\pm$5.56 & 616.277$\pm$63.75 & 780.205$\pm$10.69 & 871.685$\pm$6.17 \\
\midrule 
 & \textbf{AGRAW$_{1}$} & \textbf{56.447$\pm$2.39} & 1295.659$\pm$69.00 & 654.983$\pm$5.71 & 803.660$\pm$4.31 & 932.449$\pm$8.15 & 868.156$\pm$139.03 & 1209.506$\pm$11.16 & 1352.825$\pm$11.09 \\
 & \textbf{AGRAW$_{2}$} & \textbf{56.405$\pm$2.01} & 1820.463$\pm$611.90 & 650.418$\pm$6.10 & 798.790$\pm$4.62 & 926.054$\pm$8.94 & 843.577$\pm$140.92 & 1216.810$\pm$10.40 & 1316.239$\pm$7.36 \\
ABRUPT & \textbf{LED} & \textbf{167.647$\pm$3.85} & 211.517$\pm$1.94 & 1125.144$\pm$5.31 & 1387.918$\pm$5.90 & 1641.316$\pm$10.64 & 1200.967$\pm$259.25 & 2181.654$\pm$23.49 & 2352.130$\pm$186.76 \\
2M & \textbf{MIXED} & \textbf{27.654$\pm$0.22} & 347.565$\pm$20.12 & 479.585$\pm$3.43 & 655.376$\pm$5.57 & 759.208$\pm$6.96 & 647.524$\pm$106.95 & 1017.468$\pm$8.42 & 1120.406$\pm$8.70 \\
 & \textbf{RandRBF} & \textbf{299.654$\pm$4.82} & 996.595$\pm$116.80 & 1018.670$\pm$6.51 & 1252.965$\pm$15.31 & 1468.210$\pm$16.88 & 901.017$\pm$147.37 & 2002.443$\pm$32.11 & 2052.680$\pm$219.90 \\
 & \textbf{SINE} & \textbf{32.225$\pm$0.28} & 586.486$\pm$25.12 & 492.545$\pm$4.17 & 671.387$\pm$4.91 & 772.239$\pm$8.40 & 660.636$\pm$111.33 & 1051.882$\pm$8.49 & 1146.401$\pm$6.25 \\
 & \textbf{WAVEF} & \textbf{126.033$\pm$4.32} & 565.137$\pm$46.15 & 755.907$\pm$4.43 & 998.454$\pm$6.86 & 1158.732$\pm$8.30 & 853.317$\pm$169.27 & 1543.309$\pm$15.84 & 1724.598$\pm$47.52 \\
\bottomrule 
\end{tabular} 
\end{adjustbox} 
\end{table}

\begin{table}[t!] 
\caption{Mean run-time (sec) of classifiers with the RDDM detector, in gradual datasets, with 95\% confidence intervals (k=40)} 
\vspace{1mm}
\label{tab:det-time-RDDM-Grad40} 
\begin{adjustbox}{max width=\textwidth} 
\begin{tabular}{cccccccccc} 
\toprule
DS Type  & \textbf{RDDM} & \multicolumn{2}{c}{\textbf{CLASSIFIER}} & \multicolumn{6}{l}{kNN --- with k = 40, varying the size of the window (w)} \\ 
and Size & DATASET & NB & HT & w=600 & w=800 & w=1000 & w=1200 & w=1400 & w=1600 \\ 
\toprule
 & \textbf{AGRAW$_{1}$} & \textbf{0.314$\pm$0.02} & 0.479$\pm$0.04 & 2.917$\pm$0.08 & 3.409$\pm$0.13 & 3.822$\pm$0.12 & 3.910$\pm$0.14 & 4.359$\pm$0.14 & 4.460$\pm$0.21 \\
 & \textbf{AGRAW$_{2}$} & \textbf{0.316$\pm$0.02} & 0.538$\pm$0.03 & 2.942$\pm$0.07 & 3.639$\pm$0.13 & 4.120$\pm$0.12 & 4.294$\pm$0.15 & 5.005$\pm$0.20 & 5.338$\pm$0.25 \\
GRAD. & \textbf{LED} & \textbf{0.868$\pm$0.04} & 1.045$\pm$0.04 & 5.062$\pm$0.11 & 6.136$\pm$0.10 & 6.987$\pm$0.16 & 7.124$\pm$0.22 & 8.431$\pm$0.20 & 8.461$\pm$0.23 \\
10K & \textbf{MIXED} & \textbf{0.157$\pm$0.02} & 0.263$\pm$0.03 & 1.998$\pm$0.05 & 2.546$\pm$0.06 & 2.740$\pm$0.06 & 2.887$\pm$0.08 & 3.266$\pm$0.08 & 3.272$\pm$0.08 \\
 & \textbf{RandRBF} & \textbf{1.496$\pm$0.05} & 2.289$\pm$0.07 & 4.631$\pm$0.07 & 5.451$\pm$0.13 & 6.437$\pm$0.17 & 6.731$\pm$0.19 & 8.035$\pm$0.21 & 8.729$\pm$0.23 \\
 & \textbf{SINE} & \textbf{0.182$\pm$0.02} & 0.387$\pm$0.03 & 1.972$\pm$0.08 & 2.607$\pm$0.06 & 2.798$\pm$0.09 & 2.998$\pm$0.10 & 3.387$\pm$0.10 & 3.396$\pm$0.09 \\
 & \textbf{WAVEF} & \textbf{0.647$\pm$0.03} & 1.017$\pm$0.05 & 3.547$\pm$0.09 & 4.563$\pm$0.11 & 5.270$\pm$0.12 & 5.695$\pm$0.17 & 6.456$\pm$0.24 & 6.877$\pm$0.31 \\
\midrule 
 & \textbf{AGRAW$_{1}$} & \textbf{0.594$\pm$0.03} & 1.710$\pm$0.16 & 6.135$\pm$0.14 & 7.471$\pm$0.18 & 8.387$\pm$0.19 & 9.074$\pm$0.18 & 10.464$\pm$0.23 & 11.302$\pm$0.30 \\
 & \textbf{AGRAW$_{2}$} & \textbf{0.557$\pm$0.03} & 1.701$\pm$0.59 & 6.234$\pm$0.11 & 7.629$\pm$0.18 & 8.653$\pm$0.22 & 9.362$\pm$0.22 & 11.245$\pm$0.23 & 12.097$\pm$0.24 \\
GRAD. & \textbf{LED} & \textbf{1.614$\pm$0.06} & 2.018$\pm$0.07 & 10.767$\pm$0.17 & 12.680$\pm$0.22 & 15.006$\pm$0.23 & 16.474$\pm$0.23 & 19.034$\pm$0.28 & 20.612$\pm$0.24 \\
20K & \textbf{MIXED} & \textbf{0.275$\pm$0.02} & 0.509$\pm$0.03 & 4.343$\pm$0.12 & 5.735$\pm$0.14 & 6.380$\pm$0.16 & 7.030$\pm$0.15 & 8.251$\pm$0.13 & 8.841$\pm$0.13 \\
 & \textbf{RandRBF} & \textbf{2.998$\pm$0.08} & 4.330$\pm$0.09 & 9.681$\pm$0.15 & 11.924$\pm$0.18 & 13.773$\pm$0.27 & 14.706$\pm$0.28 & 17.191$\pm$0.34 & 18.937$\pm$0.36 \\
 & \textbf{SINE} & \textbf{0.316$\pm$0.02} & 0.958$\pm$0.04 & 4.461$\pm$0.12 & 6.022$\pm$0.12 & 6.689$\pm$0.11 & 7.366$\pm$0.17 & 8.568$\pm$0.17 & 9.248$\pm$0.17 \\
 & \textbf{WAVEF} & \textbf{1.292$\pm$0.04} & 2.134$\pm$0.08 & 7.374$\pm$0.18 & 9.522$\pm$0.10 & 11.076$\pm$0.31 & 12.315$\pm$0.25 & 14.139$\pm$0.30 & 15.389$\pm$0.36 \\
\midrule 
 & \textbf{AGRAW$_{1}$} & \textbf{1.424$\pm$0.04} & 9.735$\pm$0.76 & 15.937$\pm$0.18 & 19.482$\pm$0.22 & 22.216$\pm$0.30 & 24.720$\pm$0.39 & 29.291$\pm$0.36 & 32.002$\pm$0.47 \\
 & \textbf{AGRAW$_{2}$} & \textbf{1.424$\pm$0.04} & 6.230$\pm$0.74 & 16.006$\pm$0.21 & 19.788$\pm$0.24 & 22.415$\pm$0.23 & 24.967$\pm$0.36 & 29.376$\pm$0.33 & 31.828$\pm$0.39 \\
GRAD. & \textbf{LED} & \textbf{4.155$\pm$0.15} & 4.816$\pm$0.11 & 27.394$\pm$0.27 & 34.001$\pm$0.32 & 40.239$\pm$0.44 & 44.755$\pm$0.43 & 52.791$\pm$0.48 & 58.284$\pm$0.47 \\
50K & \textbf{MIXED} & \textbf{0.670$\pm$0.03} & 2.424$\pm$0.13 & 11.673$\pm$0.22 & 15.692$\pm$0.25 & 17.702$\pm$0.20 & 19.226$\pm$0.46 & 23.995$\pm$0.30 & 26.306$\pm$0.31 \\
 & \textbf{RandRBF} & \textbf{7.488$\pm$0.14} & 10.396$\pm$0.13 & 24.794$\pm$0.25 & 30.630$\pm$0.43 & 36.039$\pm$0.49 & 40.068$\pm$0.60 & 47.021$\pm$0.57 & 53.383$\pm$0.72 \\
 & \textbf{SINE} & \textbf{0.764$\pm$0.04} & 3.520$\pm$0.10 & 11.956$\pm$0.24 & 15.988$\pm$0.21 & 18.231$\pm$0.26 & 20.339$\pm$0.30 & 24.569$\pm$0.28 & 26.689$\pm$0.27 \\
 & \textbf{WAVEF} & \textbf{3.085$\pm$0.10} & 6.816$\pm$0.51 & 18.754$\pm$0.17 & 24.354$\pm$0.27 & 27.915$\pm$0.34 & 31.753$\pm$0.38 & 38.247$\pm$0.47 & 41.980$\pm$0.48 \\
\midrule 
 & \textbf{AGRAW$_{1}$} & \textbf{2.790$\pm$0.06} & 30.644$\pm$1.29 & 32.532$\pm$0.22 & 39.650$\pm$0.34 & 45.750$\pm$0.38 & 50.767$\pm$0.57 & 60.108$\pm$0.55 & 66.468$\pm$0.72 \\
 & \textbf{AGRAW$_{2}$} & \textbf{2.696$\pm$0.10} & 23.296$\pm$2.44 & 32.247$\pm$0.29 & 39.704$\pm$0.38 & 45.680$\pm$0.46 & 51.330$\pm$0.52 & 61.069$\pm$0.57 & 66.988$\pm$0.57 \\
GRAD. & \textbf{LED} & \textbf{8.308$\pm$0.13} & 9.760$\pm$0.12 & 55.273$\pm$0.42 & 69.616$\pm$0.42 & 82.692$\pm$0.84 & 95.814$\pm$0.71 & 110.338$\pm$0.88 & 124.212$\pm$0.76 \\
100K & \textbf{MIXED} & \textbf{1.315$\pm$0.05} & 7.067$\pm$0.27 & 23.636$\pm$0.27 & 31.987$\pm$0.33 & 36.814$\pm$0.28 & 40.673$\pm$0.44 & 49.186$\pm$0.46 & 55.031$\pm$0.38 \\
 & \textbf{RandRBF} & \textbf{14.953$\pm$0.31} & 22.444$\pm$0.56 & 51.439$\pm$0.46 & 62.943$\pm$0.52 & 75.024$\pm$0.92 & 82.600$\pm$0.79 & 99.216$\pm$0.79 & 111.968$\pm$0.94 \\
 & \textbf{SINE} & \textbf{1.590$\pm$0.04} & 10.438$\pm$0.22 & 24.554$\pm$0.23 & 33.403$\pm$0.35 & 38.437$\pm$0.38 & 42.722$\pm$0.32 & 49.901$\pm$0.40 & 57.058$\pm$0.39 \\
 & \textbf{WAVEF} & \textbf{6.370$\pm$0.14} & 16.845$\pm$1.33 & 38.388$\pm$0.37 & 49.577$\pm$0.36 & 59.497$\pm$0.43 & 65.984$\pm$0.60 & 77.419$\pm$0.59 & 88.446$\pm$0.70 \\
\midrule 
 & \textbf{AGRAW$_{1}$} & \textbf{14.481$\pm$0.40} & 257.248$\pm$18.51 & 166.325$\pm$1.15 & 200.955$\pm$1.63 & 231.319$\pm$1.14 & 258.464$\pm$2.09 & 307.368$\pm$3.10 & 341.592$\pm$2.77 \\
 & \textbf{AGRAW$_{2}$} & \textbf{14.326$\pm$0.43} & 288.189$\pm$35.27 & 165.315$\pm$1.67 & 198.840$\pm$1.30 & 231.394$\pm$2.25 & 262.382$\pm$2.01 & 306.668$\pm$1.56 & 336.606$\pm$2.58 \\
GRAD. & \textbf{LED} & \textbf{42.103$\pm$0.86} & 51.246$\pm$0.54 & 281.936$\pm$2.06 & 349.085$\pm$2.87 & 419.025$\pm$3.66 & 481.918$\pm$3.92 & 554.435$\pm$4.80 & 627.611$\pm$7.39 \\
500K & \textbf{MIXED} & \textbf{6.869$\pm$0.15} & 67.277$\pm$4.85 & 121.244$\pm$1.03 & 166.205$\pm$1.18 & 190.021$\pm$1.25 & 213.245$\pm$1.33 & 255.349$\pm$2.21 & 285.847$\pm$1.67 \\
 & \textbf{RandRBF} & \textbf{75.476$\pm$1.86} & 174.009$\pm$21.76 & 256.474$\pm$2.82 & 313.074$\pm$2.77 & 370.087$\pm$3.81 & 426.301$\pm$5.86 & 501.879$\pm$5.38 & 557.291$\pm$5.48 \\
 & \textbf{SINE} & \textbf{8.028$\pm$0.30} & 115.575$\pm$7.05 & 126.030$\pm$1.44 & 167.559$\pm$1.05 & 194.181$\pm$1.54 & 219.377$\pm$1.75 & 260.356$\pm$1.34 & 290.814$\pm$1.96 \\
 & \textbf{WAVEF} & \textbf{33.021$\pm$0.39} & 116.076$\pm$26.57 & 189.763$\pm$1.65 & 249.983$\pm$1.98 & 294.085$\pm$2.68 & 332.151$\pm$3.75 & 394.933$\pm$2.72 & 438.278$\pm$5.27 \\
\midrule 
 & \textbf{AGRAW$_{1}$} & \textbf{29.030$\pm$0.69} & 577.423$\pm$37.19 & 332.974$\pm$3.19 & 404.563$\pm$1.39 & 464.358$\pm$3.46 & 529.960$\pm$9.82 & 612.587$\pm$4.03 & 680.089$\pm$5.41 \\
 & \textbf{AGRAW$_{2}$} & \textbf{28.928$\pm$0.53} & 640.909$\pm$95.79 & 328.368$\pm$3.24 & 405.818$\pm$1.24 & 466.032$\pm$3.03 & 533.479$\pm$4.91 & 608.133$\pm$1.92 & 657.908$\pm$8.62 \\
GRAD. & \textbf{LED} & \textbf{85.729$\pm$2.25} & 101.050$\pm$0.77 & 568.420$\pm$3.08 & 711.131$\pm$6.21 & 824.748$\pm$8.28 & 971.181$\pm$9.38 & 1082.603$\pm$7.46 & 1210.083$\pm$13.57 \\
1M & \textbf{MIXED} & \textbf{13.690$\pm$0.16} & 156.387$\pm$16.55 & 247.224$\pm$1.02 & 327.919$\pm$2.99 & 378.099$\pm$3.27 & 432.098$\pm$5.37 & 510.037$\pm$3.37 & 568.151$\pm$3.34 \\
 & \textbf{RandRBF} & \textbf{150.971$\pm$2.64} & 442.408$\pm$64.02 & 506.234$\pm$3.32 & 640.992$\pm$5.58 & 746.090$\pm$9.97 & 868.162$\pm$11.41 & 998.247$\pm$11.78 & 1084.871$\pm$19.46 \\
 & \textbf{SINE} & \textbf{16.254$\pm$0.39} & 264.026$\pm$14.01 & 252.286$\pm$1.49 & 343.519$\pm$3.15 & 387.452$\pm$1.94 & 444.156$\pm$4.77 & 527.466$\pm$2.69 & 580.101$\pm$3.43 \\
 & \textbf{WAVEF} & \textbf{65.128$\pm$1.18} & 266.549$\pm$40.59 & 378.101$\pm$2.95 & 494.582$\pm$3.28 & 585.655$\pm$4.64 & 672.894$\pm$6.54 & 779.516$\pm$7.10 & 862.580$\pm$9.61 \\
\midrule 
 & \textbf{AGRAW$_{1}$} & \textbf{56.290$\pm$2.30} & 1269.585$\pm$103.37 & 649.787$\pm$3.72 & 806.505$\pm$2.70 & 926.528$\pm$10.17 & 1064.057$\pm$10.06 & 1223.623$\pm$9.30 & 1341.320$\pm$12.45 \\
 & \textbf{AGRAW$_{2}$} & \textbf{56.638$\pm$2.39} & 1727.610$\pm$585.69 & 654.795$\pm$4.54 & 806.453$\pm$4.95 & 920.115$\pm$10.46 & 1066.953$\pm$7.54 & 1201.898$\pm$6.80 & 1297.267$\pm$12.29 \\
GRAD. & \textbf{LED} & \textbf{168.232$\pm$2.84} & 209.991$\pm$11.74 & 1115.579$\pm$7.19 & 1404.268$\pm$13.15 & 1646.879$\pm$11.57 & 1976.008$\pm$13.89 & 2172.260$\pm$22.52 & 2088.124$\pm$401.05 \\
2M & \textbf{MIXED} & \textbf{27.330$\pm$0.70} & 338.511$\pm$28.79 & 482.268$\pm$2.74 & 663.455$\pm$3.74 & 752.071$\pm$7.67 & 869.063$\pm$7.50 & 1022.426$\pm$7.20 & 1095.822$\pm$14.85 \\
 & \textbf{RandRBF} & \textbf{302.316$\pm$4.12} & 1224.013$\pm$400.43 & 1018.492$\pm$10.30 & 1265.647$\pm$13.60 & 1485.799$\pm$17.17 & 1730.521$\pm$18.83 & 1976.657$\pm$32.54 & 1799.438$\pm$375.37 \\
 & \textbf{SINE} & \textbf{32.172$\pm$0.46} & 566.160$\pm$26.68 & 499.153$\pm$3.31 & 676.171$\pm$4.93 & 768.053$\pm$6.53 & 905.905$\pm$5.53 & 1031.987$\pm$5.96 & 1143.559$\pm$11.89 \\
 & \textbf{WAVEF} & \textbf{129.694$\pm$2.52} & 495.468$\pm$60.19 & 757.090$\pm$3.94 & 1011.892$\pm$3.93 & 1159.754$\pm$5.99 & 1343.119$\pm$10.20 & 1554.748$\pm$10.70 & 1572.563$\pm$155.40 \\
\bottomrule 
\end{tabular} 
\end{adjustbox} 
\end{table}

Following, we compare the run-time consumption of \gls{knn} {\em with} and {\em without} using \gls{rddm} to detect concept drifts. 
As discussed earlier, the classification operation is potentially performed much more frequently with a detector, and this ought to directly affect the run-time.
Besides, as $w$ is the variable with the most significant influence on time classification, its enlargement is likely to cause more impact on the run-time of \gls{knn} when accompanied by \gls{rddm}.

\begin{figure}
\vspace{2mm}
\centering
\includegraphics[width=0.8\linewidth]{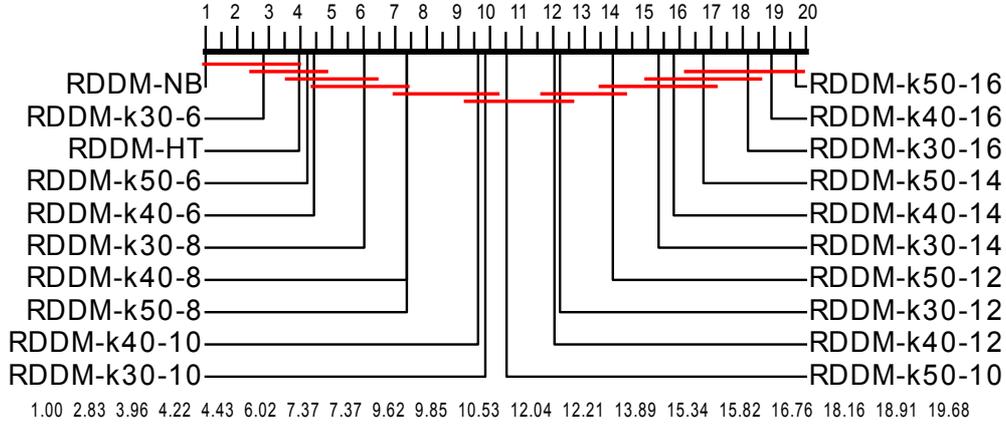}
\vspace{-15mm}
\caption{Comparison of the run-time results of classifiers with the RDDM detector in all datasets with 95\% confidence.}
\label{fig:f2-RDDM-Time-All}
\vspace{-2mm}
\end{figure}

Once again, the presentation of the results is complemented with an evaluation using the $F_F$ statistic and the Nemenyi post-hoc test \cite{demsar:2006}. 
Although the results without concept drift detection were also assessed, this time, only the results using \gls{rddm} to detect concept drifts are presented because these results are reasonably similar.
The main differences are the relative position of \gls{ht}, which is lower in the evaluation without drift detection, and a few inversions in the order of \gls{knn} versions with the same $w$.
Figure \ref{fig:f2-RDDM-Time-All} clearly demonstrates that the run-time of \gls{knn} is primarily determined by the window size used: the larger it is, the longer \gls{knn} will take to perform the classifications.

\subsection{Answers to RQ5--RQ8} \label{rq5to8}

To conclude this section, we address more research questions. 
The description of \textbf{RQ5} was: \textit{What is the impact of the window size ($w$) on the accuracy of \gls{knn}?}
The analysis allows the determination that no specific value of $w$ overcomes others in all scenarios.
In general, $w$ in the $[1200, 1600]$ interval yielded the best results. 
Nevertheless, specific generators were better classified with smaller windows, e.g., $w \in \{600, 800\}$ achieved the best results in the Agrawal datasets.
Furthermore, the volatility rates, i.e., the differences between the best and worst ranked $w$ values in terms of accuracy, showed that differences up to 6\% can be found and, thus, we recommend that specific assessment on this hyper-parameter should be based on experimentation with a sample of the data stream to be classified.

Next, we have \textbf{RQ6}: \textit{How does drift detection affect the accuracy of \gls{knn} given different $w$ values?}
Similar to the previous research question, the addition of drift detection to the process, represented by \gls{rddm}, did not change the overall trend in the results. 
Again, no window size was clearly better than the others statistically, and all values of $w \geq 1200$ fell within the confidence interval.
Finally, a volatility analysis also showed that the differences between the best and worst ranked values of $w$ reached 5\% and, thus, in the association of \gls{knn} with a concept drift detector, it is still advisable to adopt some specific assessment of different window sizes.

Proceeding to the research questions regarding run-time, we move on to \textbf{RQ7} and repeat its description: \textit{What is the impact of the window size ($w$) on \gls{knn}'s run-time?}
The results of the experiments confirmed the theoretical assessment of the time complexity of \gls{knn}, more precisely, that $w$ has a strong influence on the run-time of \gls{knn}.
Accordingly, choosing very high values for $w$ may prevent the practical use of \gls{knn} as a classifier for data streams and, thus, this parameter must be carefully defined depending on the actual restrictions of each specific problem.

Finally, the definition of \textbf{RQ8} was: \textit{How does concept drift detection impact the run-time of \gls{knn} for different values of $w$?}
Before explicitly answering this research question, we remind the reader that $w$ is the most critical variable in the classification time complexity of \gls{knn} and the use of a concept drift detector is likely to cause the classification operation to be requested much more frequently.
Thus, it is possible to deduce that higher values for $w$, especially with the adoption of a concept drift detector, are likely to increase the run-time of \gls{knn} considerably.
Hence, we emphatically recommend that the adoption of larger windows sizes in this scenario should be carefully evaluated based on experimentation considering the actual restrictions of the data stream.

\section{Comparison of \gls{knn} to Naïve Bayes and Hoeffding Tree}\label{comp:nb:ht}

In this section we target the comparison of different parametrizations of \gls{knn} against \gls{nb} and \gls{ht} based on the results previously depicted in Sections \ref{results:k} and \ref{results:w}.
In the accuracy analysis, we only refer to the results discussed in Section \ref{sec:results:w:acc}, where different window size values $w$ were assessed in conjunction with different neighborhood sizes $k$, because these encompass the results discussed in Section \ref{sec:results:k:acc} as well.

Firstly focusing on the abrupt and gradual dataset experiments with no drift detector, the results in Tables \ref{tab:det-acc-NoDet-Abr40}, \ref{tab:det-acc-NoDet-Grad40}, and \ref{tab:det-acc-NoDet-Abr30}--\ref{tab:det-acc-NoDet-Grad50} show that \gls{knn} achieves the best accuracy rates in nearly all the datasets. 
The exceptions are the larger RandRBF datasets, in both abrupt and gradual datasets, which are dominated by \gls{ht}.
Consequently, the chart presented in Figure \ref{fig:Acc-w-NoDet-Acc-All}, which captures the statistical results where different combinations of $k$ and $w$ were assessed, shows that the best versions of \gls{knn} were significantly better than \gls{ht} and that all \gls{knn} versions were statistically superior to \gls{nb}.
The fact that \gls{knn} uses only the last $w$ instances of the stream to perform the classification makes it naturally adapt to new probability distributions.
On the other hand, \gls{nb} and \gls{ht} take into account all previously seen instances for classification purposes and may have their predictions compromised by outdated data in drift scenarios.

Now, proceeding to the accuracy results of the tests using \gls{rddm} as drift detector in the abrupt and gradual dataset experiments, we refer to the results given in Tables \ref{tab:det-acc-RDDM-Abr40}, \ref{tab:det-acc-RDDM-Grad40} and \ref{tab:det-acc-RDDM-Abr30}--\ref{tab:det-acc-RDDM-Grad50}.
These results show that some configurations of \gls{knn} are still able to achieve superior results in several experiments but, comparatively, \gls{knn} does {\em not} benefit from drift detection as much as \gls{nb} and \gls{ht}, especially in the gradual drift datasets.
As a result, \gls{nb} achieves the best accuracy in the LED experiments, while \gls{ht} becomes competitive in a larger number of datasets.
Consequently, the critical difference analysis depicted in Figure \ref{fig:Acc-w-RDDM-Acc-All} shows that the relative rankings of \gls{nb} and \gls{ht} became higher than those in Figure \ref{fig:Acc-w-NoDet-Acc-All}.
Moreover, although five versions of \gls{knn} still achieved the best ranks, none of them was significantly better than \gls{ht}, and several \gls{knn} versions were statistically indistinguishable from \gls{nb}.

We conclude this accuracy analysis emphasizing these results are relevant because \gls{nb} and \gls{ht} are commonly considered the {\em off-the-shelf} solutions for data stream mining, while \gls{knn} is roughly overlooked, possibly due to its expected intensive computational time.

Changing the focus to run-time, we refer to the detailed results included in the run-time tables presented throughout the article.
A relevant and challenging to miss result is the fact that the run-time of \gls{nb} was the best in {\em nearly all scenarios}, irrespective of the use of drift detection, which is a consequence of its simplicity.
Despite being less efficient than \gls{knn} in training, which has constant time complexity, this difference is overcome in the classification.
As discussed in Subsection \ref{survey:nb}, \gls{nb} only depends on the number of classes $c$ and attributes of the dataset $d$ whereas \gls{knn} depends on the number of attributes $d$, and also on the number of closest neighbors $k$ and the size of the instance window $w$.

Regarding the run-times of \gls{ht}, on the other hand, they may or may not be better than those of \gls{knn}, depending on the scenario.
According to the experiments, the most decisive factor for \gls{ht} to have good performance is the size of the dataset, with its relative efficiency dropping as the size of the datasets grow. 
This happens because \gls{ht} has some dependency on the number of instances already processed in the stream and are continuously grown as new training instances are made available to the classifier, a behavior also observed in \cite{barddal_reg_trees1:2019}, though there are parameters in \gls{ht} that limit its growth to a certain point, avoiding the risk of making its memory and run-time consumption impracticable for data streams.
In addition, both for training and classification, it is necessary to traverse the tree from its root to the leaf node, an operation that depends on the tree's height and tends to increase with more instances.

According to the results of the experiments without drift detection, in scenarios containing up to 100 thousand instances, the run-times of \gls{ht} were generally better than those of \gls{knn} and, on the larger datasets, the opposite was true. 
When using \gls{rddm} to detect drifts, these findings were also true when \gls{knn} was configured with smaller values of $w$ (600 and 800); with larger values of $w$, \gls{knn} was generally slower than \gls{ht} in most datasets.

\subsection{Answers to RQ9 and RQ10}

After all the explanations above, we can now answer the remaining two research questions.
The first of them, \textbf{RQ9}, was stated as follows: \textit{How does \gls{knn} compare to \gls{nb} and \gls{ht} in terms of accuracy?}

The answer to \textbf{RQ9} is: across the multiple experiments conducted, it has been observed that most configurations of \gls{knn} were statistically superior to \gls{nb}, irrespective of the use of drift detection, and also significantly better than \gls{ht} when no drift detector was used. 
On the other hand, when they were associated with the \gls{rddm} drift detector, we observed that \gls{ht} and the best-ranked versions of \gls{knn} were statistically indistinguishable. However, five versions of \gls{knn} still achieved the top ranks considering all the 49 tested datasets.

Finally, \textbf{RQ10} was defined as: \textit{How does \gls{knn} compare to \gls{nb} and \gls{ht} regarding run-time?}
The experiments confirmed that \gls{nb} is the fastest of the tested classifiers, achieving the best result in nearly all datasets. 
\gls{knn} is faster during training, but it becomes considerably less efficient in the classification.
In relation to \gls{ht}, some versions of \gls{knn} were significantly better than \gls{ht} in the tests without drift detection.
In addition, when using \gls{rddm}, the run-time of \gls{ht} was statistically superior to the best-ranked \gls{knn} versions in accuracy and similar to some of the lower-ranked ones, which is due to the model resets that occur over time.
Consequently, the use of \gls{ht}s without drift detectors must be carefully analyzed if the model is trained \textit{ad eternum}.
Furthermore, \gls{knn} becomes an interesting choice as its complexity is constant regardless of how many instances are used for training.

\section{Conclusions and Future Work}\label{conc}

This article presents an in-depth evaluation of the suitability of \gls{knn} for classifying data streams with concept drifts.
For this purpose, \gls{knn} was compared to \acrfull{nb} and \acrfull{ht}, the most commonly used classifiers in the area.
In particular, the time complexity of the classifiers was also analyzed, and a large set of experiments was carried out to assess the accuracy and run-time performance of \gls{knn} with different parameter configurations and how they compare to those of \gls{nb} and \gls{ht}.
More specifically, we introduced and answered ten research questions.

According to our evaluation of the experiment results, we defend that \gls{knn} proved to be a suitable option to consider.
Among its advantages is that it naturally adapts to concept drifts since it only uses the last $w$ instances of the stream for predictions.
Consequently, when the different learners were compared without drift detection, \gls{knn} was statistically superior to the other methods, regardless of the adopted parameter values.
Numerically speaking, \gls{knn} overcame \gls{nb} in 100\% of the experiments and \gls{ht} in 94\% of the scenarios tested. Furthermore, the average accuracy differences between \gls{knn} and \gls{ht} was of 10.18\% and \gls{knn} and \gls{nb} was of 16.27\%.
In addition, when the classifiers were configured with the \gls{rddm} detector, representing the drift detection scenario, \gls{knn} using $w \geq$ 1400 with $k$ between 30 and 50 achieved better accuracy ranks than \gls{ht} and were significantly better than \gls{nb}.

On the other hand, \gls{knn} also has limitations.
One of its major drawbacks is the deterioration of the run-time when configured with larger window sizes and/or when a drift detector is attached.
We observed a clear trade-off between accuracy and run-time in the configuration of the \gls{knn} parameters, especially the window size $w$. 
In other words, the experiments demonstrated that, even though windows with 1400 instances or more do increase the likelihood of achieving higher accuracy rates, the run-time consumption also grows considerably.

Additionally, it is important to note that scenarios with many attributes can also affect the performance of \gls{knn} in terms of accuracy, processing time, and memory consumption \cite{kouiroukidis:2011}. 
Therefore, when considering using \gls{knn} to classify data streams, all these factors must be carefully analyzed, accounting for the environment's restrictions. 
Despite conducting all the experiments using artificial scenarios, its configurations were comprehensive enough to contemplate most of the characteristics of real-world datasets found in practice. 
Therefore, it is feasible to assume that the discussions presented throughout this article apply in different contexts.

Although this study broadly covers \gls{knn} in data streams, some limitations exist and must be enumerated: these could also be seen as possibilities of future work. 
The first concerns using only the Euclidean distance: other computationally more efficient alternatives could be beneficial to the \gls{knn} classifier in some way. 
Another limitation refers to the scenario used, i.e., supervised classification: exploring other contexts would yield different information.

As other directions for future works, we envisage the assessment of \gls{knn}'s behavior when it is applied to different dimensionality and coupled with feature selection algorithms \cite{barddal_abfs:2019}. 
Another possibly worthwhile investigation, considering that \gls{nb} was significantly faster but normally less accurate than both \gls{knn} and \gls{ht}, is an in-depth investigation of how different configurations of ensembles of \gls{nb} classifiers perform against single \gls{knn} and \gls{ht} classifiers.

\section*{Acknowledgments}
Roberto S.~M.~Barros is supported by research grant number 310092/2019-1 from CNPq, and Silas G.~T.~C.~Santos is supported by post-doctorate grant number 88887.374884/2019-00 from CAPES.

\bibliography{refs}

\newpage

\setcounter{page}{1}
 
\section*{Appendix}
All the accuracy and run-time tables of results omitted from the main body of the article are provided here in the appendix.
Nevertheless, this appendix can also be seen as complementary material.

\begin{table} 
\caption{Mean accuracies of the classifiers with no drift detector, in the gradual datasets, in percentage (\%), with a 95\% confidence interval (Part~1)} 
\label{tab:det-acc-NoDet-Grad1} 
\begin{adjustbox}{max width=\textwidth} 
\begin{tabular}{cccccccc} 
\toprule 
DS Type & \textbf{NO DETECTOR} & NB & kNN5 & kNN10 & kNN15 & kNN20 & kNN25 \\ 
and Size & DATASET &  HT & kNN30 & kNN35 & kNN40 & kNN45 & kNN50 \\ 
\toprule 
 & \textbf{AGRAW$_{1}$} & 57.15691$\pm$0.19 & 61.75522$\pm$0.24 & 60.98401$\pm$0.19 & 61.30536$\pm$0.21 & 61.59338$\pm$0.19 & 63.19000$\pm$0.22 \\  &  & 58.69050$\pm$0.71 & 64.41822$\pm$0.19 & 65.40591$\pm$0.16 & 66.08142$\pm$0.16 & 66.13313$\pm$0.17 & \textbf{66.33262$\pm$0.18} \\  \\[-1.72mm]
 & \textbf{AGRAW$_{2}$} & 67.54421$\pm$0.28 & 72.59629$\pm$0.18 & 73.02791$\pm$0.18 & 73.14080$\pm$0.23 & 73.68921$\pm$0.22 & 74.38014$\pm$0.26 \\  &  & 69.44918$\pm$0.43 & 75.56940$\pm$0.23 & 76.37001$\pm$0.23 & 77.05881$\pm$0.24 & 77.26907$\pm$0.22 & \textbf{77.56103$\pm$0.22} \\  \\
\bottomrule 
\end{tabular} 
\end{adjustbox} 
\end{table}

\begin{table} 
\caption{Mean accuracies of the classifiers with no drift detector, in the gradual datasets, in percentage (\%), with a 95\% confidence interval (Part~2)} 
\label{tab:det-acc-NoDet-Grad2} 
\begin{adjustbox}{max width=\textwidth} 
\begin{tabular}{cccccccc} 
\toprule 
DS Type & \textbf{NO DETECTOR} & NB & kNN5 & kNN10 & kNN15 & kNN20 & kNN25 \\ 
and Size & DATASET &  HT & kNN30 & kNN35 & kNN40 & kNN45 & kNN50 \\ 
\toprule 
 & \textbf{AGRAW$_{1}$} & 57.69231$\pm$0.05 & 65.50224$\pm$0.07 & 63.76845$\pm$0.06 & 63.85355$\pm$0.07 & 64.11151$\pm$0.07 & 65.85048$\pm$0.06 \\  &  & 59.53123$\pm$0.63 & 67.55707$\pm$0.05 & 68.64163$\pm$0.04 & 69.69023$\pm$0.04 & 69.66239$\pm$0.04 & \textbf{69.98244$\pm$0.04} \\  \\[-1.72mm]
 & \textbf{AGRAW$_{2}$} & 70.41979$\pm$0.05 & 81.72446$\pm$0.05 & 81.14441$\pm$0.06 & 81.31200$\pm$0.05 & 81.25157$\pm$0.05 & 82.16218$\pm$0.05 \\  &  & 72.84759$\pm$0.49 & 83.37217$\pm$0.06 & 84.67848$\pm$0.05 & 85.52473$\pm$0.05 & 86.04447$\pm$0.05 & \textbf{86.31104$\pm$0.05} \\  \\
\bottomrule 
\end{tabular} 
\end{adjustbox} 
\end{table}

\begin{table} 
\caption{Mean accuracies of classifiers with the RDDM detector, in the gradual datasets, in percentage (\%), with a 95\% confidence interval (Part~1)} 
\label{tab:det-acc-RDDM-Grad1} 
\begin{adjustbox}{max width=\textwidth} 
\begin{tabular}{cccccccc} 
\toprule 
DS Type & \textbf{RDDM} & NB & kNN5 & kNN10 & kNN15 & kNN20 & kNN25 \\ 
and Size & DATASET &  HT & kNN30 & kNN35 & kNN40 & kNN45 & kNN50 \\ 
\toprule 
 & \textbf{AGRAW$_{1}$} & 62.05290$\pm$0.29 & 61.68505$\pm$0.23 & 61.54286$\pm$0.23 & 62.35493$\pm$0.23 & 63.14568$\pm$0.25 & 64.16449$\pm$0.20 \\  &  & 62.92415$\pm$0.28 & 65.23403$\pm$0.20 & 65.82949$\pm$0.17 & \textbf{66.23798$\pm$0.17} & 66.17342$\pm$0.17 & 66.16596$\pm$0.19 \\  \\[-1.72mm]
 & \textbf{AGRAW$_{2}$} & 74.29287$\pm$1.50 & 73.51382$\pm$0.25 & 73.73693$\pm$0.22 & 74.62410$\pm$0.33 & 75.30997$\pm$0.30 & 76.31426$\pm$0.30 \\  &  & 78.65467$\pm$1.02 & 77.42512$\pm$0.27 & 78.31431$\pm$0.27 & 78.85404$\pm$0.27 & 79.09180$\pm$0.25 & \textbf{79.32209$\pm$0.24} \\  \\
\bottomrule 
\end{tabular} 
\end{adjustbox} 
\end{table}

\begin{table} 
\caption{Mean accuracies of classifiers with the RDDM detector, in the gradual datasets, in percentage (\%), with a 95\% confidence interval (Part~2)} 
\label{tab:det-acc-RDDM-Grad2} 
\begin{adjustbox}{max width=\textwidth} 
\begin{tabular}{cccccccc} 
\toprule 
DS Type & \textbf{RDDM} & NB & kNN5 & kNN10 & kNN15 & kNN20 & kNN25 \\ 
and Size & DATASET &  HT & kNN30 & kNN35 & kNN40 & kNN45 & kNN50 \\ 
\toprule 
 & \textbf{AGRAW$_{1}$} & 66.36269$\pm$0.05 & 65.45609$\pm$0.07 & 63.76528$\pm$0.06 & 63.93425$\pm$0.09 & 64.21672$\pm$0.09 & 65.87483$\pm$0.06 \\  &  & \textbf{78.12037$\pm$0.97} & 67.55184$\pm$0.05 & 68.61143$\pm$0.05 & 69.64152$\pm$0.04 & 69.60861$\pm$0.04 & 69.91641$\pm$0.05 \\  \\[-1.72mm]
 & \textbf{AGRAW$_{2}$} & 86.66851$\pm$0.08 & 81.71481$\pm$0.06 & 81.13637$\pm$0.07 & 81.33306$\pm$0.05 & 81.32460$\pm$0.06 & 82.27954$\pm$0.07 \\  &  & \textbf{88.74043$\pm$0.20} & 83.41904$\pm$0.06 & 84.70543$\pm$0.05 & 85.54518$\pm$0.05 & 86.06146$\pm$0.05 & 86.32210$\pm$0.05 \\  \\
\bottomrule 
\end{tabular} 
\end{adjustbox} 
\end{table}

\begin{table} 
\caption{Mean run-time of the classifiers with no drift detector, in the abrupt datasets, in seconds, with a 95\% confidence interval (Part~1)} 
\label{tab:det-time-NoDet-Abr1} 
\begin{adjustbox}{max width=\textwidth} 
\begin{tabular}{cccccccc} 
\toprule 
DS Type & \textbf{NO DETECTOR} & NB & kNN5 & kNN10 & kNN15 & kNN20 & kNN25 \\ 
and Size & DATASET &  HT & kNN30 & kNN35 & kNN40 & kNN45 & kNN50 \\ 
\toprule 
 & \textbf{AGRAW$_{1}$} & \textbf{0.0982$\pm$0.01} & 0.9201$\pm$0.01 & 0.5287$\pm$0.01 & 0.9868$\pm$0.03 & 0.9659$\pm$0.05 & 1.0246$\pm$0.03 \\  &   & 0.1261$\pm$0.03 & 1.0304$\pm$0.04 & 1.0438$\pm$0.04 & 1.0377$\pm$0.03 & 1.0077$\pm$0.04 & 1.0223$\pm$0.04 \\  \\[-1.72mm] 
 & \textbf{AGRAW$_{2}$} & 0.0929$\pm$0.01 & 0.8626$\pm$0.04 & 0.5295$\pm$0.01 & 0.9744$\pm$0.04 & 1.0148$\pm$0.02 & 1.0325$\pm$0.03 \\  &   & \textbf{0.0795$\pm$0.02} & 1.0471$\pm$0.03 & 0.9821$\pm$0.04 & 1.0192$\pm$0.03 & 1.0324$\pm$0.03 & 1.0716$\pm$0.04 \\  \\
\bottomrule 
\end{tabular} 
\end{adjustbox} 
\end{table}

\begin{table} 
\caption{Mean run-time of the classifiers with no drift detector, in the abrupt datasets, in seconds, with a 95\% confidence interval (Part~2)}
\label{tab:det-time-NoDet-Abr2} 
\begin{adjustbox}{max width=\textwidth} 
\begin{tabular}{cccccccc} 
\toprule 
DS Type & \textbf{NO DETECTOR} & NB & kNN5 & kNN10 & kNN15 & kNN20 & kNN25 \\ 
and Size & DATASET &  HT & kNN30 & kNN35 & kNN40 & kNN45 & kNN50 \\ 
\toprule 
 & \textbf{AGRAW$_{1}$} & \textbf{3.5761$\pm$0.16} & 48.7839$\pm$1.19 & 50.3205$\pm$0.98 & 51.7470$\pm$1.31 & 54.8391$\pm$1.11 & 52.9043$\pm$1.14 \\  &   & 164.2418$\pm$8.54 & 57.2963$\pm$0.83 & 55.0329$\pm$0.66 & 55.1852$\pm$0.94 & 56.4761$\pm$0.96 & 55.3622$\pm$1.19 \\  \\[-1.72mm] 
 & \textbf{AGRAW$_{2}$} & \textbf{3.6301$\pm$0.18} & 49.1385$\pm$0.52 & 51.2435$\pm$0.79 & 52.6055$\pm$0.61 & 53.3026$\pm$0.94 & 53.9001$\pm$1.27 \\  &   & 184.0367$\pm$38.37 & 54.5673$\pm$0.93 & 53.7096$\pm$1.33 & 54.6432$\pm$1.08 & 55.4264$\pm$0.74 & 55.6916$\pm$1.27 \\  \\
\bottomrule 
\end{tabular} 
\end{adjustbox} 
\end{table}

\begin{table} 
\caption{Mean run-time of the classifiers with no drift detector, in the gradual datasets, in seconds, with a 95\% confidence interval (Part~1)}
\label{tab:det-time-NoDet-Grad1} 
\begin{adjustbox}{max width=\textwidth} 
\begin{tabular}{cccccccc} 
\toprule 
DS Type & \textbf{NO DETECTOR} & NB & kNN5 & kNN10 & kNN15 & kNN20 & kNN25 \\ 
and Size & DATASET &  HT & kNN30 & kNN35 & kNN40 & kNN45 & kNN50 \\ 
\toprule 
 & \textbf{AGRAW$_{1}$} & \textbf{0.0853$\pm$0.01} & 0.8920$\pm$0.02 & 0.8906$\pm$0.04 & 0.9147$\pm$0.06 & 1.0089$\pm$0.02 & 1.0026$\pm$0.05 \\  &   & 0.1244$\pm$0.03 & 1.0428$\pm$0.03 & 0.9819$\pm$0.04 & 1.0419$\pm$0.02 & 1.0174$\pm$0.05 & 1.0536$\pm$0.03 \\  \\[-1.72mm] 
 & \textbf{AGRAW$_{2}$} & 0.0853$\pm$0.01 & 0.8894$\pm$0.03 & 0.8908$\pm$0.04 & 0.9885$\pm$0.01 & 0.9718$\pm$0.04 & 1.0356$\pm$0.01 \\  &   & \textbf{0.0783$\pm$0.02} & 1.0331$\pm$0.03 & 1.0180$\pm$0.01 & 1.0419$\pm$0.02 & 1.0713$\pm$0.03 & 1.0760$\pm$0.01 \\  \\
\bottomrule 
\end{tabular} 
\end{adjustbox} 
\end{table}

\begin{table} 
\caption{Mean run-time of the classifiers with no drift detector, in the gradual datasets, in seconds, with a 95\% confidence interval (Part~2)} 
\label{tab:det-time-NoDet-Grad2} 
\begin{adjustbox}{max width=\textwidth} 
\begin{tabular}{cccccccc} 
\toprule 
DS Type & \textbf{NO DETECTOR} & NB & kNN5 & kNN10 & kNN15 & kNN20 & kNN25 \\ 
and Size & DATASET &  HT & kNN30 & kNN35 & kNN40 & kNN45 & kNN50 \\ 
\toprule 
 & \textbf{AGRAW$_{1}$} & \textbf{3.8057$\pm$0.18} & 48.6762$\pm$0.84 & 50.1668$\pm$0.98 & 51.9479$\pm$0.89 & 54.9040$\pm$1.25 & 54.7818$\pm$1.52 \\  &   & 165.4906$\pm$8.54 & 55.9238$\pm$0.53 & 54.3625$\pm$1.03 & 56.8617$\pm$0.84 & 55.2924$\pm$0.73 & 56.5447$\pm$1.41 \\  \\[-1.72mm] 
 & \textbf{AGRAW$_{2}$} & \textbf{3.6291$\pm$0.22} & 49.1040$\pm$0.65 & 49.7485$\pm$0.99 & 52.4315$\pm$0.88 & 53.3627$\pm$0.91 & 54.7895$\pm$1.53 \\  &   & 181.4916$\pm$37.79 & 56.0651$\pm$0.75 & 53.4233$\pm$1.12 & 54.2193$\pm$0.64 & 55.5962$\pm$0.83 & 55.8145$\pm$0.93 \\  \\
\bottomrule 
\end{tabular} 
\end{adjustbox} 
\end{table}

\begin{table} 
\caption{Mean run-time of the classifiers with the RDDM detector, in the abrupt datasets, in seconds, with a 95\% confidence interval (Part~1)} 
\label{tab:det-time-RDDM-Abr1} 
\begin{adjustbox}{max width=\textwidth} 
\begin{tabular}{cccccccc} 
\toprule 
DS Type & \textbf{RDDM} & NB & kNN5 & kNN10 & kNN15 & kNN20 & kNN25 \\ 
and Size & DATASET &  HT & kNN30 & kNN35 & kNN40 & kNN45 & kNN50 \\ 
\toprule  
 & \textbf{AGRAW$_{1}$} & \textbf{0.3186$\pm$0.02} & 3.7733$\pm$0.12 & 3.5994$\pm$0.12 & 3.6413$\pm$0.13 & 3.6038$\pm$0.13 & 3.7167$\pm$0.12 \\  &   & 0.5182$\pm$0.03 & 3.5987$\pm$0.12 & 3.4571$\pm$0.13 & 3.7932$\pm$0.13 & 3.6481$\pm$0.12 & 3.6217$\pm$0.15 \\  \\[-1.72mm] 
 & \textbf{AGRAW$_{2}$} & \textbf{0.2973$\pm$0.02} & 3.8177$\pm$0.11 & 3.9435$\pm$0.12 & 3.9253$\pm$0.11 & 3.7997$\pm$0.13 & 4.0747$\pm$0.13 \\  &   & 0.5767$\pm$0.03 & 4.0002$\pm$0.13 & 3.8090$\pm$0.12 & 4.1247$\pm$0.12 & 3.9516$\pm$0.13 & 3.9608$\pm$0.14 \\  \\
\bottomrule 
\end{tabular} 
\end{adjustbox} 
\end{table}

\begin{table} 
\caption{Mean run-time of the classifiers with the RDDM detector, in the abrupt datasets, in seconds, with a 95\% confidence interval (Part~2)} 
\label{tab:det-time-RDDM-Abr2} 
\begin{adjustbox}{max width=\textwidth} 
\begin{tabular}{cccccccc} 
\toprule 
DS Type & \textbf{RDDM} & NB & kNN5 & kNN10 & kNN15 & kNN20 & kNN25 \\ 
and Size & DATASET &  HT & kNN30 & kNN35 & kNN40 & kNN45 & kNN50 \\ 
\toprule  
 & \textbf{AGRAW$_{1}$} & \textbf{14.3855$\pm$0.45} & 230.0645$\pm$1.95 & 232.3154$\pm$1.91 & 232.1544$\pm$1.87 & 232.0193$\pm$2.89 & 237.3415$\pm$2.59 \\  &   & 267.2791$\pm$10.18 & 235.6440$\pm$2.17 & 226.4760$\pm$2.10 & 231.1928$\pm$2.78 & 234.7085$\pm$1.74 & 237.5302$\pm$1.03 \\  \\[-1.72mm] 
 & \textbf{AGRAW$_{2}$} & \textbf{14.1827$\pm$0.42} & 227.1787$\pm$1.51 & 231.2469$\pm$1.40 & 230.1563$\pm$2.37 & 230.2414$\pm$2.35 & 235.3479$\pm$2.75 \\  &   & 310.2627$\pm$37.71 & 232.9152$\pm$1.39 & 224.7824$\pm$2.02 & 231.3684$\pm$2.92 & 235.5567$\pm$1.33 & 235.9290$\pm$1.80 \\  \\
\bottomrule 
\end{tabular} 
\end{adjustbox} 
\end{table}

\begin{table} 
\caption{Mean run-time of the classifiers with the RDDM detector, in the gradual datasets, in seconds, with a 95\% confidence interval (Part~1)} 
\label{tab:det-time-RDDM-Grad1} 
\begin{adjustbox}{max width=\textwidth} 
\begin{tabular}{cccccccc} 
\toprule 
DS Type & \textbf{RDDM} & NB & kNN5 & kNN10 & kNN15 & kNN20 & kNN25 \\ 
and Size & DATASET &  HT & kNN30 & kNN35 & kNN40 & kNN45 & kNN50 \\ 
\toprule  
 & \textbf{AGRAW$_{1}$} & \textbf{0.3143$\pm$0.02} & 3.7445$\pm$0.12 & 3.7835$\pm$0.13 & 3.8567$\pm$0.12 & 3.6229$\pm$0.12 & 3.8470$\pm$0.10 \\  &   & 0.4790$\pm$0.04 & 3.6665$\pm$0.12 & 3.6877$\pm$0.12 & 3.8225$\pm$0.12 & 3.7564$\pm$0.11 & 3.7396$\pm$0.13 \\  \\[-1.72mm] 
 & \textbf{AGRAW$_{2}$} & \textbf{0.3156$\pm$0.02} & 3.8318$\pm$0.13 & 4.0165$\pm$0.13 & 3.9865$\pm$0.13 & 3.9330$\pm$0.11 & 4.0935$\pm$0.13 \\  &   & 0.5378$\pm$0.03 & 4.0644$\pm$0.12 & 3.9251$\pm$0.13 & 4.1201$\pm$0.12 & 4.0727$\pm$0.11 & 4.0095$\pm$0.12 \\  \\
\bottomrule 
\end{tabular} 
\end{adjustbox} 
\end{table}

\begin{table} 
\caption{Mean run-time of the classifiers with the RDDM detector, in the gradual datasets, in seconds, with a 95\% confidence interval (Part~2)} 
\label{tab:det-time-RDDM-Grad2} 
\begin{adjustbox}{max width=\textwidth} 
\begin{tabular}{cccccccc} 
\toprule 
DS Type & \textbf{RDDM} & NB & kNN5 & kNN10 & kNN15 & kNN20 & kNN25 \\ 
and Size & DATASET &  HT & kNN30 & kNN35 & kNN40 & kNN45 & kNN50 \\ 
\toprule
 & \textbf{AGRAW$_{1}$} & \textbf{14.4805$\pm$0.40} & 229.7982$\pm$1.32 & 229.6450$\pm$1.95 & 227.3569$\pm$2.49 & 230.2809$\pm$1.61 & 237.1345$\pm$1.81 \\  &   & 257.2483$\pm$18.51 & 235.0347$\pm$2.26 & 232.4794$\pm$1.99 & 231.3189$\pm$1.14 & 233.6929$\pm$1.71 & 235.3278$\pm$1.30 \\  \\[-1.72mm] 
 & \textbf{AGRAW$_{2}$} & \textbf{14.3263$\pm$0.43} & 229.3661$\pm$1.73 & 230.4289$\pm$2.08 & 232.4347$\pm$3.45 & 230.2129$\pm$2.62 & 234.6813$\pm$1.87 \\  &   & 288.1889$\pm$35.27 & 234.1349$\pm$1.68 & 232.8291$\pm$2.18 & 231.3941$\pm$2.25 & 235.5673$\pm$2.31 & 234.2452$\pm$0.95 \\  \\
\bottomrule 
\end{tabular} 
\end{adjustbox} 
\end{table}

\begin{table} 
\caption{Mean accuracies of classifiers with no drift detector, in the abrupt datasets, in percentage (\%), with a 95\% confidence interval (k=30)} 
\label{tab:det-acc-NoDet-Abr30} 
\begin{adjustbox}{max width=\textwidth} 
\begin{tabular}{cccccccccc} 
\toprule 
DS Type  & \textbf{NO DET.} & \multicolumn{2}{c}{\textbf{CLASSIFIER}} & \multicolumn{6}{l}{kNN --- with k = 30, varying the size of the window (w)} \\ 
and Size & DATASET & NB & HT & w=600 & w=800 & w=1000 & w=1200 & w=1400 & w=1600 \\ 
\toprule 
 & \textbf{AGRAW$_{1}$} & 57.1202$\pm$0.18 & 58.7460$\pm$0.73 & \textbf{66.9738$\pm$0.16} & 66.2080$\pm$0.15 & 64.5865$\pm$0.19 & 62.9442$\pm$0.18 & 61.7068$\pm$0.19 & 60.9274$\pm$0.19 \\
 & \textbf{AGRAW$_{2}$} & 67.5743$\pm$0.30 & 69.9352$\pm$0.29 & \textbf{79.4833$\pm$0.20} & 77.9282$\pm$0.23 & 76.0549$\pm$0.23 & 74.4206$\pm$0.29 & 73.0559$\pm$0.29 & 71.9723$\pm$0.28 \\
ABRUPT &  \textbf{LED}  & 57.0178$\pm$0.25 & 55.8691$\pm$0.42 & 60.1646$\pm$0.41 & 61.0352$\pm$0.38 & \textbf{61.1341$\pm$0.35} & 60.9493$\pm$0.33 & 60.5651$\pm$0.32 & 60.0758$\pm$0.33 \\
10K  &  \textbf{MIXED}  & 57.6353$\pm$0.14 & 59.3308$\pm$0.40 & \textbf{82.3687$\pm$0.17} & 79.3725$\pm$0.18 & 75.9498$\pm$0.15 & 72.4408$\pm$0.17 & 68.8254$\pm$0.16 & 65.1320$\pm$0.15 \\
 &   \textbf{RandRBF}   & 30.8708$\pm$0.67 & 30.8309$\pm$0.67 & 30.0641$\pm$0.28 & 31.3897$\pm$0.28 & 32.5348$\pm$0.30 & 33.1462$\pm$0.32 & 33.8250$\pm$0.30 & \textbf{34.3637$\pm$0.33} \\
 &      \textbf{SINE}   & 56.3338$\pm$0.10 & 59.9403$\pm$0.49 & \textbf{81.3276$\pm$0.19} & 78.8508$\pm$0.20 & 76.2093$\pm$0.19 & 73.4115$\pm$0.19 & 70.4472$\pm$0.21 & 67.5238$\pm$0.24 \\
\bottomrule 
\end{tabular} 
\end{adjustbox} 
\end{table}

\begin{table} 
\caption{Mean accuracies of classifiers with no drift detector, in the gradual datasets, in percentage (\%), with a 95\% confidence interval (k=30)} 
\label{tab:det-acc-NoDet-Grad30} 
\begin{adjustbox}{max width=\textwidth} 
\begin{tabular}{cccccccccc} 
\toprule 
DS Type  & \textbf{NO DET.} & \multicolumn{2}{c}{\textbf{CLASSIFIER}} & \multicolumn{6}{l}{kNN --- with k = 30, varying the size of the window (w)} \\ 
and Size & DATASET & NB & HT & w=600 & w=800 & w=1000 & w=1200 & w=1400 & w=1600 \\ 
\toprule 
 & \textbf{AGRAW$_{1}$} & 57.1569$\pm$0.19 & 58.6905$\pm$0.71 & \textbf{66.5481$\pm$0.17} & 65.9383$\pm$0.15 & 64.4182$\pm$0.19 & 62.8614$\pm$0.19 & 61.5648$\pm$0.20 & 60.8480$\pm$0.21 \\
 & \textbf{AGRAW$_{2}$} & 67.5442$\pm$0.28 & 69.4492$\pm$0.43 & \textbf{78.6148$\pm$0.19} & 77.2397$\pm$0.22 & 75.5694$\pm$0.23 & 74.0157$\pm$0.29 & 72.7601$\pm$0.28 & 71.7408$\pm$0.28 \\
GRAD. &   \textbf{LED}  & 57.0161$\pm$0.25 & 55.8402$\pm$0.41 & 59.7426$\pm$0.39 & 60.7360$\pm$0.36 & \textbf{60.9442$\pm$0.36} & 60.7802$\pm$0.31 & 60.4453$\pm$0.31 & 59.9799$\pm$0.33 \\
10K  &  \textbf{MIXED}  & 57.6343$\pm$0.12 & 59.4987$\pm$0.38 & \textbf{81.2716$\pm$0.17} & 78.7588$\pm$0.16 & 75.6382$\pm$0.17 & 72.2428$\pm$0.16 & 68.7475$\pm$0.15 & 65.0872$\pm$0.14 \\
 &   \textbf{RandRBF}   & 30.9061$\pm$0.68 & 30.8662$\pm$0.68 & 29.9545$\pm$0.27 & 31.3417$\pm$0.27 & 32.4718$\pm$0.29 & 33.1711$\pm$0.31 & 33.8352$\pm$0.32 & \textbf{34.3860$\pm$0.32} \\
 &      \textbf{SINE}   & 56.3515$\pm$0.10 & 59.2024$\pm$0.28 & \textbf{80.2146$\pm$0.17} & 78.2331$\pm$0.19 & 75.8276$\pm$0.20 & 73.1260$\pm$0.18 & 70.2780$\pm$0.21 & 67.3588$\pm$0.23 \\
\bottomrule 
\end{tabular} 
\end{adjustbox} 
\end{table}

\begin{table} 
\caption{Mean accuracies of classifiers with no drift detector, in the abrupt datasets, in percentage (\%), with a 95\% confidence interval (k=50)} 
\label{tab:det-acc-NoDet-Abr50} 
\begin{adjustbox}{max width=\textwidth} 
\begin{tabular}{cccccccccc} 
\toprule 
DS Type  & \textbf{NO DET.} & \multicolumn{2}{c}{\textbf{CLASSIFIER}} & \multicolumn{6}{l}{kNN --- with k = 50, varying the size of the window (w)} \\ 
and Size & DATASET & NB & HT & w=600 & w=800 & w=1000 & w=1200 & w=1400 & w=1600 \\ 
\toprule 
 & \textbf{AGRAW$_{1}$} & 57.1202$\pm$0.18 & 58.7460$\pm$0.73 & 65.6666$\pm$0.16 & 66.4263$\pm$0.12 & \textbf{66.4633$\pm$0.18} & 65.8418$\pm$0.17 & 64.8472$\pm$0.18 & 63.7042$\pm$0.18 \\
 & \textbf{AGRAW$_{2}$} & 67.5743$\pm$0.30 & 69.9352$\pm$0.29 & \textbf{79.8480$\pm$0.21} & 79.0762$\pm$0.21 & 78.0439$\pm$0.24 & 76.7892$\pm$0.24 & 75.3499$\pm$0.25 & 73.9502$\pm$0.26 \\
ABRUPT &  \textbf{LED}  & 57.0178$\pm$0.25 & 55.8691$\pm$0.42 & 60.2788$\pm$0.34 & 61.3736$\pm$0.32 & \textbf{61.5976$\pm$0.31} & 61.4487$\pm$0.30 & 60.9678$\pm$0.30 & 60.5463$\pm$0.32 \\
10K  &  \textbf{MIXED}  & 57.6353$\pm$0.14 & 59.3308$\pm$0.40 & \textbf{80.4958$\pm$0.20} & 77.9741$\pm$0.18 & 74.8950$\pm$0.17 & 71.5925$\pm$0.17 & 68.0953$\pm$0.16 & 64.4154$\pm$0.15 \\
 &   \textbf{RandRBF}   & 30.8708$\pm$0.67 & 30.8309$\pm$0.67 & 27.8171$\pm$0.25 & 28.8699$\pm$0.34 & 29.8656$\pm$0.33 & 30.5096$\pm$0.28 & 31.0953$\pm$0.32 & \textbf{31.6292$\pm$0.33} \\
 &      \textbf{SINE}   & 56.3338$\pm$0.10 & 59.9403$\pm$0.49 & \textbf{80.2152$\pm$0.17} & 78.0083$\pm$0.19 & 75.4451$\pm$0.20 & 72.7026$\pm$0.17 & 69.9131$\pm$0.18 & 66.9283$\pm$0.18 \\
\bottomrule 
\end{tabular} 
\end{adjustbox} 
\end{table}

\begin{table} 
\caption{Mean accuracies of classifiers with no drift detector, in the gradual datasets, in percentage (\%), with a 95\% confidence interval (k=50)} 
\label{tab:det-acc-NoDet-Grad50} 
\begin{adjustbox}{max width=\textwidth} 
\begin{tabular}{cccccccccc} 
\toprule 
DS Type  & \textbf{NO DET.} & \multicolumn{2}{c}{\textbf{CLASSIFIER}} & \multicolumn{6}{l}{kNN --- with k = 50, varying the size of the window (w)} \\ 
and Size & DATASET & NB & HT & w=600 & w=800 & w=1000 & w=1200 & w=1400 & w=1600 \\ 
\toprule 
 & \textbf{AGRAW$_{1}$} & 57.1569$\pm$0.19 & 58.6905$\pm$0.71 & 65.3461$\pm$0.16 & 66.2581$\pm$0.15 & \textbf{66.3326$\pm$0.18} & 65.7915$\pm$0.17 & 64.8220$\pm$0.18 & 63.6696$\pm$0.18 \\
 & \textbf{AGRAW$_{2}$} & 67.5442$\pm$0.28 & 69.4492$\pm$0.43 & \textbf{78.9671$\pm$0.21} & 78.4793$\pm$0.21 & 77.5610$\pm$0.22 & 76.3787$\pm$0.24 & 75.0400$\pm$0.25 & 73.7094$\pm$0.27 \\
GRAD. &   \textbf{LED}  & 57.0161$\pm$0.25 & 55.8402$\pm$0.41 & 59.8674$\pm$0.34 & 61.1083$\pm$0.29 & \textbf{61.4335$\pm$0.32} & 61.3185$\pm$0.30 & 60.9215$\pm$0.30 & 60.4125$\pm$0.32 \\
10K  &  \textbf{MIXED}  & 57.6343$\pm$0.12 & 59.4987$\pm$0.38 & \textbf{79.5089$\pm$0.18} & 77.4944$\pm$0.18 & 74.5938$\pm$0.18 & 71.4096$\pm$0.17 & 67.9685$\pm$0.15 & 64.3678$\pm$0.14 \\
 &   \textbf{RandRBF}   & 30.9061$\pm$0.68 & 30.8662$\pm$0.68 & 27.7281$\pm$0.27 & 28.8236$\pm$0.34 & 29.7639$\pm$0.33 & 30.4933$\pm$0.28 & 31.1235$\pm$0.31 & \textbf{31.6047$\pm$0.33} \\
 &      \textbf{SINE}   & 56.3515$\pm$0.10 & 59.2024$\pm$0.28 & \textbf{79.2225$\pm$0.18} & 77.4134$\pm$0.18 & 75.1024$\pm$0.20 & 72.4819$\pm$0.16 & 69.7079$\pm$0.18 & 66.8150$\pm$0.19 \\
\bottomrule 
\end{tabular} 
\end{adjustbox} 
\end{table}

\begin{table} 
\caption{Mean accuracies of classifiers with the RDDM detector, in the abrupt datasets, in percentage (\%), with a 95\% confidence interval (k=30)} 
\label{tab:det-acc-RDDM-Abr30} 
\begin{adjustbox}{max width=\textwidth} 
\begin{tabular}{cccccccccc} 
\toprule 
DS Type  & \textbf{RDDM} & \multicolumn{2}{c}{\textbf{CLASSIFIER}} & \multicolumn{6}{l}{kNN --- with k = 30, varying the size of the window (w)} \\ 
and Size & DATASET & NB & HT & w=600 & w=800 & w=1000 & w=1200 & w=1400 & w=1600 \\ 
\toprule 
 & \textbf{AGRAW$_{1}$} & 63.5577$\pm$0.28 & 64.6877$\pm$0.32 & \textbf{67.1350$\pm$0.21} & 66.8672$\pm$0.19 & 66.1983$\pm$0.21 & 65.6651$\pm$0.24 & 65.4927$\pm$0.21 & 65.4655$\pm$0.21 \\
 & \textbf{AGRAW$_{2}$} & 79.6292$\pm$1.01 & \textbf{81.5760$\pm$0.88} & 80.8054$\pm$0.19 & 79.9589$\pm$0.24 & 78.9740$\pm$0.28 & 78.1470$\pm$0.36 & 77.8296$\pm$0.48 & 77.4983$\pm$0.54 \\
ABRUPT &  \textbf{LED}  & \textbf{69.8033$\pm$0.30} & 69.7821$\pm$0.31 & 60.1756$\pm$0.42 & 61.4042$\pm$0.40 & 62.0506$\pm$0.41 & 62.3654$\pm$0.36 & 62.5525$\pm$0.37 & 62.6411$\pm$0.42 \\
10K  &  \textbf{MIXED}  & 90.2191$\pm$0.24 & 90.1677$\pm$0.25 & 90.6887$\pm$0.19 & 91.2581$\pm$0.17 & 91.4531$\pm$0.18 & 91.6181$\pm$0.18 & 91.7236$\pm$0.19 & \textbf{91.7549$\pm$0.19} \\
 &   \textbf{RandRBF}   & 30.5285$\pm$0.45 & 32.0130$\pm$0.41 & 29.7601$\pm$0.29 & 30.8671$\pm$0.30 & 31.8088$\pm$0.33 & 32.3540$\pm$0.30 & 32.7399$\pm$0.28 & \textbf{33.1623$\pm$0.33} \\
 &      \textbf{SINE}   & 86.5796$\pm$0.25 & 87.9844$\pm$0.21 & 88.3579$\pm$0.21 & 88.7910$\pm$0.20 & 89.0756$\pm$0.20 & 89.2508$\pm$0.18 & 89.3240$\pm$0.19 & \textbf{89.3910$\pm$0.18} \\
\bottomrule 
\end{tabular} 
\end{adjustbox} 
\end{table}

\begin{table} 
\caption{Mean accuracies of classifiers with the RDDM detector, in the abrupt datasets, in percentage (\%), with a 95\% confidence interval (k=50)} 
\label{tab:det-acc-RDDM-Abr50} 
\begin{adjustbox}{max width=\textwidth} 
\begin{tabular}{cccccccccc} 
\toprule 
DS Type  & \textbf{RDDM} & \multicolumn{2}{c}{\textbf{CLASSIFIER}} & \multicolumn{6}{l}{kNN --- with k = 50, varying the size of the window (w)} \\ 
and Size & DATASET & NB & HT & w=600 & w=800 & w=1000 & w=1200 & w=1400 & w=1600 \\ 
\toprule 
 & \textbf{AGRAW$_{1}$} & 63.5577$\pm$0.28 & 64.6877$\pm$0.32 & 65.7916$\pm$0.18 & 66.6430$\pm$0.16 & \textbf{67.0335$\pm$0.20} & 66.9132$\pm$0.18 & 66.7771$\pm$0.20 & 66.6144$\pm$0.18 \\
 & \textbf{AGRAW$_{2}$} & 79.6292$\pm$1.01 & \textbf{81.5760$\pm$0.88} & 81.1501$\pm$0.22 & 81.2219$\pm$0.24 & 81.1279$\pm$0.31 & 80.6356$\pm$0.32 & 80.3096$\pm$0.38 & 79.6653$\pm$0.44 \\
ABRUPT &  \textbf{LED}  & \textbf{69.8033$\pm$0.30} & 69.7821$\pm$0.31 & 59.8452$\pm$0.39 & 61.2983$\pm$0.35 & 62.0373$\pm$0.38 & 62.3398$\pm$0.35 & 62.4046$\pm$0.38 & 62.5849$\pm$0.38 \\
10K  &  \textbf{MIXED}  & \textbf{90.2191$\pm$0.24} & 90.1677$\pm$0.25 & 87.7845$\pm$0.23 & 88.7057$\pm$0.25 & 89.1890$\pm$0.25 & 89.4569$\pm$0.25 & 89.5874$\pm$0.23 & 89.6273$\pm$0.24 \\
 &   \textbf{RandRBF}   & 30.5285$\pm$0.45 & \textbf{32.0130$\pm$0.41} & 27.8014$\pm$0.27 & 28.6218$\pm$0.36 & 29.4883$\pm$0.34 & 30.0018$\pm$0.28 & 30.4386$\pm$0.31 & 30.9027$\pm$0.34 \\
 &      \textbf{SINE}   & 86.5796$\pm$0.25 & 87.9844$\pm$0.21 & 86.8872$\pm$0.17 & 87.5173$\pm$0.21 & 87.8298$\pm$0.21 & 88.0175$\pm$0.20 & 88.1487$\pm$0.19 & \textbf{88.1890$\pm$0.21} \\
\bottomrule 
\end{tabular} 
\end{adjustbox} 
\end{table}

\begin{table} 
\caption{Mean accuracies of classifiers with the RDDM detector, in the gradual datasets, in percentage (\%), with a 95\% confidence interval (k=30)} 
\label{tab:det-acc-RDDM-Grad30} 
\begin{adjustbox}{max width=\textwidth} 
\begin{tabular}{cccccccccc} 
\toprule 
DS Type  & \textbf{RDDM} & \multicolumn{2}{c}{\textbf{CLASSIFIER}} & \multicolumn{6}{l}{kNN --- with k = 30, varying the size of the window (w)} \\ 
and Size & DATASET & NB & HT & w=600 & w=800 & w=1000 & w=1200 & w=1400 & w=1600 \\ 
\toprule 
 & \textbf{AGRAW$_{1}$} & 62.0529$\pm$0.29 & 62.9242$\pm$0.28 & \textbf{66.4353$\pm$0.19} & 66.0887$\pm$0.17 & 65.2340$\pm$0.20 & 64.6966$\pm$0.23 & 64.4591$\pm$0.21 & 64.4977$\pm$0.20 \\
 & \textbf{AGRAW$_{2}$} & 74.2929$\pm$1.50 & 78.6547$\pm$1.02 & \textbf{79.0769$\pm$0.20} & 78.3941$\pm$0.22 & 77.4251$\pm$0.27 & 76.6098$\pm$0.34 & 76.0011$\pm$0.34 & 75.7271$\pm$0.45 \\
GRAD. &   \textbf{LED}  & \textbf{67.8506$\pm$0.30} & 67.8066$\pm$0.30 & 59.5479$\pm$0.40 & 60.6033$\pm$0.36 & 60.9922$\pm$0.39 & 61.1815$\pm$0.33 & 61.2638$\pm$0.33 & 61.1837$\pm$0.37 \\
10K  &  \textbf{MIXED}  & 83.8911$\pm$0.30 & 83.6951$\pm$0.32 & 84.8636$\pm$0.23 & 85.3415$\pm$0.18 & 85.5335$\pm$0.17 & 85.5721$\pm$0.17 & 85.6458$\pm$0.18 & \textbf{85.7418$\pm$0.20} \\
 &   \textbf{RandRBF}   & 30.3931$\pm$0.46 & 31.9290$\pm$0.40 & 29.6924$\pm$0.28 & 30.8771$\pm$0.26 & 31.7877$\pm$0.29 & 32.3534$\pm$0.29 & 32.8638$\pm$0.28 & \textbf{33.2568$\pm$0.32} \\
 &      \textbf{SINE}   & 81.8545$\pm$0.19 & 82.6644$\pm$0.20 & 82.7569$\pm$0.21 & 83.0501$\pm$0.21 & 83.2879$\pm$0.22 & 83.3327$\pm$0.21 & 83.4463$\pm$0.22 & \textbf{83.5191$\pm$0.21} \\
\bottomrule 
\end{tabular} 
\end{adjustbox} 
\end{table}

\begin{table} 
\caption{Mean accuracies of classifiers with the RDDM detector, in the gradual datasets, in percentage (\%), with a 95\% confidence interval (k=50)} 
\label{tab:det-acc-RDDM-Grad50} 
\begin{adjustbox}{max width=\textwidth} 
\begin{tabular}{cccccccccc} 
\toprule 
DS Type  & \textbf{RDDM} & \multicolumn{2}{c}{\textbf{CLASSIFIER}} & \multicolumn{6}{l}{kNN --- with k = 50, varying the size of the window (w)} \\ 
and Size & DATASET & NB & HT & w=600 & w=800 & w=1000 & w=1200 & w=1400 & w=1600 \\ 
\toprule
 & \textbf{AGRAW$_{1}$} & 62.0529$\pm$0.29 & 62.9242$\pm$0.28 & 65.3192$\pm$0.18 & 66.0058$\pm$0.20 & \textbf{66.1660$\pm$0.19} & 65.9864$\pm$0.21 & 65.6753$\pm$0.16 & 65.6112$\pm$0.18 \\
 & \textbf{AGRAW$_{2}$} & 74.2929$\pm$1.50 & 78.6547$\pm$1.02 & 79.4239$\pm$0.23 & \textbf{79.4941$\pm$0.21} & 79.3221$\pm$0.24 & 78.8554$\pm$0.28 & 78.4384$\pm$0.32 & 77.9278$\pm$0.36 \\
GRAD. &   \textbf{LED}  & \textbf{67.8506$\pm$0.30} & 67.8066$\pm$0.30 & 59.5147$\pm$0.35 & 60.6729$\pm$0.35 & 61.0872$\pm$0.34 & 61.2258$\pm$0.37 & 61.2151$\pm$0.37 & 61.2115$\pm$0.36 \\
10K  &  \textbf{MIXED}  & 83.8911$\pm$0.30 & 83.6951$\pm$0.32 & 82.8372$\pm$0.24 & 83.6314$\pm$0.22 & 84.0258$\pm$0.22 & 84.2503$\pm$0.23 & 84.3524$\pm$0.21 & \textbf{84.3713$\pm$0.25} \\
 &   \textbf{RandRBF}   & 30.3931$\pm$0.46 & \textbf{31.9290$\pm$0.40} & 27.6112$\pm$0.28 & 28.6599$\pm$0.36 & 29.4716$\pm$0.34 & 30.1120$\pm$0.30 & 30.5867$\pm$0.32 & 30.9390$\pm$0.32 \\
 &      \textbf{SINE}   & 81.8545$\pm$0.19 & 82.6644$\pm$0.20 & 81.6859$\pm$0.19 & 82.1317$\pm$0.18 & 82.3994$\pm$0.17 & 82.5861$\pm$0.15 & 82.6617$\pm$0.19 & \textbf{82.7029$\pm$0.19} \\
\bottomrule 
\end{tabular} 
\end{adjustbox} 
\end{table}

\begin{table} 
\caption{Mean run-time of classifiers with no drift detector, in the abrupt datasets, in seconds, with a 95\% confidence interval (k=30)} 
\label{tab:det-time-NoDet-Abr30} 
\begin{adjustbox}{max width=\textwidth} 
\begin{tabular}{cccccccccc} 
\toprule 
DS Type  & \textbf{NO DET.} & \multicolumn{2}{c}{\textbf{CLASSIFIER}} & \multicolumn{6}{l}{kNN --- with k = 30, varying the size of the window (w)} \\ 
and Size & DATASET & NB & HT & w=600 & w=800 & w=1000 & w=1200 & w=1400 & w=1600 \\ 
\toprule 
 & \textbf{AGRAW$_{1}$} & \textbf{0.098$\pm$0.01} & 0.126$\pm$0.03 & 0.381$\pm$0.01 & 0.814$\pm$0.04 & 1.030$\pm$0.04 & 1.134$\pm$0.04 & 1.285$\pm$0.04 & 1.444$\pm$0.06 \\
 & \textbf{AGRAW$_{2}$} & 0.093$\pm$0.01 & \textbf{0.079$\pm$0.02} & 0.376$\pm$0.01 & 0.852$\pm$0.01 & 1.047$\pm$0.03 & 1.119$\pm$0.04 & 1.297$\pm$0.04 & 1.406$\pm$0.05 \\
ABRUPT & \textbf{LED} & 0.216$\pm$0.02 & \textbf{0.193$\pm$0.02} & 0.948$\pm$0.01 & 1.949$\pm$0.07 & 2.520$\pm$0.05 & 2.714$\pm$0.10 & 3.337$\pm$0.09 & 3.630$\pm$0.12 \\
10K & \textbf{MIXED} & \textbf{0.042$\pm$0.01} & 0.053$\pm$0.02 & 0.250$\pm$0.01 & 0.594$\pm$0.02 & 0.739$\pm$0.03 & 0.850$\pm$0.02 & 0.946$\pm$0.03 & 1.019$\pm$0.04 \\
 & \textbf{RandRBF} & \textbf{0.291$\pm$0.02} & 0.368$\pm$0.02 & 0.783$\pm$0.01 & 1.611$\pm$0.07 & 2.019$\pm$0.08 & 2.307$\pm$0.03 & 2.543$\pm$0.10 & 3.220$\pm$0.10 \\
 & \textbf{SINE} & \textbf{0.040$\pm$0.01} & 0.077$\pm$0.02 & 0.272$\pm$0.01 & 0.605$\pm$0.04 & 0.825$\pm$0.02 & 0.854$\pm$0.04 & 1.002$\pm$0.05 & 1.116$\pm$0.04 \\
\bottomrule 
\end{tabular} 
\end{adjustbox} 
\end{table}

\begin{table} 
\caption{Mean run-time of classifiers with no drift detector, in the abrupt datasets, in seconds, with a 95\% confidence interval (k=40)} 
\label{tab:det-time-NoDet-Abr40} 
\begin{adjustbox}{max width=\textwidth} 
\begin{tabular}{cccccccccc} 
\toprule 
DS Type  & \textbf{NO DET.} & \multicolumn{2}{c}{\textbf{CLASSIFIER}} & \multicolumn{6}{l}{kNN --- with k = 40, varying the size of the window (w)} \\ 
and Size & DATASET & NB & HT & w=600 & w=800 & w=1000 & w=1200 & w=1400 & w=1600 \\ 
\toprule  
 & \textbf{AGRAW$_{1}$} & \textbf{0.098$\pm$0.01} & 0.126$\pm$0.03 & 0.696$\pm$0.03 & 0.862$\pm$0.02 & 1.038$\pm$0.03 & 0.679$\pm$0.01 & 1.263$\pm$0.05 & 1.447$\pm$0.04 \\
 & \textbf{AGRAW$_{2}$} & 0.093$\pm$0.01 & \textbf{0.079$\pm$0.02} & 0.683$\pm$0.02 & 0.848$\pm$0.04 & 1.019$\pm$0.03 & 0.685$\pm$0.01 & 1.350$\pm$0.04 & 1.494$\pm$0.03 \\
ABRUPT & \textbf{LED} & 0.216$\pm$0.02 & \textbf{0.193$\pm$0.02} & 1.601$\pm$0.05 & 2.114$\pm$0.04 & 2.497$\pm$0.08 & 1.902$\pm$0.12 & 3.381$\pm$0.08 & 3.732$\pm$0.17 \\
10K & \textbf{MIXED} & \textbf{0.042$\pm$0.01} & 0.053$\pm$0.02 & 0.490$\pm$0.02 & 0.646$\pm$0.01 & 0.752$\pm$0.02 & 0.461$\pm$0.01 & 0.985$\pm$0.03 & 1.088$\pm$0.04 \\
 & \textbf{RandRBF} & \textbf{0.291$\pm$0.02} & 0.368$\pm$0.02 & 1.404$\pm$0.02 & 1.738$\pm$0.07 & 2.212$\pm$0.03 & 1.593$\pm$0.13 & 3.013$\pm$0.08 & 3.287$\pm$0.11 \\
 & \textbf{SINE} & \textbf{0.040$\pm$0.01} & 0.077$\pm$0.02 & 0.521$\pm$0.02 & 0.646$\pm$0.04 & 0.751$\pm$0.04 & 0.520$\pm$0.01 & 1.036$\pm$0.03 & 1.140$\pm$0.03 \\
\bottomrule 
\end{tabular} 
\end{adjustbox} 
\end{table}

\begin{table} 
\caption{Mean run-time of classifiers with no drift detector, in the abrupt datasets, in seconds, with a 95\% confidence interval (k=50)} 
\label{tab:det-time-NoDet-Abr50} 
\begin{adjustbox}{max width=\textwidth} 
\begin{tabular}{cccccccccc} 
\toprule 
DS Type  & \textbf{NO DET.} & \multicolumn{2}{c}{\textbf{CLASSIFIER}} & \multicolumn{6}{l}{kNN --- with k = 50, varying the size of the window (w)} \\ 
and Size & DATASET & NB & HT & w=600 & w=800 & w=1000 & w=1200 & w=1400 & w=1600 \\ 
\toprule  
 & \textbf{AGRAW$_{1}$} & \textbf{0.098$\pm$0.01} & 0.126$\pm$0.03 & 0.708$\pm$0.02 & 0.902$\pm$0.02 & 1.022$\pm$0.04 & 1.226$\pm$0.03 & 1.333$\pm$0.05 & 1.479$\pm$0.06 \\
 & \textbf{AGRAW$_{2}$} & 0.093$\pm$0.01 & \textbf{0.079$\pm$0.02} & 0.684$\pm$0.03 & 0.863$\pm$0.03 & 1.072$\pm$0.04 & 1.214$\pm$0.02 & 1.304$\pm$0.05 & 1.587$\pm$0.06 \\
ABRUPT & \textbf{LED} & 0.216$\pm$0.02 & \textbf{0.193$\pm$0.02} & 1.612$\pm$0.05 & 2.076$\pm$0.04 & 2.544$\pm$0.05 & 2.978$\pm$0.04 & 3.454$\pm$0.13 & 3.964$\pm$0.15 \\
10K & \textbf{MIXED} & \textbf{0.042$\pm$0.01} & 0.053$\pm$0.02 & 0.494$\pm$0.02 & 0.662$\pm$0.01 & 0.760$\pm$0.04 & 0.879$\pm$0.04 & 0.994$\pm$0.03 & 1.096$\pm$0.04 \\
 & \textbf{RandRBF} & \textbf{0.291$\pm$0.02} & 0.368$\pm$0.02 & 1.516$\pm$0.05 & 1.688$\pm$0.11 & 2.251$\pm$0.05 & 2.497$\pm$0.09 & 2.911$\pm$0.12 & 3.493$\pm$0.13 \\
 & \textbf{SINE} & \textbf{0.040$\pm$0.01} & 0.077$\pm$0.02 & 0.546$\pm$0.02 & 0.685$\pm$0.02 & 0.840$\pm$0.02 & 0.945$\pm$0.04 & 1.087$\pm$0.02 & 1.164$\pm$0.03 \\
\bottomrule 
\end{tabular} 
\end{adjustbox} 
\end{table}

\begin{table} 
\caption{Mean run-time of classifiers with no drift detector, in the gradual datasets, in seconds, with a 95\% confidence interval (k=30)} 
\label{tab:det-time-NoDet-Grad30} 
\begin{adjustbox}{max width=\textwidth} 
\begin{tabular}{cccccccccc} 
\toprule 
DS Type  & \textbf{NO DET.} & \multicolumn{2}{c}{\textbf{CLASSIFIER}} & \multicolumn{6}{l}{kNN --- with k = 30, varying the size of the window (w)} \\ 
and Size & DATASET & NB & HT & w=600 & w=800 & w=1000 & w=1200 & w=1400 & w=1600 \\ 
\toprule 
 & \textbf{AGRAW$_{1}$} & \textbf{0.085$\pm$0.01} & 0.124$\pm$0.03 & 0.669$\pm$0.02 & 0.888$\pm$0.02 & 1.043$\pm$0.03 & 1.213$\pm$0.02 & 1.298$\pm$0.05 & 1.447$\pm$0.05 \\
 & \textbf{AGRAW$_{2}$} & 0.085$\pm$0.01 & \textbf{0.078$\pm$0.02} & 0.685$\pm$0.01 & 0.866$\pm$0.02 & 1.033$\pm$0.03 & 1.031$\pm$0.06 & 1.284$\pm$0.05 & 1.441$\pm$0.04 \\
GRAD. & \textbf{LED} & 0.203$\pm$0.02 & \textbf{0.188$\pm$0.03} & 1.579$\pm$0.05 & 2.053$\pm$0.03 & 2.461$\pm$0.10 & 2.821$\pm$0.08 & 3.400$\pm$0.11 & 3.805$\pm$0.16 \\
10K & \textbf{MIXED} & \textbf{0.040$\pm$0.01} & 0.048$\pm$0.02 & 0.424$\pm$0.03 & 0.586$\pm$0.03 & 0.778$\pm$0.03 & 0.748$\pm$0.05 & 0.937$\pm$0.04 & 1.042$\pm$0.04 \\
 & \textbf{RandRBF} & \textbf{0.282$\pm$0.01} & 0.372$\pm$0.03 & 1.337$\pm$0.06 & 1.736$\pm$0.05 & 2.101$\pm$0.06 & 2.200$\pm$0.12 & 2.778$\pm$0.09 & 3.171$\pm$0.14 \\
 & \textbf{SINE} & \textbf{0.037$\pm$0.01} & 0.074$\pm$0.02 & 0.494$\pm$0.03 & 0.660$\pm$0.03 & 0.811$\pm$0.02 & 0.897$\pm$0.02 & 0.976$\pm$0.04 & 1.194$\pm$0.02 \\
\bottomrule 
\end{tabular} 
\end{adjustbox} 
\end{table}

\begin{table} 
\caption{Mean run-time of classifiers with no drift detector, in the gradual datasets, in seconds, with a 95\% confidence interval (k=40)} 
\label{tab:det-time-NoDet-Grad40} 
\begin{adjustbox}{max width=\textwidth} 
\begin{tabular}{cccccccccc} 
\toprule 
DS Type  & \textbf{NO DET.} & \multicolumn{2}{c}{\textbf{CLASSIFIER}} & \multicolumn{6}{l}{kNN --- with k = 40, varying the size of the window (w)} \\ 
and Size & DATASET & NB & HT & w=600 & w=800 & w=1000 & w=1200 & w=1400 & w=1600 \\ 
\toprule  
 & \textbf{AGRAW$_{1}$} & \textbf{0.085$\pm$0.01} & 0.124$\pm$0.03 & 0.709$\pm$0.01 & 0.884$\pm$0.01 & 1.042$\pm$0.02 & 1.199$\pm$0.03 & 1.362$\pm$0.04 & 1.475$\pm$0.03 \\
 & \textbf{AGRAW$_{2}$} & 0.085$\pm$0.01 & \textbf{0.078$\pm$0.02} & 0.689$\pm$0.03 & 0.865$\pm$0.02 & 1.042$\pm$0.02 & 1.200$\pm$0.02 & 1.354$\pm$0.02 & 1.464$\pm$0.05 \\
GRAD. & \textbf{LED} & 0.203$\pm$0.02 & \textbf{0.188$\pm$0.03} & 1.611$\pm$0.04 & 2.034$\pm$0.06 & 2.446$\pm$0.08 & 3.003$\pm$0.14 & 3.210$\pm$0.07 & 3.840$\pm$0.08 \\
10K & \textbf{MIXED} & \textbf{0.040$\pm$0.01} & 0.048$\pm$0.02 & 0.497$\pm$0.02 & 0.609$\pm$0.04 & 0.760$\pm$0.02 & 0.851$\pm$0.03 & 1.010$\pm$0.02 & 1.078$\pm$0.04 \\
 & \textbf{RandRBF} & \textbf{0.282$\pm$0.01} & 0.372$\pm$0.03 & 1.413$\pm$0.05 & 1.820$\pm$0.04 & 2.070$\pm$0.09 & 2.526$\pm$0.10 & 2.848$\pm$0.08 & 3.279$\pm$0.10 \\
 & \textbf{SINE} & \textbf{0.037$\pm$0.01} & 0.074$\pm$0.02 & 0.521$\pm$0.02 & 0.660$\pm$0.04 & 0.705$\pm$0.04 & 0.926$\pm$0.02 & 1.054$\pm$0.03 & 1.180$\pm$0.03 \\
\bottomrule 
\end{tabular} 
\end{adjustbox} 
\end{table}

\begin{table} 
\caption{Mean run-time of classifiers with no drift detector, in the gradual datasets, in seconds, with a 95\% confidence interval (k=50)} 
\label{tab:det-time-NoDet-Grad50} 
\begin{adjustbox}{max width=\textwidth} 
\begin{tabular}{cccccccccc} 
\toprule 
DS Type  & \textbf{NO DET.} & \multicolumn{2}{c}{\textbf{CLASSIFIER}} & \multicolumn{6}{l}{kNN --- with k = 50, varying the size of the window (w)} \\ 
and Size & DATASET & NB & HT & w=600 & w=800 & w=1000 & w=1200 & w=1400 & w=1600 \\ 
\toprule 
 & \textbf{AGRAW$_{1}$} & \textbf{0.085$\pm$0.01} & 0.124$\pm$0.03 & 0.669$\pm$0.04 & 0.791$\pm$0.05 & 1.054$\pm$0.03 & 1.148$\pm$0.05 & 1.361$\pm$0.05 & 1.497$\pm$0.06 \\
 & \textbf{AGRAW$_{2}$} & 0.085$\pm$0.01 & \textbf{0.078$\pm$0.02} & 0.648$\pm$0.03 & 0.894$\pm$0.02 & 1.076$\pm$0.01 & 1.206$\pm$0.04 & 1.384$\pm$0.05 & 1.507$\pm$0.05 \\
GRAD. & \textbf{LED} & 0.203$\pm$0.02 & \textbf{0.188$\pm$0.03} & 1.638$\pm$0.04 & 2.068$\pm$0.05 & 2.553$\pm$0.06 & 2.918$\pm$0.07 & 3.608$\pm$0.13 & 4.004$\pm$0.10 \\
10K & \textbf{MIXED} & \textbf{0.040$\pm$0.01} & 0.048$\pm$0.02 & 0.487$\pm$0.03 & 0.658$\pm$0.02 & 0.725$\pm$0.04 & 0.888$\pm$0.03 & 0.919$\pm$0.07 & 1.147$\pm$0.02 \\
 & \textbf{RandRBF} & \textbf{0.282$\pm$0.01} & 0.372$\pm$0.03 & 1.489$\pm$0.06 & 1.852$\pm$0.07 & 2.157$\pm$0.11 & 2.495$\pm$0.11 & 3.062$\pm$0.14 & 3.530$\pm$0.13 \\
 & \textbf{SINE} & \textbf{0.037$\pm$0.01} & 0.074$\pm$0.02 & 0.511$\pm$0.03 & 0.694$\pm$0.01 & 0.817$\pm$0.03 & 0.963$\pm$0.03 & 1.100$\pm$0.03 & 1.159$\pm$0.06 \\
\bottomrule 
\end{tabular} 
\end{adjustbox} 
\end{table}

\begin{table} 
\caption{Mean run-time of classifiers with the RDDM detector, in the abrupt datasets, in seconds, with a 95\% confidence interval (k=30)} 
\label{tab:det-time-RDDM-Abr30} 
\begin{adjustbox}{max width=\textwidth} 
\begin{tabular}{cccccccccc} 
\toprule
DS Type  & \textbf{RDDM} & \multicolumn{2}{c}{\textbf{CLASSIFIER}} & \multicolumn{6}{l}{kNN --- with k = 30, varying the size of the window (w)} \\ 
and Size & DATASET & NB & HT & w=600 & w=800 & w=1000 & w=1200 & w=1400 & w=1600 \\ 
\toprule
 & \textbf{AGRAW$_{1}$} & \textbf{0.319$\pm$0.02} & 0.518$\pm$0.03 & 2.720$\pm$0.09 & 3.211$\pm$0.11 & 3.599$\pm$0.12 & 3.670$\pm$0.11 & 4.211$\pm$0.12 & 4.249$\pm$0.17 \\
 & \textbf{AGRAW$_{2}$} & \textbf{0.297$\pm$0.02} & 0.577$\pm$0.03 & 2.891$\pm$0.09 & 3.516$\pm$0.11 & 4.000$\pm$0.13 & 4.306$\pm$0.09 & 4.955$\pm$0.19 & 5.098$\pm$0.21 \\
ABRUPT & \textbf{LED} & \textbf{0.825$\pm$0.03} & 1.088$\pm$0.04 & 4.728$\pm$0.10 & 5.801$\pm$0.13 & 6.580$\pm$0.13 & 6.639$\pm$0.22 & 7.815$\pm$0.16 & 8.146$\pm$0.15 \\
10K & \textbf{MIXED} & \textbf{0.159$\pm$0.01} & 0.277$\pm$0.02 & 1.990$\pm$0.06 & 2.555$\pm$0.08 & 2.819$\pm$0.08 & 2.923$\pm$0.07 & 3.432$\pm$0.07 & 3.451$\pm$0.09 \\
 & \textbf{RandRBF} & \textbf{1.464$\pm$0.06} & 2.355$\pm$0.07 & 4.387$\pm$0.09 & 5.393$\pm$0.12 & 6.261$\pm$0.16 & 5.981$\pm$0.22 & 7.654$\pm$0.16 & 8.463$\pm$0.22 \\
 & \textbf{SINE} & \textbf{0.181$\pm$0.01} & 0.487$\pm$0.04 & 2.046$\pm$0.06 & 2.656$\pm$0.07 & 2.996$\pm$0.08 & 3.023$\pm$0.09 & 3.604$\pm$0.09 & 3.546$\pm$0.11 \\
\bottomrule 
\end{tabular} 
\end{adjustbox} 
\end{table}

\begin{table} 
\caption{Mean run-time of classifiers with the RDDM detector, in the abrupt datasets, in seconds, with a 95\% confidence interval (k=50)} 
\label{tab:det-time-RDDM-Abr50} 
\begin{adjustbox}{max width=\textwidth} 
\begin{tabular}{cccccccccc} 
\toprule
DS Type  & \textbf{RDDM} & \multicolumn{2}{c}{\textbf{CLASSIFIER}} & \multicolumn{6}{l}{kNN --- with k = 50, varying the size of the window (w)} \\ 
and Size & DATASET & NB & HT & w=600 & w=800 & w=1000 & w=1200 & w=1400 & w=1600 \\ 
\toprule
 & \textbf{AGRAW$_{1}$} & \textbf{0.319$\pm$0.02} & 0.518$\pm$0.03 & 2.815$\pm$0.09 & 3.385$\pm$0.09 & 3.622$\pm$0.15 & 3.990$\pm$0.14 & 4.222$\pm$0.19 & 4.363$\pm$0.16 \\
 & \textbf{AGRAW$_{2}$} & \textbf{0.297$\pm$0.02} & 0.577$\pm$0.03 & 2.945$\pm$0.06 & 3.620$\pm$0.11 & 3.961$\pm$0.14 & 4.541$\pm$0.16 & 4.995$\pm$0.16 & 5.304$\pm$0.25 \\
ABRUPT & \textbf{LED} & \textbf{0.825$\pm$0.03} & 1.088$\pm$0.04 & 4.874$\pm$0.10 & 5.981$\pm$0.13 & 6.761$\pm$0.16 & 7.491$\pm$0.14 & 7.913$\pm$0.17 & 8.248$\pm$0.16 \\
10K & \textbf{MIXED} & \textbf{0.159$\pm$0.01} & 0.277$\pm$0.02 & 2.054$\pm$0.07 & 2.670$\pm$0.08 & 2.961$\pm$0.06 & 3.253$\pm$0.06 & 3.509$\pm$0.10 & 3.555$\pm$0.09 \\
 & \textbf{RandRBF} & \textbf{1.464$\pm$0.06} & 2.355$\pm$0.07 & 4.586$\pm$0.11 & 5.485$\pm$0.20 & 6.438$\pm$0.18 & 7.236$\pm$0.23 & 8.055$\pm$0.24 & 8.595$\pm$0.25 \\
 & \textbf{SINE} & \textbf{0.181$\pm$0.01} & 0.487$\pm$0.04 & 2.131$\pm$0.05 & 2.735$\pm$0.08 & 2.967$\pm$0.07 & 3.247$\pm$0.08 & 3.560$\pm$0.11 & 3.717$\pm$0.08 \\
\bottomrule 
\end{tabular} 
\end{adjustbox} 
\end{table}

\begin{table} 
\caption{Mean run-time of classifiers with the RDDM detector, in the gradual datasets, in seconds, with a 95\% confidence interval (k=30)} 
\label{tab:det-time-RDDM-Grad30} 
\begin{adjustbox}{max width=\textwidth} 
\begin{tabular}{cccccccccc} 
\toprule
DS Type  & \textbf{RDDM} & \multicolumn{2}{c}{\textbf{CLASSIFIER}} & \multicolumn{6}{l}{kNN --- with k = 30, varying the size of the window (w)} \\ 
and Size & DATASET & NB & HT & w=600 & w=800 & w=1000 & w=1200 & w=1400 & w=1600 \\ 
\toprule
 & \textbf{AGRAW$_{1}$} & \textbf{0.314$\pm$0.02} & 0.479$\pm$0.04 & 2.793$\pm$0.09 & 3.414$\pm$0.11 & 3.666$\pm$0.12 & 3.992$\pm$0.13 & 4.163$\pm$0.17 & 4.188$\pm$0.18 \\
 & \textbf{AGRAW$_{2}$} & \textbf{0.316$\pm$0.02} & 0.538$\pm$0.03 & 2.908$\pm$0.07 & 3.597$\pm$0.07 & 4.064$\pm$0.12 & 4.545$\pm$0.12 & 4.998$\pm$0.17 & 5.303$\pm$0.14 \\
GRAD. & \textbf{LED} & \textbf{0.868$\pm$0.04} & 1.045$\pm$0.04 & 4.975$\pm$0.09 & 5.908$\pm$0.10 & 6.862$\pm$0.17 & 7.358$\pm$0.16 & 7.977$\pm$0.20 & 8.197$\pm$0.18 \\
10K & \textbf{MIXED} & \textbf{0.157$\pm$0.02} & 0.263$\pm$0.03 & 1.930$\pm$0.05 & 2.439$\pm$0.09 & 2.753$\pm$0.08 & 2.928$\pm$0.08 & 3.079$\pm$0.09 & 3.170$\pm$0.10 \\
 & \textbf{RandRBF} & \textbf{1.496$\pm$0.05} & 2.289$\pm$0.07 & 4.410$\pm$0.09 & 5.287$\pm$0.13 & 6.228$\pm$0.16 & 6.796$\pm$0.28 & 7.621$\pm$0.23 & 8.405$\pm$0.33 \\
 & \textbf{SINE} & \textbf{0.182$\pm$0.02} & 0.387$\pm$0.03 & 1.974$\pm$0.06 & 2.573$\pm$0.06 & 2.856$\pm$0.10 & 3.013$\pm$0.09 & 3.213$\pm$0.13 & 3.430$\pm$0.10 \\
\bottomrule 
\end{tabular} 
\end{adjustbox} 
\end{table}

\begin{table} 
\caption{Mean run-time of classifiers with the RDDM detector, in the gradual datasets, in seconds, with a 95\% confidence interval (k=50)} 
\label{tab:det-time-RDDM-Grad50} 
\begin{adjustbox}{max width=\textwidth} 
\begin{tabular}{cccccccccc} 
\toprule
DS Type  & \textbf{RDDM} & \multicolumn{2}{c}{\textbf{CLASSIFIER}} & \multicolumn{6}{l}{kNN --- with k = 50, varying the size of the window (w)} \\ 
and Size & DATASET & NB & HT & w=600 & w=800 & w=1000 & w=1200 & w=1400 & w=1600 \\ 
\toprule
 & \textbf{AGRAW$_{1}$} & \textbf{0.314$\pm$0.02} & 0.479$\pm$0.04 & 2.899$\pm$0.06 & 3.406$\pm$0.11 & 3.740$\pm$0.13 & 4.132$\pm$0.15 & 4.351$\pm$0.18 & 4.493$\pm$0.17 \\
 & \textbf{AGRAW$_{2}$} & \textbf{0.316$\pm$0.02} & 0.538$\pm$0.03 & 2.918$\pm$0.09 & 3.546$\pm$0.13 & 4.009$\pm$0.12 & 4.584$\pm$0.14 & 4.954$\pm$0.21 & 5.489$\pm$0.20 \\
GRAD. & \textbf{LED} & \textbf{0.868$\pm$0.04} & 1.045$\pm$0.04 & 5.000$\pm$0.13 & 6.088$\pm$0.10 & 6.907$\pm$0.16 & 7.803$\pm$0.21 & 8.027$\pm$0.17 & 8.519$\pm$0.20 \\
10K & \textbf{MIXED} & \textbf{0.157$\pm$0.02} & 0.263$\pm$0.03 & 1.977$\pm$0.07 & 2.572$\pm$0.07 & 2.724$\pm$0.08 & 3.065$\pm$0.11 & 3.261$\pm$0.07 & 3.420$\pm$0.10 \\
 & \textbf{RandRBF} & \textbf{1.496$\pm$0.05} & 2.289$\pm$0.07 & 4.598$\pm$0.13 & 5.625$\pm$0.14 & 6.477$\pm$0.17 & 7.594$\pm$0.27 & 8.196$\pm$0.17 & 9.007$\pm$0.25 \\
 & \textbf{SINE} & \textbf{0.182$\pm$0.02} & 0.387$\pm$0.03 & 2.040$\pm$0.07 & 2.625$\pm$0.07 & 2.884$\pm$0.07 & 3.095$\pm$0.11 & 3.341$\pm$0.08 & 3.530$\pm$0.09 \\
\bottomrule 
\end{tabular} 
\end{adjustbox} 
\end{table}

\end{document}